\documentclass{article}

\usepackage{PRIMEarxiv}

\usepackage[utf8]{inputenc} 
\usepackage[T1]{fontenc}    
\usepackage{hyperref}       
\usepackage{url}            
\usepackage{booktabs}       
\usepackage{amsfonts}       
\usepackage{nicefrac}       
\usepackage{microtype}      
\usepackage{lipsum}
\usepackage{graphicx}
\graphicspath{{media/}}     

\usepackage[small]{caption}
\usepackage[switch]{lineno}
\usepackage{xcolor}         
\usepackage{times}
\usepackage{soul}
\usepackage{url}
\usepackage{amsmath}
\usepackage{amsthm}
\usepackage{cancel}
\usepackage{balance}
\usepackage{bm}
\usepackage{breqn}
\usepackage{multirow} 
\usepackage{wrapfig}
\usepackage{enumitem}
\usepackage{subfig}
\usepackage{algorithm}
\usepackage{algpseudocode}

\theoremstyle{plain}
\newtheorem{theorem}{Theorem}

\newtheorem{lemma}[theorem]{Lemma}
\newtheorem{corollary}[theorem]{Corollary}
\theoremstyle{definition}

\newtheorem{assumption}[theorem]{Assumption}
\theoremstyle{remark}

\title{Reducing Variance Caused by Communication in Decentralized Multi-agent Deep Reinforcement Learning
}

\author{
  Changxi Zhu \\
  Department of Information and Computing Sciences \\
  Utrecht University \\
  Utrecht, Netherlands\\
  \texttt{c.zhu@uu.nl} \\
   \And
  Mehdi Dastani \\
  Department of Information and Computing Sciences \\
  Utrecht University \\
  Utrecht, Netherlands\\
  \texttt{m.m.dastani@uu.nl} \\
  \AND
  Shihan Wang \\
  Department of Information and Computing Sciences \\
  Utrecht University \\
  Utrecht, Netherlands\\
  \texttt{s.wang2@uu.nl} \\
}

\begin{document}
\maketitle

\begin{abstract}
In decentralized multi-agent deep reinforcement learning (MADRL), communication can help agents to gain a better understanding of the environment to better coordinate their behaviors. Nevertheless, communication may involve uncertainty, which potentially introduces variance to the learning of decentralized agents. In this paper, we focus on a specific decentralized MADRL setting with communication and conduct a theoretical analysis to study the variance that is caused by communication in policy gradients. We propose modular techniques to reduce the variance in policy gradients during training. We adopt our modular techniques into two existing algorithms for decentralized MADRL with communication and evaluate them on multiple tasks in the StarCraft Multi-Agent Challenge and Traffic Junction domains. The results show that decentralized MADRL communication methods extended with our proposed techniques not only achieve high-performing agents but also reduce variance in policy gradients during training.
\end{abstract}


\section{Introduction}

Numerous real-world scenarios involve multiple agents interacting within a shared environment, spanning domains like autonomous driving \cite{Shai2016MARLAD}, robotics \cite{Kober2013Robotics}, and game playing \cite{Silver2017Go,Brown2019superhuman}. Multi-agent Deep Reinforcement Learning (MADRL) has been widely used to develop cooperative behaviors of agents in partially observable environments \cite{Gronauer2022Survey,Oroojlooyjadid2023Survey,Yang2020Survey}. MADRL agents can communicate various types of information, including observations, intentions, and experiences, to mitigate the limitations in agent observability and enhance the coordination of their behaviors \cite{Changxi2024Survey,Mohamed2019LearningComm,Gronauer2022Survey}. In recent years, there has been growing research interest in MADRL that focuses on communication via a vector or range of discrete or continuous values as encoded messages, rather than directly sharing agents' private and massive local information \cite{Changxi2024Survey}. These research works are known as MADRL with learning communication (Comm-MADRL), which aims to establish adaptive and learnable communication protocols. Within the Comm-MADRL field, several settings have been utilized, focusing on whether agents are trained in a centralized or decentralized manner, and whether communication occurs during training or policy execution \cite{Gronauer2022Survey,Changxi2024Survey}.

Practical considerations such as security and privacy, especially in realistic applications like UAVs \cite{Oubbati2022UAV}, require that agents act independently and keep control of their individual information during execution, i.e., no information sharing during policy execution. This leaves us to the following MADRL settings: Centralized Training and Decentralized Execution (CTDE), where communication is neither used during training nor during policy execution, or Decentralized Training and Decentralized Execution (DTDE), where communication can be used during training but not during policy execution. Considering centralized training in CTDE suffers from computational challenges \cite{Gronauer2022Survey,Gupta2017Cooperative}, we will focus on DTDE with communication setting. This setting, which is based on actor-critic methods, is known in the literature as \textbf{D}ecentralized \textbf{C}ommunicating \textbf{C}ritics and \textbf{D}ecentralized \textbf{A}ctors (DCCDA) \cite{Iqbal2019MAAC,Liu2020G2ANet}. The DCCDA setting enables communication among individual critics (Q-functions) during training while disabling communication among individual actors (policies) during both training and execution. In DCCDA, agents can enjoy the benefit of communication for better coordination and computational efficiency during training, while they can operate fully decentralized (without communication) during the execution. Despite the promising applications of DCCDA, communicated messages are often initiated in a stochastic manner \cite{Foerster2016Comm,Jiang2018ATOC}, adding uncertainty from the receiver's perspective. This can result in high variance in the policy gradient estimator of receiver agents, leading to low sample efficiency and performance degradation.

In this work, we conduct the first theoretical analysis of the variance in policy gradients within Comm-MADRL under the DCCDA setting. Variance analysis is a vigorous method that allows us to investigate the variability and dispersion of policy learning. Previous research has focused on variance analysis in policy gradients without communication \cite{Lyu2023Centralized,Lyu2021Contrasting}, and thus not measuring variance caused by communication. Specifically, Lyu et al. \cite{Lyu2023Centralized} analyze the variance in policy gradients under non-communication settings with CTDE and DTDE, where a centralized critic and decentralized critics are used, respectively, to guide the learning of decentralized actors. They claim that the variance in policy gradients under CTDE without communication can be equal to or higher than DTDE without communication, i.e., $Var(CTDE) \geq Var(DTDE)$.\footnote{For simplicity, we use CTDE to refer to CTDE without communication and the same for DTDE. For clarity, we add a figure in Appendix \ref{app:comSettings} to illustrate the differences in MADRL settings.} However, it remains unclear how communication among agents (in the DCCDA setting) affects the variance in policy gradients. Through our variance analysis, we prove that in both idealistic communication setting (where agents communicate sound \& complete information) and non-idealistic communication setting (where sound \& complete information is corrupted with noise), policy gradients under DCCDA have equal or higher variance than CTDE, i.e, $Var(DCCDA) \geq Var(CTDE)$.

Our variance analysis motivates us to propose a novel message-dependent \emph{baseline} technique to reduce the variance caused by communication during policy updates. We also observe that the introduction of our variance reduction technique affects the learning performance. To mitigate this effect, we introduce a regularization technique (i.e., a KL divergence term) on decentralized actors, ensuring the experience generated by non-communicating actors aligns with communicating critics and thereby improving the learning of critics. The two proposed techniques can be applied to any Comm-MADRL method under DCCDA. To show the effectiveness and efficiency of our proposed techniques, we extend two existing MADRL methods under the DCCDA setting and evaluate the two methods with and without our techniques on multiple tasks in two benchmark environments, StarCraft Multi-Agent Challenge \cite{Samvelyan2019StarCraft} and Traffic Junction \cite{Singh2019IC3Net}. The results show that our proposed techniques can reduce the variance in policy gradients caused by communication under DCCDA and significantly improve learning performance.

\section{Related Works}
\label{sec:relatedWorks}

\paragraph{\textnormal{\textbf{Learning Communication in MADRL}}.} Previous works mainly focus on learning efficient and effective communication to improve task-level performance under either CTDE with communication setting (where communication is used in both training and execution) \cite{Wang2020NDQ,Das2019TarMAC}, or DTDE with communication setting \cite{Iqbal2019MAAC,Liu2020G2ANet}. In CTDE with communication, existing research works utilize either a shared Q-function (e.g., DIAL \cite{Foerster2016Comm}), centralized but factorized Q-functions (e.g., NDQ \cite{Wang2020NDQ}, TMC \cite{Zhang2020TMC}, MAIC \cite{Yuan2022MAIC}, MASIA \cite{Guan2023Aggregation}, and T2MAC \cite{Sun2024T2MAC}), or a joint value function (e.g., TarMAC \cite{Das2019TarMAC}, GAMCL \cite{Mao2020GatedACML}, I2C \cite{Ding2020I2C}, and TEM \cite{Guo2023Email}) to enable efficient training of communication architectures. Compared to CTDE with communication, communication in the DTDE setting is under-explored. The existing works mainly rely on actor-critic methods \cite{Jiang2018ATOC,Liu2020G2ANet,Niu2021MAGIC,Chen2024RGMComm}. When communication is allowed between individual critics not actors (i.e., the DCCDA setting), learning communication relies on MAAC \cite{Iqbal2019MAAC} and its variant GAAC \cite{Liu2020G2ANet}. In MAAC, individual agents aggregate and integrate encoded information from other agents into local Q-functions. GAAC proposes to incorporate graph neural networks in the critic (Q) network to effectively combine information from neighboring agents. In addition to DCCDA methods, there are specific DTDE methods that allow communication among individual actors. Concretely, ATOC \cite{Jiang2018ATOC} enables communication in actors and uses non-communicating critics for training. MAGIC \cite{Niu2021MAGIC} allows each agent's critic and actor to share the same neural network components, including the communication architecture. IMAC \cite{Wang2020IMAC} and RGMComm \cite{Chen2024RGMComm} use a setup where each agent integrates encoded messages into its policies and incorporates the actions and observations of all other agents into individual critics, implicitly assuming full observability. In contrast, DCCDA methods do not involve message sharing during execution but learning what or when to communicate local information during training. As a result, DCCDA methods allow decentralized execution without communication and benefit from communication during training.

\paragraph{\textnormal{\textbf{Variance Reduction in MADRL}}.} Variance reduction is an essential topic in MADRL \cite{Tucker2018Baseline,Kuba2021Setting}. Previous works have built a theoretical analysis of the variance in policy gradients without considering learning communication. Lyu et al. \cite{Lyu2021Contrasting,Lyu2023Centralized} theoretically contrast policy gradients under CTDE and DTDE without communication settings and claim that the uncertainty of other agents' observations and actions appeared in centralized Q-functions can increase the variance in policy gradients. One of the most successfully applied and extensively studied methods to reduce variance is known as the \textit{baseline} technique \cite{Wu2018Baseline,Foerster2018COMA,Kuba2021Setting}. Specifically, Wu et al. \cite{Wu2018Baseline} utilizes an action-dependent baseline to consider the influence of different dimensions of actions. Foerster et al. \cite{Foerster2018COMA} introduces a counterfactual baseline that marginalizes out a single agent’s action, while keeping the other agents’ actions fixed. More recently, Kuba et al. \cite{Kuba2021Setting} mathematically analyze the variance of policy gradients under CTDE and quantify how agents contribute to the total variance in Markov games, where states are available to agents. They propose a state-based baseline technique to achieve minimal variance when estimating policy gradients under CTDE. In summary, existing baseline techniques consider the source of variance from the uncertainty in states or actions. In contrast, our baseline technique considers the source of variance from the uncertainty in messages based on partial information. To the best of our knowledge, this is the first work to study variance in policy gradients considering learning communication in decentralized MADRL and partially observable environments, and propose a baseline technique to decrease such variance.

\section{Preliminaries}

\subsection{Multi-Agent Reinforcement Learning}
We consider cooperative multi-agent tasks where a team of agents interacts within the same environment to achieve some common goals. The tasks are generally modeled as decentralized partially observable Markov decision processes (Dec-POMDPs) \cite{OliehoekA16DecPOMDP}. A Dec-POMDP is defined by a tuple $\left\langle\mathcal{I}, \mathcal{S},\rho^0,\left\{\mathcal{A}_{i}\right\}, P, \left\{\mathcal{O}_{i}\right\}, O, \mathcal{R}, \gamma \right\rangle$, where $\mathcal{I}$ is a set of (finite) agents indexed as $\{1,...,N\}$, $\mathcal{S}$ is a set of environment states, $\rho^{0}$ is the initial state distribution, $\mathcal{A}_{i}$ is a set of actions available to agent $i$, and $\mathcal{O}_{i}$ is a set of observations of agent $i$. We denote a joint action space as $\boldsymbol{\mathcal{A}} =\times_{i\in \mathcal{I}}\mathcal{A}_{i}$ and a joint observation space as $\boldsymbol{\mathcal{O}} =\times_{i\in \mathcal{I}}\mathcal{O}_{i}$. Therefore, transition function $P: \mathcal{S} \times \boldsymbol{\mathcal{A}} \rightarrow \Delta(\mathcal{S})$ specifies the transition probability $p(s'|s,\boldsymbol{a})$ from state $s \in \mathcal{S}$ to new state $s' \in \mathcal{S}$ given joint action $\boldsymbol{a} = \langle a_1,...,a_N \rangle$ and $\boldsymbol{a} \in \boldsymbol{\mathcal{A}}$. With the environment transitioning to new state $s'$, given joint action $\boldsymbol{a}$, the probability of a joint observation $\boldsymbol{o}=\langle o_1,...,o_N \rangle$ ($\boldsymbol{o} \in \boldsymbol{\mathcal{O}}$) is determined according to the observation function $O: \mathcal{S} \times \boldsymbol{\mathcal{A}} \rightarrow \Delta(\mathcal{O})$. Each agent then receives a shared reward according to the reward function $\mathcal{R}: \mathcal{S} \times \boldsymbol{\mathcal{A}} \rightarrow \mathbb{R}$. The rewards $r_t=\mathcal{R}(s_t, \boldsymbol{a}_t)$ are discounted by the discount factor $\gamma$ over time step $t$. The joint policy $\boldsymbol{\pi}$ of agents induces an on-policy joint Q-function: $Q^{\boldsymbol{\pi}}(s, \boldsymbol{a})=\mathbb{E}_{s_t \sim P, \boldsymbol{a}_t \sim \boldsymbol{\pi}}[\sum_{t=0}^{T}\gamma^t r_t|s_0=s, \boldsymbol{a}_0=\boldsymbol{a}]$, which is the expected discounted return by applying the joint action $\boldsymbol{a}$ and following the joint policy afterward till the time horizon $T$. Whenever the state $s$ is not observable, we use the joint history $\boldsymbol{h}=\{h_1,...,h_N\}$ instead, where $h_i=(o^i_0,a^i_0,...,o^i_t)$ is the individual observation-action history of agent $i$ up to time step $t$. Therefore, we obtain the history-based joint Q-function $Q^{\boldsymbol{\pi}}(\boldsymbol{h}, \boldsymbol{a})$ \cite{Lyu2021Contrasting}. During implementation, histories are often processed using LSTM neural networks \cite{Omidshafiei2017DEC}, which stack past observations and actions into fixed-size memory cells. For notational readability, we omit the time step $t$.  

\subsection{Policy Gradients under Different Settings}
\label{sec:pgSettings}

\paragraph{\textnormal{\textbf{Policy gradients under CTDE}.}} In various policy gradient methods under CTDE, e.g., MAPPO \cite{Yu2022PPO} and COMA \cite{Foerster2018COMA}, a centralized and joint critic (e.g., a joint Q-function) is used to guide the learning of decentralized actors (policies). Following the setting of Lyu et al. \cite{Lyu2023Centralized}, the CTDE policy gradient of agent $i$ is defined as follows:
$$
g^i_{CTDE} \doteq \mathbb{E}_{\boldsymbol{h},\boldsymbol{a}}[Q^{\boldsymbol{\pi}}(\boldsymbol{h}, \boldsymbol{a})\nabla_{\theta_i}\log \pi_i(a_i|h_i, \theta_i)]
$$
where $Q^{\boldsymbol{\pi}}(\boldsymbol{h}, \boldsymbol{a})$ is the on-policy joint values and $\theta_i$ is the parameters of policy $\pi_i$. We further follow the work of Lyu et al. to use $\hat{g}^i_{CTDE}$ to denote the (single-sample) estimate of $g^i_{CTDE}$, i.e., $\hat{g}^i_{CTDE}=Q^{\boldsymbol{\pi}}(\boldsymbol{h}, \boldsymbol{a})\nabla_{\theta_i}\log \pi_i(a_i|h_i, \theta_i)$.

\paragraph{\textnormal{\textbf{Policy gradients under DTDE}.}} In various policy gradient methods under DTDE, e.g., IPPO \cite{Yu2022PPO}, an individual and local critic is used to guide the learning of decentralized policies. Similarly, by following the setting of Lyu et al. \cite{Lyu2023Centralized}, the DTDE policy gradient of agent $i$ is defined as follows:
$$
g^i_{DTDE} \doteq \mathbb{E}_{\boldsymbol{h},\boldsymbol{a}}[Q^{\pi}_i(h_i, a_i)\nabla_{\theta_i}\log\pi_i(a_i|h_i, \theta_i)] 
$$
where $Q^{\pi}_i(h_i, a_i)$ is the on-policy values of agent $i$. Similarly, $\hat{g}^i_{DTDE}$ is used to denote the (single-sample) estimate of $g^i_{DTDE}$, i.e., $\hat{g}^i_{DTDE}=Q^{\pi}_i(h_i, a_i)\nabla_{\theta_i}\log \pi_i(a_i|h_i, \theta_i)$. 

\paragraph{\textnormal{\textbf{Policy gradients under DCCDA}}.} Similar to policy gradients in CTDE and DTDE, we formulate the policy gradients under DCCDA based on the literature \cite{Iqbal2019MAAC,Liu2020G2ANet}. Essentially, we define messages $m_i$ as being generated from a probabilistic message function based on each agent's \textcolor{black}{history ($h_i$) and actions from the actor ($a_i$)}: $m_i \sim f^{msg}(\cdot|h_i,a_i)$, where message function $f^{msg}$ is typically implemented using neural networks with parameters $\theta^{msg}$. In Comm-MADRL, $\theta^{msg}$ is often learned through backpropagation or RL. By allowing full communication, we define decentralized communicating critics as $Q(h_i, a_i, m_{-i})$, where $m_{-i}=\{m_1,...,m_{i-1},m_{i+1},...,m_{N}\}$ is the received messages of agent $i$ from all the other agents (denoted as $-i$). The DCCDA policy gradient of receiver agent $i$ given by on-policy values is then defined as follows:
$$
g^i_{DCCDA} \doteq \mathbb{E}_{\boldsymbol{h},\boldsymbol{a},\boldsymbol{m}}[Q^{\pi}_i(h_i, a_i, m_{-i})\nabla_{\theta_i}\log \pi_i(a_i|h_i, \theta_i)]
$$
where $Q^{\pi}_i(h_i, a_i, m_{-i})$ is the on-policy Q-values of agent $i$. Then, $\hat{g}^i_{DCCDA}$ is used to denote the (single-sample) estimate of $g^i_{DCCDA}$, i.e., $\hat{g}^i_{DCCDA}=Q^{\pi}_i(h_i, a_i, m_{-i})\nabla_{\theta_i}\log \pi_i(a_i|h_i, \theta_i)$. 

\section{Methods}
\label{sec:methods}

During training in DCCDA, agents communicate a range of values (or a vector of values) as messages rather than the entire local information, which avoids sharing private information and reduces communication overhead. The messages are then integrated into agents' critics, guiding the gradient updates of actors (policies). During execution in DCCDA, agents can discard communicating critics and use actors to make decisions independently and locally. We are interested in how messages affect the policy updates of receiver agents in the training period. We specifically focus on how policy gradients diverge, in terms of the variance measurement. Inspired by previous variance analysis in MADRL with CTDE and DTDE settings \cite{Lyu2021Contrasting,Kuba2021Setting,Lyu2023Centralized}, we conduct variance analysis in DCCDA policy gradients, focusing on the variance induced by communication. In our variance analysis, we consider both idealistic communication and non-idealistic communication settings. In both scenarios, we demonstrate that the variance of DCCDA policy gradients can be equal to or higher than that of CTDE policy gradients. Motivated by the variance analysis, we propose techniques for practical learning of agents, to reduce the potential variance introduced by communication and to improve value learning.

\subsection{Variance Analysis}
\label{sec:varAnalysis}

\paragraph{\textnormal{\textbf{Idealistic Communication Setting.}}} We first consider an idealistic communication setting by assuming the existence of a perfect message decoder. Under such idealistic scenarios, the decentralized communicating critics $Q^\pi_i(h_i, a_i, m_{-i})$ and the centralized critics $Q^\pi(\boldsymbol{h}, \boldsymbol{a})$ can be related as communication induces complete and sound information from all agents. However, the probabilistic nature of messages (as commonly used by MADRL with communication methods) can lead to variance in policy gradient samples. Hence, we come to the following theorem:
\begin{theorem}
The DCCDA sample gradient has a variance greater or equal than that of the
CTDE sample gradient in idealistic communication setting: $Var(\hat{g}^i_{DCCDA}) \geq Var(\hat{g}^i_{CTDE})$.
\label{theo:DCCDAVarIdeal}
\end{theorem}
\textbf{Proof Sketch} (full proof in Appendix \ref{app:variance}). We leverage the Bellman equation to find the equivalence between $Q^{\pi}_i(h_i, a_i, m_{-i})$, used as critics in $\hat{g}^i_{DCCDA}$, and $Q^\pi(\boldsymbol{h}, \boldsymbol{a})$, used as critics in $\hat{g}^i_{CTDE}$. Essentially, as $Q^{\pi}(\boldsymbol{h}, \boldsymbol{a})$ and the expected value of $Q^{\pi}_i(h_i, a_i, m_{-i})$ over messages converge to the unique fixed point, we get: $Q^{\pi}(\boldsymbol{h}, \boldsymbol{a})=\mathbb{E}_{m_{-i}|\boldsymbol{h},\boldsymbol{a}}[Q^{\pi}_i(h_i, a_i, m_{-i})]$. Based on this, we find that $\hat{g}^i_{DCCDA}$ and $\hat{g}^i_{CTDE}$ are equal in expectation such that the difference between $Var(\hat{g}^i_{DCCDA})$ and $Var(\hat{g}^i_{CTDE})$ end ups with an expectation of the square of gradients minus the square of the expectation of gradients. According to Jensen's inequality, we conclude that $Var(\hat{g}^i_{DCCDA})$ is equal to or higher than $Var(\hat{g}^i_{CTDE})$.

\paragraph{\textnormal{\textbf{Non-idealistic Communication Setting.}}} We now consider the variance analysis under a non-idealistic communication setting, where messages received by agent $i$ are corrupted by a noise term $\epsilon_i$. The noise term can come from the imperfection of decoders, e.g., due to the use of neural networks. To simplify the analysis, we lift noise in received messages to Q-values, where $m_{-i}=<h_{-i},a_{-i},\epsilon_i>$ for receiver agent $i$, leading to 
$Q^{\pi}_i(h_i,a_i, m_{-i})=Q^{\pi}_i(h_i,a_i, <h_{-i},a_{-i},\epsilon_i>)=Q^{\pi}_i(\boldsymbol{h}, \boldsymbol{a}, \epsilon_i)$. The individual but centralized Q-function with additive noise, $Q^{\pi}_i(\boldsymbol{h}, \boldsymbol{a}, \epsilon_i)$, is used as the critics of decentralized actors, forming a noise version of DCCDA policy gradients $\hat{g}^i_{DCCDA-noise}$. 
Inspired by Wang et al. \cite{Wang2020Noise}, the noise term can affect the rewards received by each agent, such as flipping the sign in case the reward is binary. We then prove that removing the effect of the noise term can increase variance, resulting in the following theorem:

\begin{theorem}
The noisy version of DCCDA sample gradient has a variance greater or equal than that of the CTDE sample gradient in non-idealistic communication setting: $Var(\hat{g}^i_{DCCDA-noise}) \geq Var(\hat{g}^i_{CTDE})$.
\label{theoremNoiseDCCDAMain}
\end{theorem}
\textbf{Proof Sketch} (full proof in Appendix \ref{app:varianceNoise}). We first relate the noise term with the probability of changes in rewards. Inspired by Wang et al. \cite{Wang2020Noise}, a surrogate reward function can be defined to remove the effect of noise in rewards. Upon the surrogate reward function, we define a surrogate Q-function $\hat{Q}^{\pi}_i(\boldsymbol{h}, \boldsymbol{a}, \epsilon_i)$. By summing up noisy terms $\epsilon_i$, the expected value of $\hat{Q}^{\pi}_i(\boldsymbol{h}, \boldsymbol{a}, \epsilon_i)$ is shown to be equal to $Q^{\pi}(\boldsymbol{h}, \boldsymbol{a})$ (defined on noise-free rewards), i.e., $Q^{\pi}(\boldsymbol{h}, \boldsymbol{a})=\mathbb{E}_{\epsilon_i}[\hat{Q}^{\pi}_i(\boldsymbol{h}, \boldsymbol{a}, \epsilon_i)]$ by induction proof. The equality greatly simplifies the variance analysis between the noise version of DCCDA policy gradients $\hat{g}^i_{DCCDA-noise}$ (using $\hat{Q}^{\pi}_i(\boldsymbol{h}, \boldsymbol{a}, \epsilon_i)$ as critics) and the CTDE policy gradients $\hat{g}^i_{CTDE}$ (using $Q^{\pi}(\boldsymbol{h}, \boldsymbol{a})$ as critics). By comparing the variance of the two gradients, we achieve that $Var(\hat{g}^i_{DCCDA-noise}) \geq Var(\hat{g}^i_{CTDE})$.

\subsection{The Baseline Technique and Regularized Policies}
\label{sec:proposedMe}

Inspired by our variance analysis, communication in DCCDA policy gradients can introduce variance in both the idealistic and non-idealistic communication settings. To mitigate the variance caused by communication, we propose a novel message-dependent baseline to reduce the variance in DCCDA policy gradients. As variance reduction may affect the learning performance of MADRL algorithms \cite{Kuba2021Setting,Wu2018Baseline}, we also investigate how to enhance the learning of decentralized communicating critics and decentralized actors in DCCDA. Specifically, the communicating critics, which are used to compute the values of the message-dependent baseline, implicitly suggest that experience is generated by policies with communication. However, in DCCDA, decentralized policies ($\pi_i(a_i|h_i, \theta_i)$) do not use communication, generating experiences that can mislead the training of communicating critics. Thus, using decentralized policies for the learning of communicating critics can be problematic. To resolve the issue, we further propose a KL divergence term to regularize policies for enhancing the learning of critics. The message-dependent baseline technique and the KL divergence term jointly form our modular techniques, which will be integrated into existing communication methods under the DCCDA setting.

We introduce a novel message-dependent baseline $b_i(h_i, m_{-i})$ to achieve minimal variance with the presence of communication. For the learning of critics, we use Q-function $Q_i(h_i, a_i, m_{-i})$ to describe the samples of the cumulative discounted return of agent $i$ with communication, where message $m_{-i}$ can be either noisy or noise-free. We assume that $Q_i(h_i, a_i, m_{-i})$ can converge to the true on-policy values $Q^{\pi}_i(h_i, a_i, m_{-i})$. Based on the definitions, we write out DCCDA policy gradients with the message-dependent baseline (denoted as DCCDA-OB) as follows:
\begin{equation*}
\begin{split}
g^i_{DCCDA-OB} &= \mathbb{E}_{\boldsymbol{h},\boldsymbol{a},\boldsymbol{m}}[(Q_i(h_i, a_i, m_{-i}) - b_i(h_i, m_{-i})) \\
&\nabla_{\theta_i}\log \pi_i(a_i|h_i, \theta_i)]
\end{split}
\end{equation*}
where actions are sampled from decentralized policies $\pi_i(\cdot|h_i)$ in practice. We then use $\hat{g}^i_{DCCDA-OB}$ to denote the (single-sample) estimate of $g^i_{DCCDA-OB}$, i.e., 
$\hat{g}^i_{DCCDA-OB}=(Q_i(h_i, a_i, m_{-i}) - b_i(h_i, m_{-i})) \nabla_{\theta_i}\log \pi_i(a_i|h_i, \theta_i)$. We seek the optimal message-dependent baseline $b^*_i(h_i, m_{-i})$ to minimize the variance $Var(\hat{g}^i_{DCCDA-OB})$ of the policy gradient estimate $\hat{g}^i_{DCCDA-OB}$. Therefore, we come to the following theorem:

\begin{theorem}
The optimal message-dependent baseline for DCCDA-OB gradient estimator is:
\begin{equation}
\begin{split}
b_i^*(h_i, m_{-i}) = \frac{\mathbb{E}_{a_i}[Q_i(h_i, a_i, m_{-i}) S]}{\mathbb{E}_{a_i}[S]}
\end{split}
\label{eq:obFinalMain}
\end{equation}
where $S=\nabla_{\theta_i}\log \pi_i(a_i|h_i, \theta_i)^T\nabla_{\theta_i}\log \pi_i(a_i|h_i, \theta_i)$.
\label{theo:DCCDAOB}
\end{theorem}
\textbf{Proof Sketch} (For the full proof see Appendix \ref{app:baseline}). The key idea is to determine an optimal baseline to minimize the variance of $Var(g^i_{DCCDA-OB})$ by analyzing the derivatives of the variance w.r.t. the baseline. In Equation \ref{eq:obFinalMain}, $S$ is the inner product of the gradient $\nabla_{\theta_i}\log \pi_i(a_i|h_i, \theta_i)$, indicating the magnitude of the gradient vector.

The resulting formula $b_i^*(h_i, m_{-i})$ aligns with previous works on baseline techniques \cite{Kuba2021Setting,Wu2018Baseline}, while we incorporate partially observable information (histories) and communication (messages) into the Q-function. Based on Theorem 
\ref{theo:DCCDAOB}, we have:
\setcounter{theorem}{0}
\begin{corollary}
The variance of DCCDA policy gradients is reduced with the optimal message-dependent baseline: $Var(\hat{g}^i_{DCCDA-OB}) \leq Var(\hat{g}^i_{DCCDA})$.
\end{corollary}

\textbf{Proof Sketch} (full proof in Appendix \ref{app:reducedVar}). The key idea is to integrate the optimal baseline $b_i^*(h_i, m_{-i})$ into the variance $Var(\hat{g}^i_{DCCDA-OB})$, which ends up with $Var(\hat{g}^i_{DCCDA})$ minus a non-negative term. Therefore, the variance with the baseline is less than or equal to the variance without the baseline.

Corollary \ref{corollaryOne} is formulated with respect to idealistic communication setting, i.e., $Var(\hat{g}^i_{DCCDA-OB}) \leq Var(\hat{g}^i_{DCCDA})$, but it holds also for non-idealistic communication setting, i.e., $Var(\hat{g}^i_{DCCDA-OB}) \leq Var(\hat{g}^i_{DCCDA-noise})$. 

To enhance the learning of critic $Q_i(h_i, \cdot, m_{-i})$ for each agent, we improve consistency between the critic (which uses communication during training) and the actor (which executes $\pi_i(\cdot|h_i, \theta_i)$ without communication). Specifically, we propose aligning the execution policy $\pi_i(\cdot|h_i, \theta_i)$ with the policy suggested by the critic. Since only the policy $\pi_i(\cdot|h_i, \theta_i)$ is available in implementations, we estimate the policy suggested by the critic using the Boltzmann softmax distribution of local Q-values \cite{BianchiG2017Boltzmann}. This results in the following KL divergence term to regularize $\pi_i(\cdot|h_i, \theta_i)$ for receiver agent $i$:
\begin{equation}
    \mathcal{L}_{KL}(\theta_i) = - D_{KL}(\pi_i(\cdot|h_i, \theta_i) || \mathrm{SoftMax} (\frac{1}{\alpha}Q_i(h_i, \cdot, m_{-i}))
\label{eq:KLObject}
\end{equation}
where $\alpha$ is a temperature parameter and the KL divergence term $\mathcal{L}_{KL}(\theta_i)$ minimizes the KL divergence between the execution policy $\pi_i(\cdot|h_i, \theta_i)$ and the policy suggested by the critics (which we desire to achieve but not modeled during execution). A higher temperature results in a more uniform policy distribution regarding the Q-values, while a lower temperature results in a more greedy policy distribution regarding the Q-values. The KL divergence term penalizes policy $\pi_i(\cdot|h_i, \theta_i)$ when it assigns a high probability to actions that the estimated policy with communication assigns a low probability to, helping to align decentralized policies with the desired behavior.


The optimal message-dependent baseline (OB) and the KL divergence term (KL) jointly constitute our proposed techniques regarding the variance reduction in policy gradients and the learning of critics. The final gradient for agent $i$ is:
\begin{equation}
    g^{i}_{DCCDA-OB-KL} = g^i_{DCCDA-OB} + \beta \nabla_{\theta_i} \mathcal{L}_{KL}(\theta_i)
\label{eq:DCCDA-OB-KL}
\end{equation}
where $\beta$ is the scaling factor. $g^{i}_{DCCDA-OB-KL}$ is used to update the policy parameter $\theta_i$ for each agent. In practice, the expectations in baseline (Equation \ref{eq:obFinalMain}) and the KL term (Equation \ref{eq:KLObject}) are computed using samples of experience (i.e., mini-batches from the experience buffer) to estimate. Concretely, we store the observation-action-message-reward tuples together with Q-values and policy distribution for each agent in the buffer. Then, we compute the inner product of the policy gradient ($S$) based on an analytical form of the softmax policy \cite{Kuba2021Setting} and the baseline value with sampled Q-values. Moreover, we estimate the KL divergence through Q-functions and policies with sampled observation-action-message tuples. The algorithmic procedures and computations are in Appendix \ref{app:extAlg}.


\begin{figure}[t]
\centering 
{\includegraphics[width=0.8\textwidth]{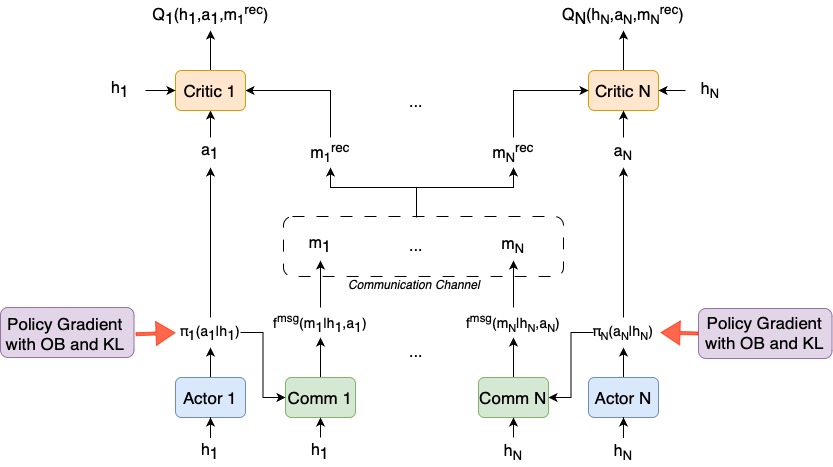}} 
\caption{{DCCDA methods integrated with OB and KL.}}
\label{fig:diagramDCCDA}
\end{figure}

\paragraph{\textnormal{\textbf{Schematic Diagram of Integrating OB and KL}}.} To illustrate how our proposed OB and KL techniques integrate with DCCDA methods, we present a schematic diagram (Figure \ref{fig:diagramDCCDA}) that includes actors, critics, communication modules, a communication channel, and training objectives. In this diagram, the actor, communication (comm), and critic modules are neural networks (e.g., MLPs or RNNs), determined by DCCDA methods. Each agent $i$ decides its action $a_i$ based on its individual history $h_i$. Then, each agent communicates messages $m_i$ sampled from $f^{msg}(m_i|h_i, a_i)$ (see details in Section \ref{sec:pgSettings}). Based on received messages $m_i^{rec}$, agents estimate their Q-values $Q_i(h_i, a_i, m_i^{rec})$. Different DCCDA methods may use different communication strategies, which will be considered in the communication channel to generate the received messages, $m_i^{rec}$.  For instance, in a broadcast communication channel, $m_i^{rec}$ is simply all other agents' messages $m_{-i}$. Red arrows highlight the individual training process for each agent's actor, which incorporates our proposed OB and KL techniques (see Equation \ref{eq:DCCDA-OB-KL}). 



\paragraph{\textnormal{\textbf{Model-agnostic techniques}}.} Our proposed techniques focus on the learning of critics and actors. The design of the communication process, including determining message content and how to exchange communicated messages, can be implemented and covered by any Comm-MADRL method under DCCDA. This reflects the adaptability and flexibility of our techniques, making our techniques model-agnostic to existing Comm-MADRL methods under DCCDA. In this paper, we show two cases of how to extend and adapt communication methods using our proposed techniques, resulting in two extended algorithms (the details of the algorithms can be found in Appendix \ref{app:extAlg}). In the next section, we demonstrate that the two methods using our techniques significantly enhance learning performance and achieve a stable learning process.



\section{Experiments}
\label{lab:experiments}

We evaluate our proposed techniques in two well-established and challenging multi-agent environments, SMAC \cite{Samvelyan2019StarCraft} and Traffic Junction \cite{Das2019TarMAC}.\footnote{The source code is not provided during the double-blind review but will be made publicly accessible upon the acceptance.} Both environments consist of a varying number of cooperative agents with shared rewards and show difficulties in coordinating agents' behaviors. We compare with the following methods: 

\begin{itemize}[leftmargin=*]
    \item \textbf{CTDE methods}: COMA \cite{Foerster2018COMA}, MAPPO \cite{Yu2022PPO}, and MAT \cite{Wen2022MAT} serve as strong baselines in MADRL. Notably, MAPPO and MAT have achieved SOTA performance across several MARL benchmarks \cite{Yu2022PPO,Wen2022MAT}. We compare CTDE methods with DCCDA methods incorporating our proposed techniques to demonstrate that variance can be reduced without compromising learning performance.
    \item \textbf{DCCDA methods}: GAAC \cite{Liu2020G2ANet} and IPPO \cite{Yu2022PPO} extended with communication (IPPO-Comm). As mentioned in related work, GAAC is the state-of-the-art method under the DCCDA setting. Due to the scarcity of communication methods under DCCDA, we adapt the state-of-the-art method under the DTDE setting, IPPO, with a communication architecture, named IPPO-Comm. Details regarding the communication structure used in IPPO-Comm can be found in Appendix \ref{app:extAlg}. 
    \item \textbf{DCCDA methods with OB-KL}: GAAC-OB-KL and IPPO-Comm-OB-KL. We extend DCCDA methods, GAAC and IPPO-Comm, with our proposed techniques (OB-KL), forming GAAC-OB-KL and IPPO-Comm-OB-KL. Notably, GAAC-OB-KL and IPPO-Comm-OB-KL not only demonstrate the effectiveness of our techniques but also highlight the possibility of integrating these techniques with various communication constraints and information structures (e.g., using communication graphs). 
\end{itemize}


We illustrate the essential components of all methods and how they differ from each other in MADRL settings, critics, and policy regularization techniques in Appendix \ref{app:compMethods}. Notably, IPPO-Comm-OB-KL and GAAC-OB-KL are the only methods that incorporate encoded messages and partial information within baseline techniques, and also employ a regularization term to align actors and critics. For evaluation, all results are reported as the median win rate with a 95\% confidence interval. The statistical reports are provided in Appendix \ref{app:statistical}. Details regarding the choice of hyperparameters, such as learning rates, are presented in Appendix \ref{app:parmChoices}. The optimization strategies for all methods are either empirical or fine-tuned to ensure fair comparisons. Additionally, our introduced parameters, the temperature $\alpha$ and the scaling factor $\beta$, are fine-tuned using a grid search, with the parameter search results presented in Appendix \ref{app:addRes}.

\begin{figure}[t]
\centering 
\includegraphics[width=0.33\textwidth]{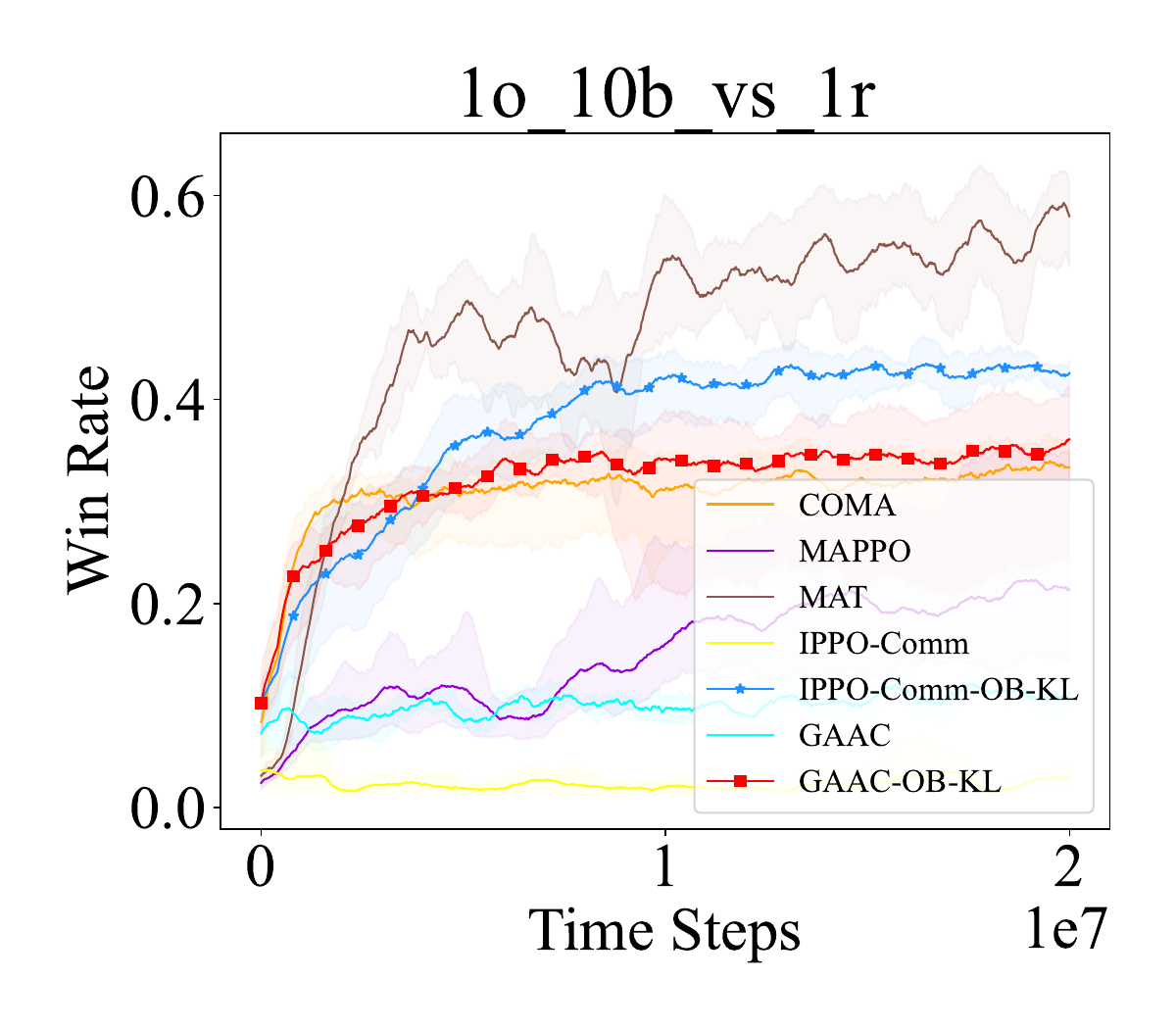} 
\includegraphics[width=0.33\textwidth]{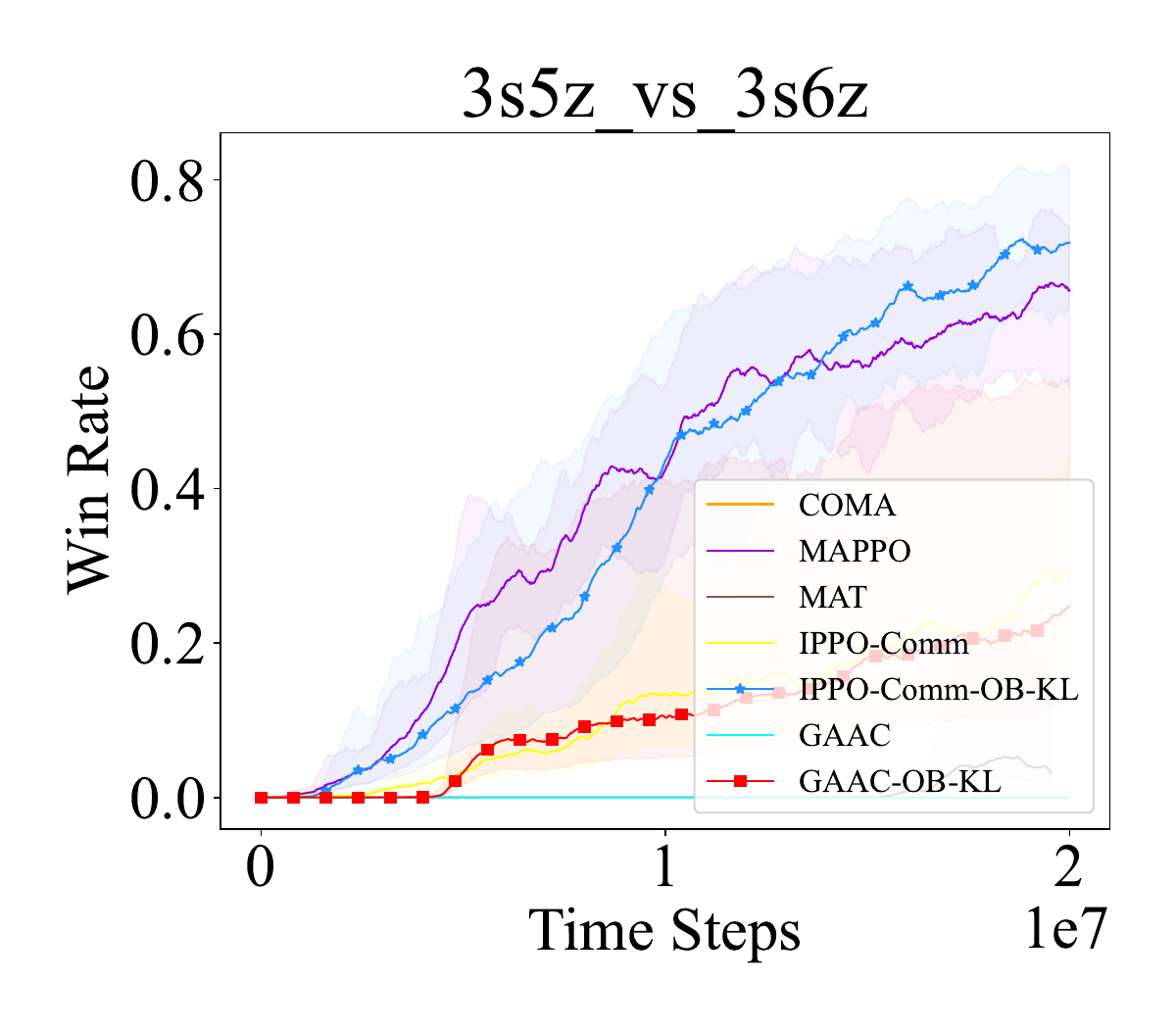}
\includegraphics[width=0.33\textwidth]{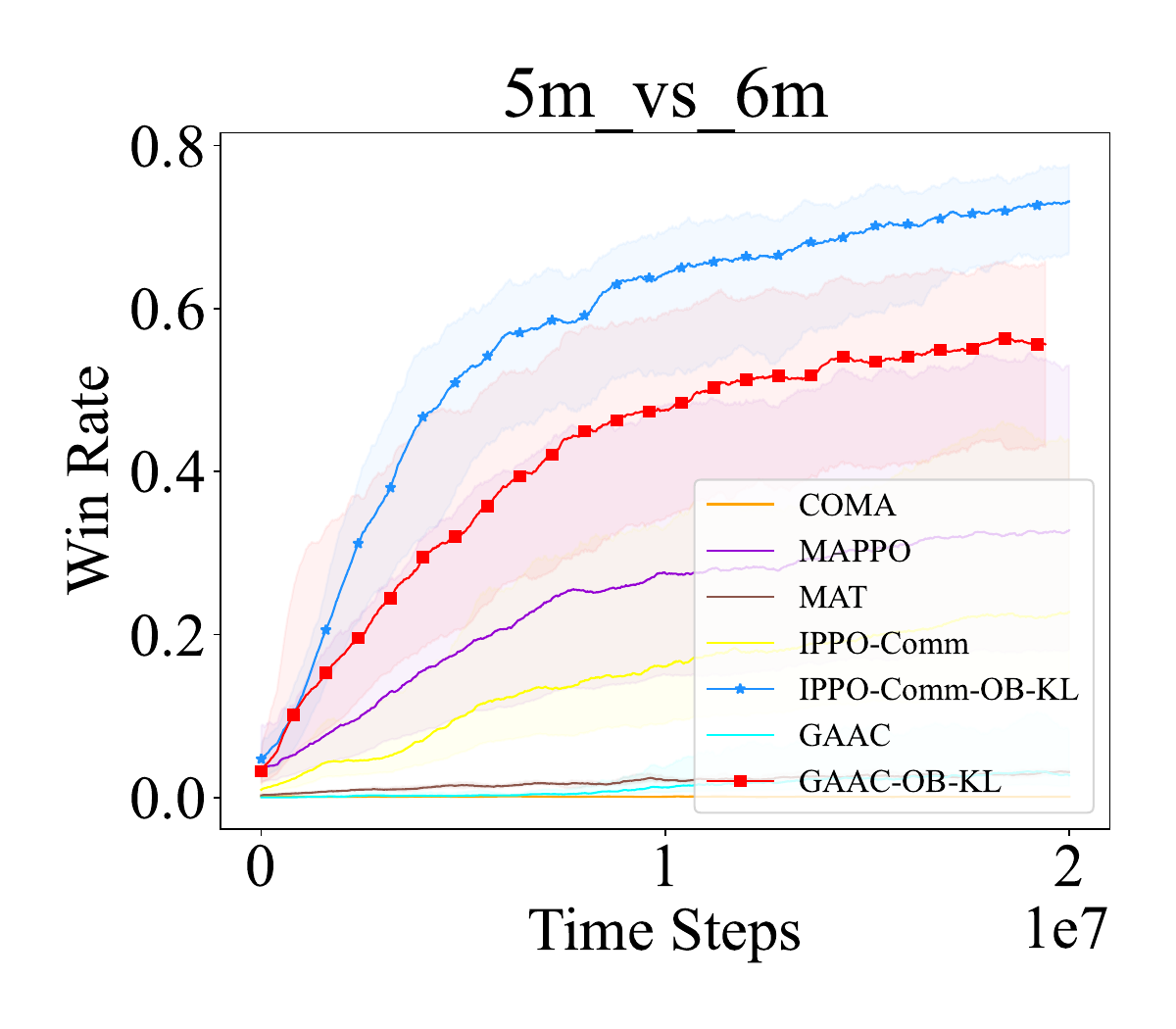} \\
\includegraphics[width=0.33\textwidth]{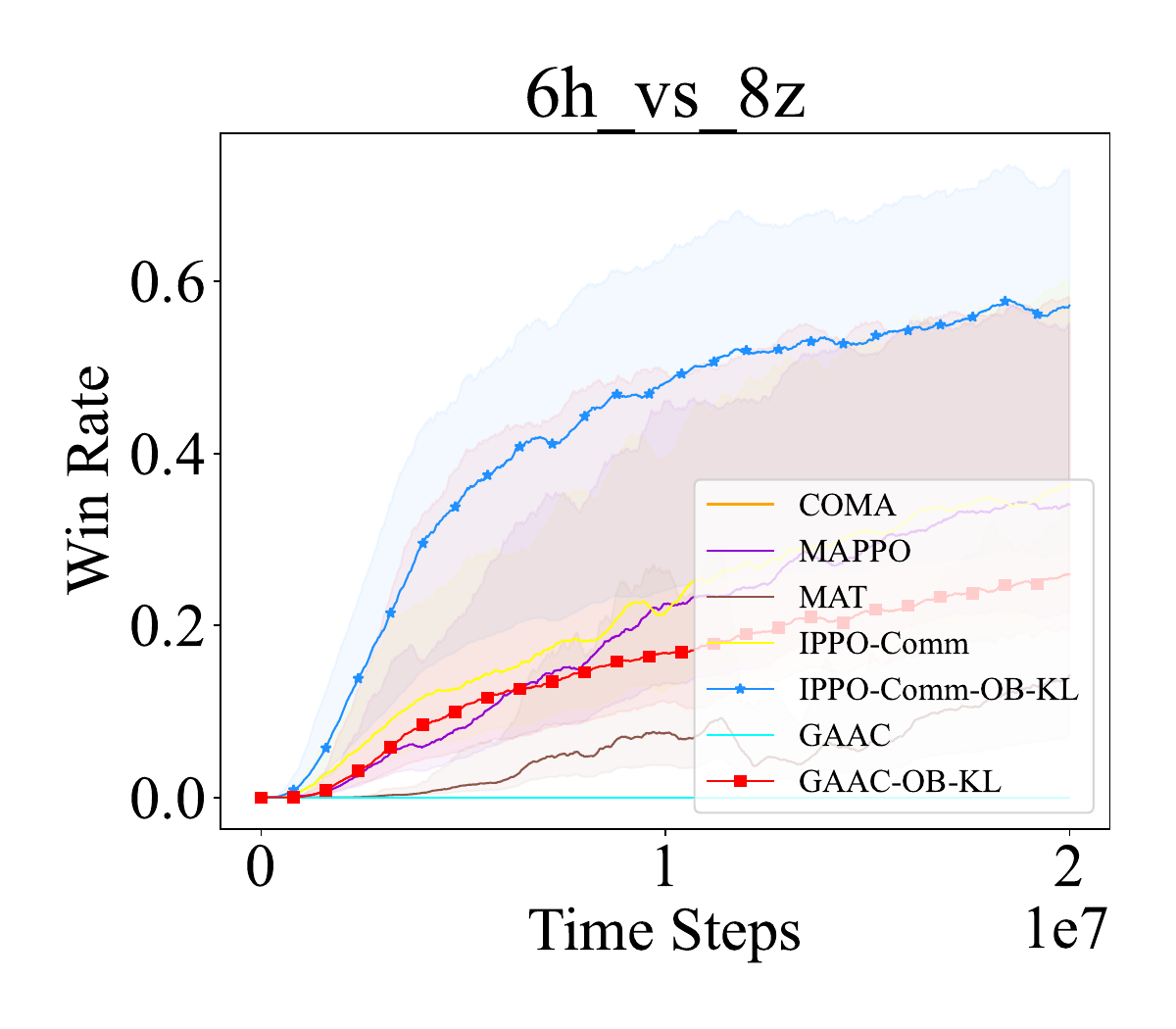}
\includegraphics[width=0.33\textwidth]{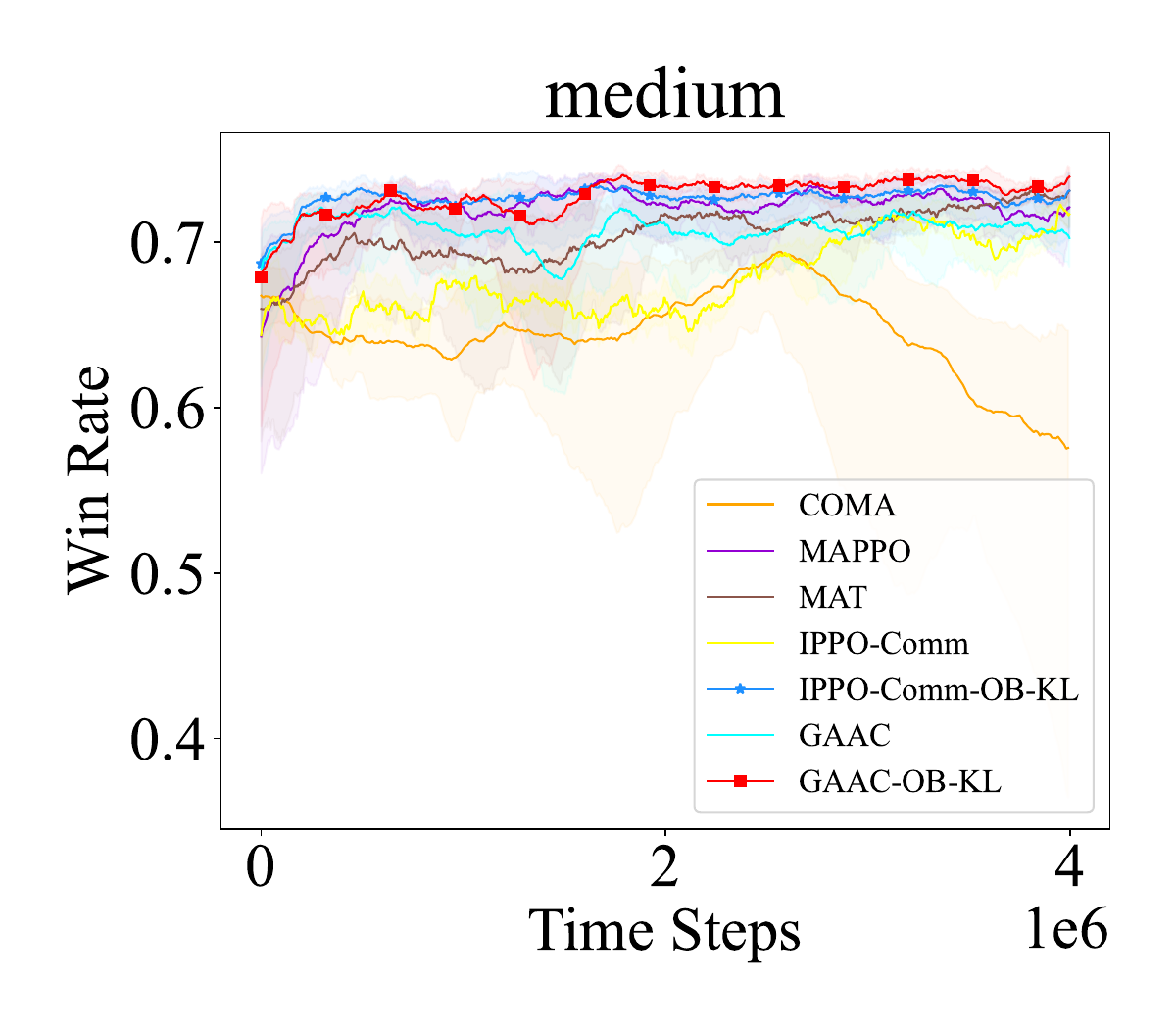}
\includegraphics[width=0.33\textwidth]{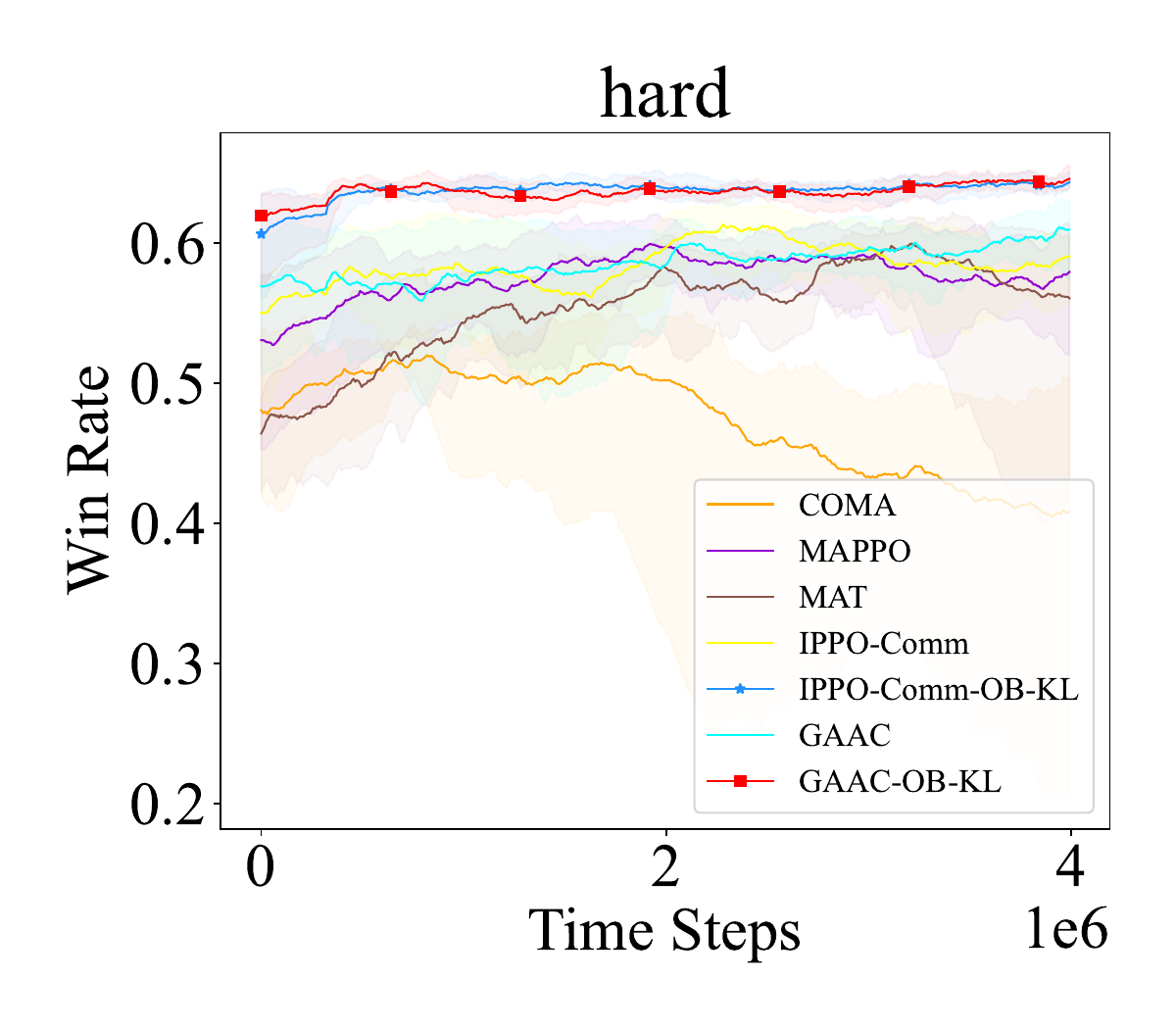}
\caption{{Averaged win rate of all methods.}}
\label{fig:allmaps}
\end{figure}

\subsection{Evaluation Results}
\label{sec:evalRes}

\paragraph{\textnormal{\textbf{Starcraft Multi-Agent Challenge (SMAC)}}.} SMAC is a real-time strategy game serving as a benchmark in the MADRL community. In SMAC, $N$ units controlled by the learned algorithm try to kill all the enemies, demanding proper cooperation strategies and micro-control of movement and attack. We choose several maps where communication plays an important role in broadening agent's view of the environment or coordinating their strategies to effectively attack the enemies: 1o\_10b\_vs\_1r, 3s5z\_vs\_3s6z, 5m\_vs\_6m, and 6h\_vs\_8z. These maps feature a diverse range of agents, roles, and terrains, each presenting different communication requirements. For example, map 1o\_10b\_vs\_1r involves agents with distinct roles, where an overseer that detects the enemy needs to communicate 10 baneling agents who act to kill the enemy. The variety and richness of the tested maps (tasks) can provide a comprehensive evaluation of different aspects of MADRL. In experiments, all methods are evaluated within 20M time steps. Each episode consists of a maximum of 100 time steps. For each seed, we compute the win rate over 32 evaluation episodes after each 25 training episodes.

The results are shown in Figure \ref{fig:allmaps}. IPPO-Comm-OB-KL and GAAC-OB-KL outperform IPPO-Comm and GAAC across all four maps. Specifically, the win rate of IPPO-Comm-OB-KL is significantly higher than all the other methods on 3 out of 4 maps. In map 1o\_10b\_vs\_1r, where agents with different roles need to coordinate, the CTDE method MAT outperforms other methods. Meanwhile, in map 3s5z\_vs\_3s6z, where agents also have different roles, the CTDE method MAPPO outperforms other methods up to approximately 15 million steps but is ultimately surpassed by IPPO-Comm-OB-KL. Despite this, in the remaining maps, where agents do not have different roles, IPPO-Comm-OB-KL surpasses all other methods. These results indicate that CTDE methods tend to perform better in scenarios where agents with different roles/skills need to collaborate, potentially mitigating the effects of high variance in policy gradients. On the other hand, IPPO-Comm-OB-KL and GAAC-OB-KL can consistently improve the learning performance compared to IPPO-Comm and GAAC.

\paragraph{\textnormal{\textbf{Traffic Junction}}.} Traffic Junction is a popular benchmark used to test communication ability, where many cars move along two-way roads with one or more road junctions following predefined routes. We test on the medium and hard maps, using the same setup as \cite{Das2019TarMAC}. We evaluate the success rate within 4M time steps. Each episode consists of a maximum of 40 time steps in the medium map and 80 time steps in the hard map. For each 25 training episodes, we compute the win rate over 32 evaluation episodes. We regard an episode as successful if no collision happens during this episode. The results are shown in Figure \ref{fig:allmaps}. GAAC-OB-KL achieves a higher win rate compared to all the other methods. IPPO-Comm-OB-KL has a similar win rate as MAT in the medium map while surpassing other methods in the hard map. 

\subsection{The Analysis of Variance in Policy Gradients}

We analyze whether the proposed techniques can reduce the variance in gradient updates of policies. We compare the standard deviation of gradient norms across 8 seeds in Table \ref{tab:graNorm}. The standard deviation is important to reflect the variance of policy gradient across independent runs. The lower values the better. As we can see, CTDE methods (COMA, MAPPO, and MAT) have a high variance of gradient norms in most maps. The variance of DCCDA methods without our proposed techniques (IPPO-Comm and GAAC) depends on the underlying optimization strategies. In our implementation, we utilize the same optimization strategy for IPPO-Comm and GAAC as MAPPO to enhance on-policy learning performance, while the variance of IPPO-Comm and GAAC is higher than MAPPO in most maps. Nevertheless, IPPO-Comm-OB-KL and GAAC-OB-KL decrease the variance of gradient norms in all maps compared to IPPO-Comm and GAAC. Compared to CTDE methods, IPPO-Comm-OB-KL achieves a much lower variance in gradient norms across all maps, and GAAC-OB-KL shows lower variance in 5 out of 6 maps. Specifically, in map 3s5z\_vs\_3s6z, GAAC-OB-KL has a slightly higher variance than MAPPO, which may stem from the higher variance of GAAC (compared to MAPPO). We also plot the changes of variance in gradient norms across training steps in Appendix \ref{app:addRes}, where IPPO-Comm-OB-KL and GAAC-OB-KL consistently demonstrate low variance in gradients throughout the learning process. The analysis of gradient norms shows that our proposed techniques can lead to not only performance improvements but also less variance in policy gradients during training.

\subsection{Ablation Studies}
\label{sec:ablStudies}

\begin{figure}[t]
\centering 
    \includegraphics[width=0.33\textwidth]{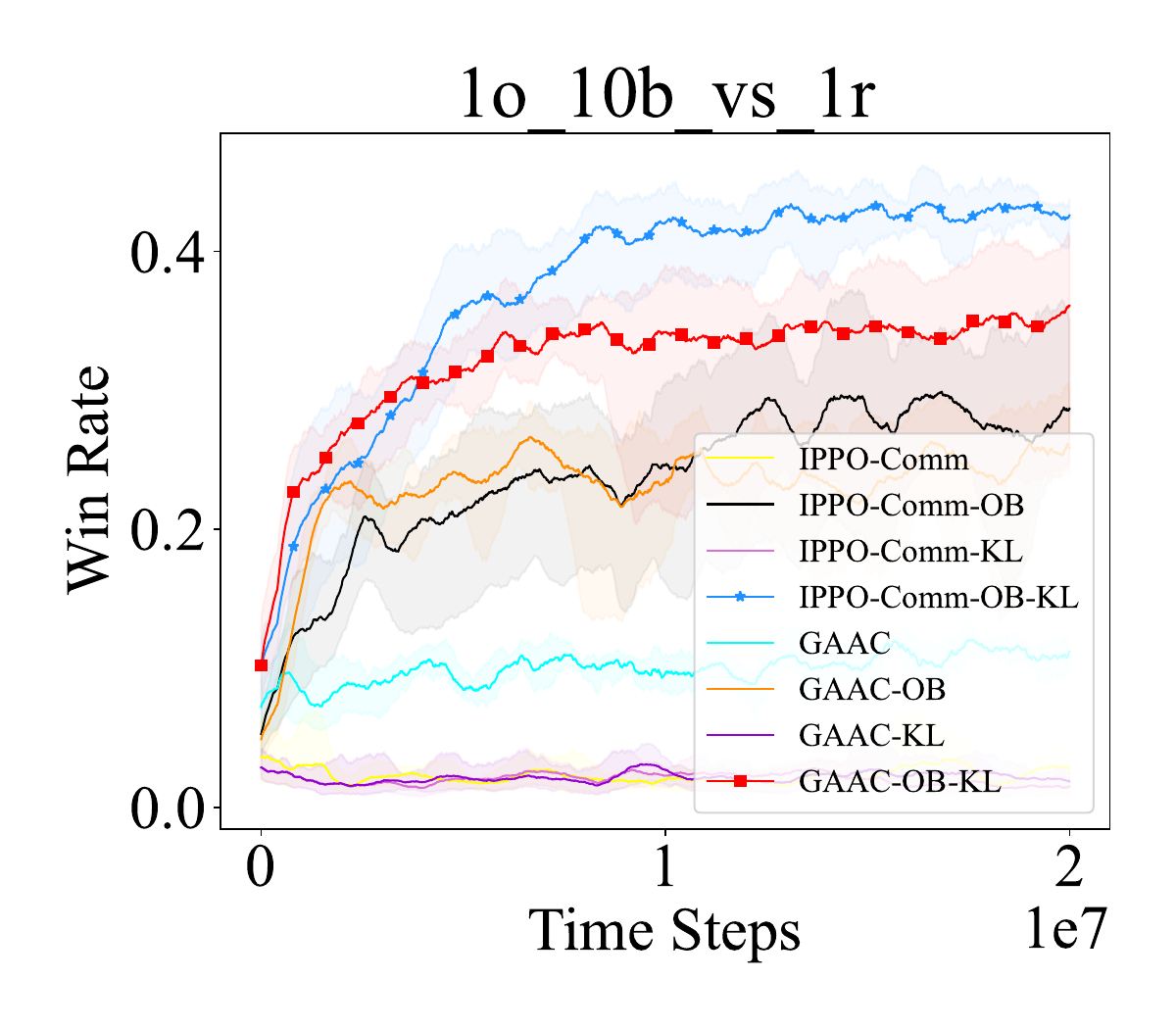} 
    \includegraphics[width=0.33\textwidth]{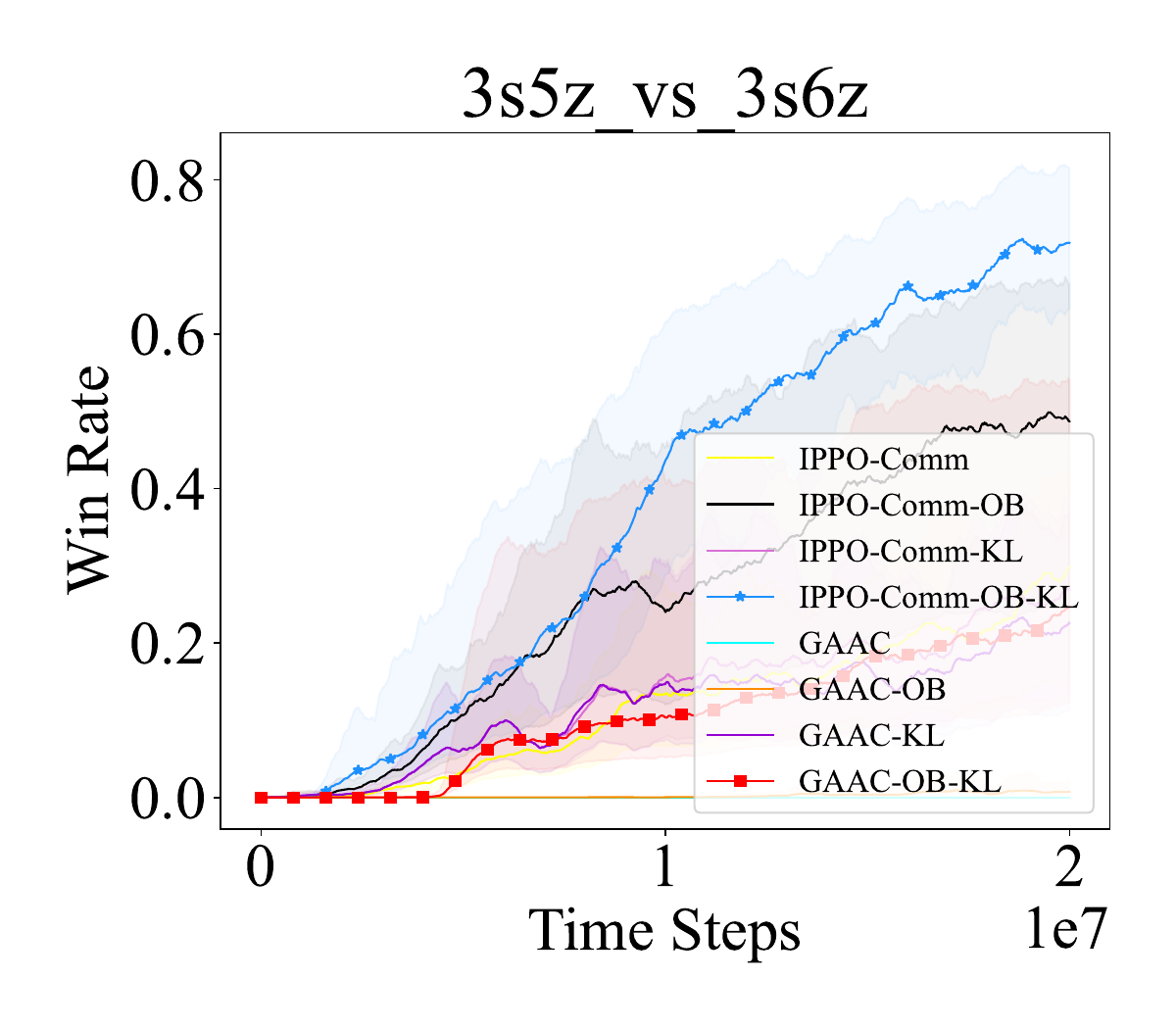}
    \includegraphics[width=0.33\textwidth]{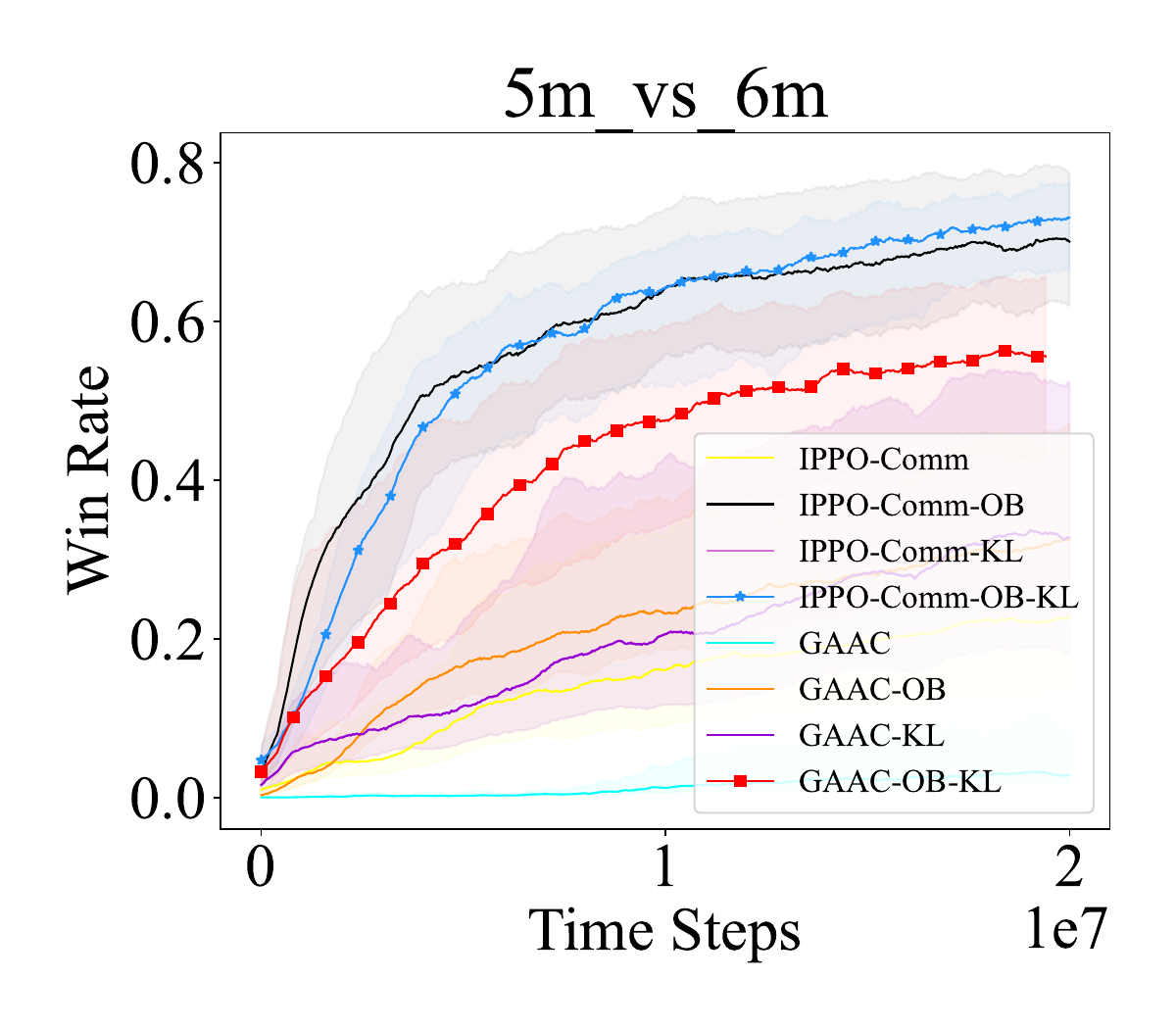}  \\
    \includegraphics[width=0.33\textwidth]{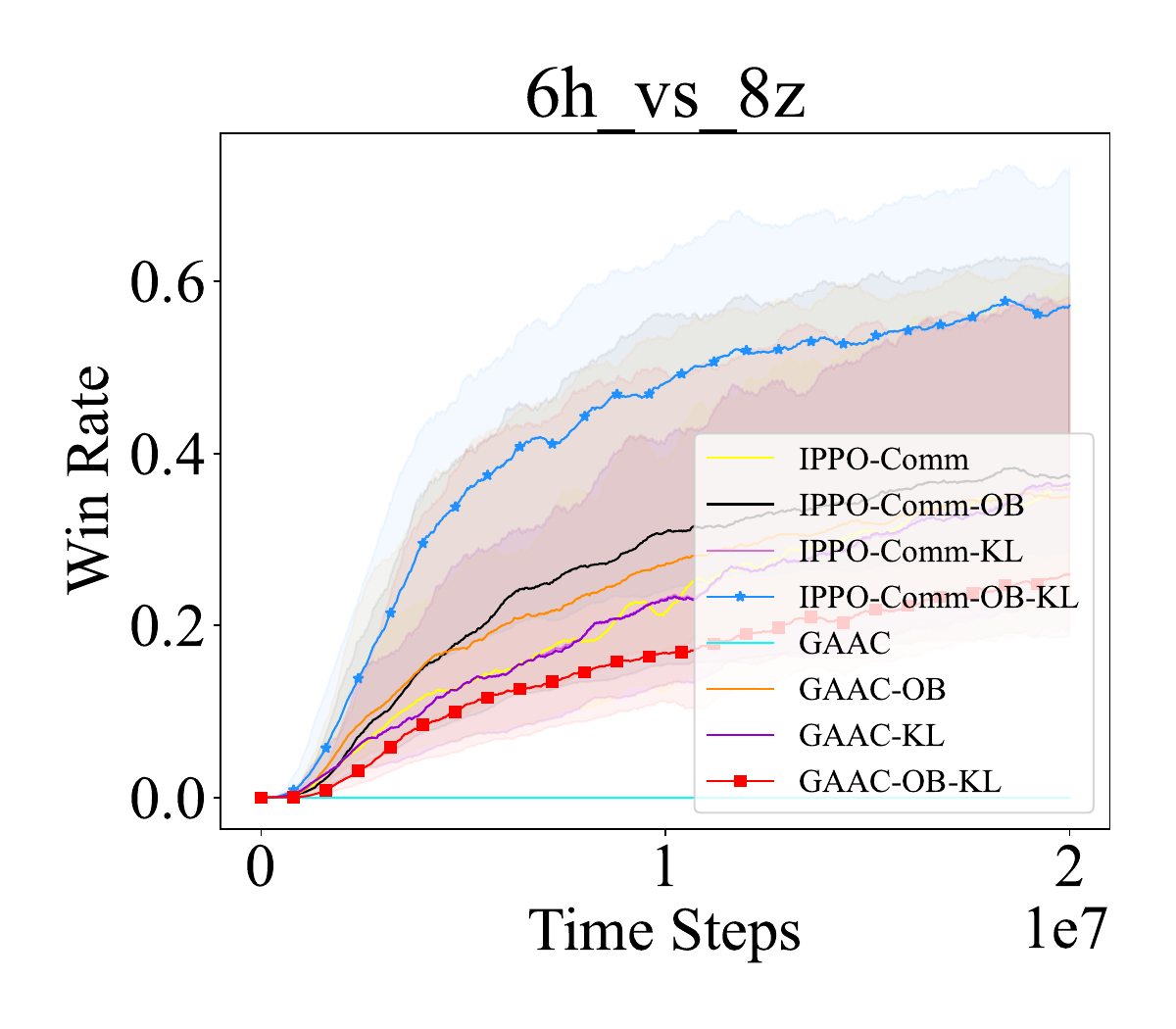} 
    \includegraphics[width=0.33\textwidth]{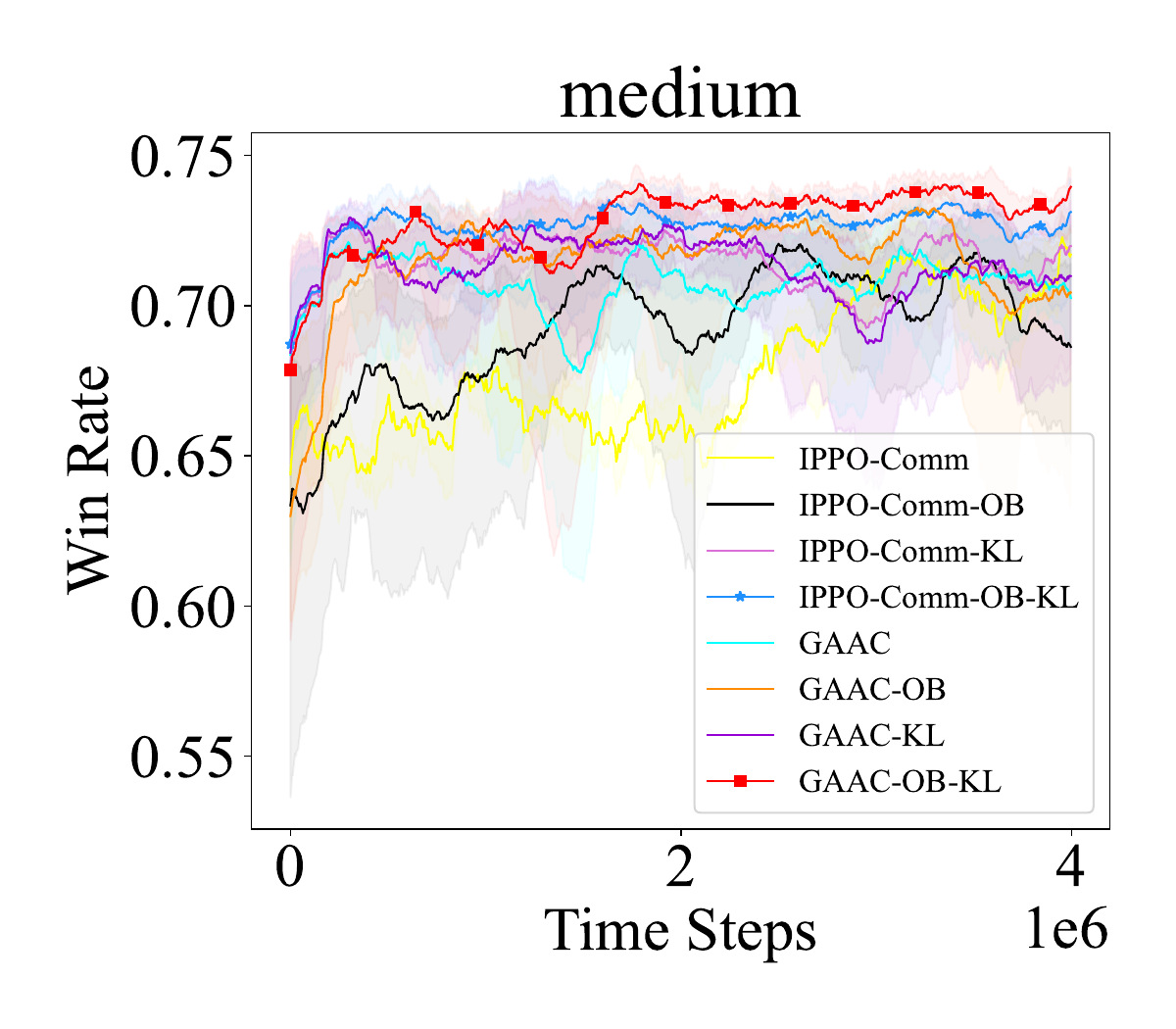} 
    \includegraphics[width=0.33\textwidth]{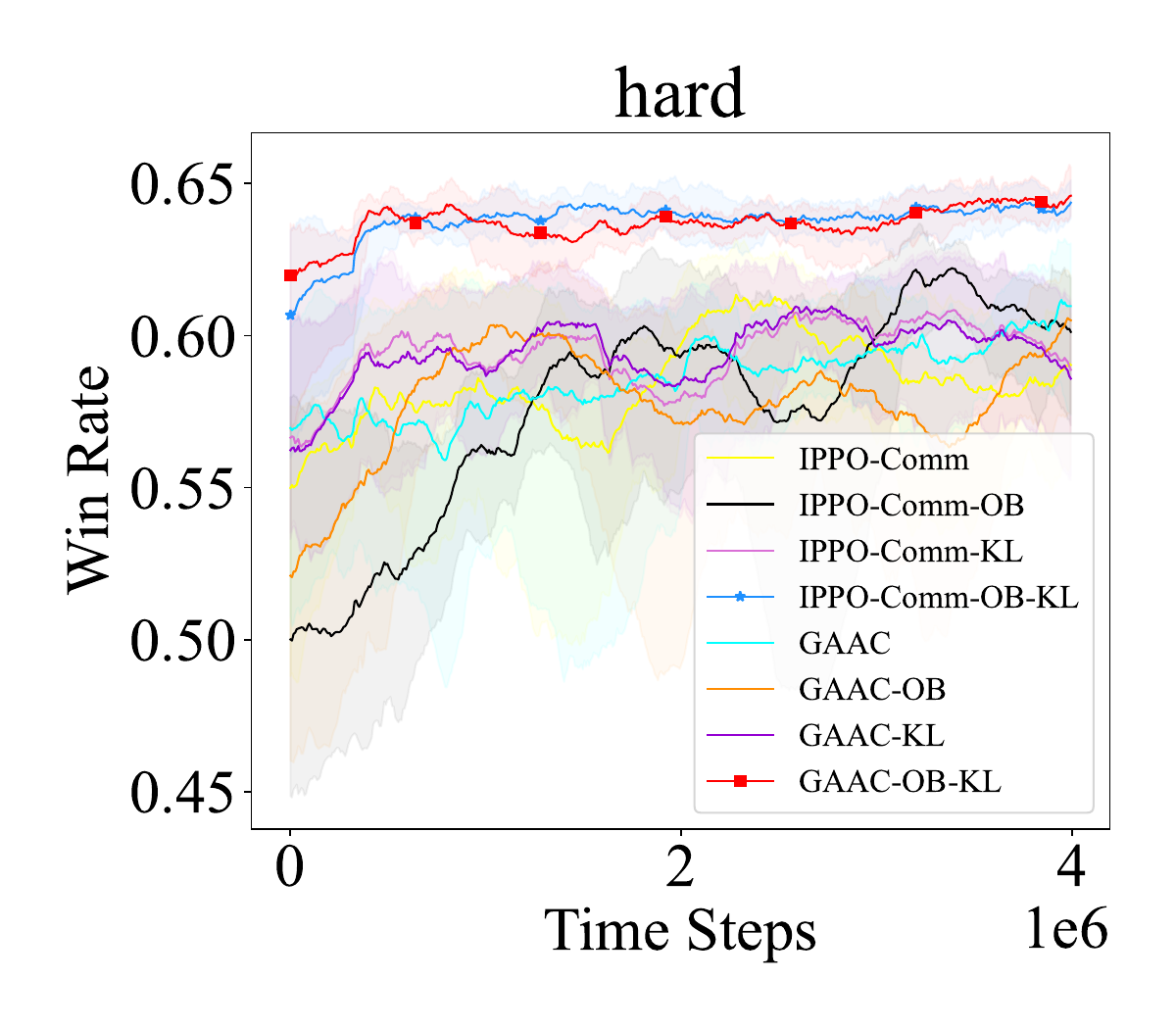} 
\caption{{Averaged win rate when ablating OB and KL.}}
\label{fig:scmapsAbl}
\end{figure}

We further conduct experiments to investigate how the proposed techniques affect learning performance. We ablate the KL divergence term and the baseline (OB) from the full version of IPPO-Comm-OB-KL and GAAC-OB-KL and report the results based on the same settings in SMAC and Traffic Junction. As shown in Figure \ref{fig:scmapsAbl}, IPPO-Comm-OB-KL surpasses IPPO-Comm-OB (without the KL term) and IPPO-Comm-KL (without the baseline) in all maps, which confirms the advantage of jointly applying the baseline and policy regularization to IPPO-Comm. On the other hand, GAAC-OB-KL outperforms GAAC-OB and GAAC-KL in all maps except for map 6h\_vs\_8z, where GAAC-OB has a higher win rate than GAAC-OB-KL. In this map, agents may not be able to correctly estimate the policy with communication based on the learned Q-values, leading to significant drops using the KL divergence term. In all the other maps, removing any one of OB and KL results in a decrease in the performance of IPPO-Comm-OB-KL and GAAC-OB-KL.




\begin{table}[t]
    \caption{The standard deviation of gradient norms, provided on a scale of 0.01. We mark the best (lowest) value in each column.}
        \label{tab:graNorm}
\begin{center}
\begin{sc}
\begin{tabular}{p{0.2\linewidth}p{0.13\linewidth}p{0.13\linewidth}p{0.1\linewidth}p{0.08\linewidth}p{0.08\linewidth}p{0.06\linewidth}}
        \toprule
          & 1o10bvs1r & 3s5zvs3s6z &  5mvs6m &  6hvs8z &  medium &  hard \\
        \midrule
 COMA & 6.84 & 3.5 & 0.92 & 3.11 & 257.66 & >1000 \\ \midrule
 MAPPO & 0.55 & 0.54 & 2.4 & 1.81 & 0.79 & 0.3 \\ \midrule
 MAT & 4.47 & 5.24 & 4.08 & 20.14 & 10.12 & 14.18 \\ \midrule
 IPPO-Comm & 0.57 & 0.82 & 2.14 & 1.57 & 0.5 & 0.43 \\ \midrule
 IPPO-Comm-OB-KL & \underline{\textbf{0.39}} & \underline{\textbf{0.24}} & \underline{\textbf{0.46}} & 0.9 & \underline{\textbf{0.01}} & 0.01 \\ \midrule
 GAAC & 1.76 & 1.09 & 0.64 & 6.28 & 1.11 & 0.67 \\ \midrule
 GAAC-OB-KL & 0.49 & 0.81 & 0.6 & \underline{\textbf{0.4}} & 0.04 & \underline{\textbf{<0.01}} \\ \bottomrule
    \end{tabular}
\end{sc}
\end{center}
\end{table}

\section{Conclusions}
In this paper, we investigate the variance of policy gradients caused by communication in decentralized multi-agent deep reinforcement learning. Specifically, we focus on the Decentralized Communicating Critics and Decentralized Actors (DCCDA) setting, where communication is allowed only among critics, while actors do not communicate during training and execution. By variance analysis, we prove that DCCDA policy gradients have a higher or equal variance than the policy gradients under CTDE without communication. We further propose a message-dependent baseline technique for variance reduction in policy gradients, and a regularization technique to improve the learning of critics with experience generated by decentralized policies. We theoretically prove that the optimal message-dependent baseline can reduce the variance in DCCDA policy gradients. The experiments on several tasks show that our proposed techniques achieve not only a higher learning performance but also reduced variance in policy gradients. In the future, we would like to investigate the theoretical variance analysis and the techniques under various communication scenarios. Moreover, we will consider how continuous values of messages affect the variance in policy gradients.

\bibliographystyle{unsrt}  
\bibliography{variance_reduction}

\newpage
\appendix
\onecolumn
\section*{Appendix}



We list important symbols and definitions used in the following deviations,
\begin{itemize}
  \item $\boldsymbol{h}_t=\{h^1_t,...,h^N_t\}$: the joint histories of $N$ agents at time step $t$, which include all previous observations, actions, and the current observation till time step $t$. We use $\boldsymbol{H}$ denote all possible values of $\boldsymbol{h}_t$, i.e., $\boldsymbol{h}_t \in \boldsymbol{H}$. We use $h^{-i}_t$ to denote the other agents' histories and therefore: $\boldsymbol{h}_t=\{h^i_t, h^{-i}_t\}$.
  \item $\boldsymbol{a}_t=\{a^1_t,...,a^N_t\}$: the joint actions of $N$ agents at time step $t$, generated by policies. We use $a^{-i}_t$ to denote the other agents' actions and therefore: $\boldsymbol{a}_t=\{a^i_t, a^{-i}_t\}$.
  \item $\boldsymbol{o}_{t+1}=\{o^1_{t+1},...,o^N_{t+1}\}$: the joint new observations of $N$ agents (at time step $t+1$). We use $o^{-i}_t$ to denote the other agents' new observations and therefore: $\boldsymbol{o}_t=\{o^i_t, o^{-i}_t\}$.
  \item $\boldsymbol{h}_{t+1}=\{\boldsymbol{h}_t, \boldsymbol{a}_t, \boldsymbol{o}_{t+1} \}$ is the next joint history of all agents. We further use $\boldsymbol{hao}$ to denote $\{\boldsymbol{h}_t, \boldsymbol{a}_t, \boldsymbol{o}_{t+1} \}$.
  \item $\boldsymbol{m}_t=\{m^1_t...,m^N_t\}$: the joint (sending) messages of $N$ agents at time step $t$. During the implementation, each message is generated from a probabilistic message function $f^{msg}$ conditioned on the sender agent’s history and action with parameters: $m_j \sim f^{msg}(m_j|h_j, a_j)$, where $j \in \{1,...,i-1,i+1,...,N\}$. With broadcasting communication, the received message of each agent $i$ is denoted by $m_{-i}=\{m^1_t...,m^{i-1}_t,m^{i+1}_t,m^N_t\}$, i.e., the set of messages received from all other agents.
  \item $\boldsymbol{\pi}(\boldsymbol{a}_t|\boldsymbol{h}_t)$: the centralized and joint policies of agents at time step $t$, considering the joint histories $\boldsymbol{h}_t$ of all agents.
  \item $\pi_i(a^i_t|h^i_t)$: the decentralized policy without communication of each agent $i$ at time step $t$, considering only the individual history $h^i_t$.
  \item $\pi_i(a^i_t|h^i_t,m^{-i}_t)$: the decentralized policy with communication of each agent $i$ at time step $t$, considering the individual history $h^i_t$ and received messages $m^{-i}_t$.
  \item $\epsilon_i$ is the noise associated to agent $i$, which corrupts received messages from other agents. We define $\epsilon_i \in \mathcal{E}$, where $\mathcal{E}$ is the set of all possible noise values, i.e., integers. Each value of noise represents a specific type of error. The noise term for each agent is sampled from a distribution, i.e., $\epsilon_i \sim p(\epsilon_i)$, where $p(\epsilon_i)$ denotes the probability distribution over $\epsilon_i \in \mathcal{E}$. We assume that a noise term will be sampled at every time step and denote as $\epsilon^i_t$ with probability $p(\epsilon^i_t)$.
\end{itemize}

Note that we omit the time step of histories, actions, observations, messages, and policies in the following deviations for notation simplification. The notation follows conventions in RL and MARL \cite{Haarnoja2018SAC,Lyu2023Centralized}.

\section{Proofs of the Theoretical Results in Idealistic Communication Setting}
\label{sec:proofsIdealistic}

We first consider an idealistic communication setting which can simplify complex distributions in the following theoretical analysis. Specifically, we have the following assumption:
\setcounter{theorem}{0}
\begin{assumption}
In idealistic communication setting, each received message correctly represents the sender's local information, i.e., histories and actions.
\label{assumptionOne}
\end{assumption}

The ideal communication assumption implies the existence of a perfect message decoder from the receiver's perspective, capable of reconstructing the sender agents' information from the received messages. Therefore, communication induces complete and sound information from all agents, which can be used to relate communicating critics and centralized critics.


Under the idealistic communication assumption, we come to the following lemma:  

\setcounter{theorem}{0}
\begin{lemma}
The on-policy centralized critic equals to the expectation of on-policy decentralized communicating critics in idealistic communication setting: $Q^{\pi}(\boldsymbol{h}, \boldsymbol{a})=\mathbb{E}_{m_{-i}|\boldsymbol{h}, \boldsymbol{a}}[Q^{\pi}_i(h_i, a_i, m_{-i})]$
\label{lemmaOne}
\end{lemma}

\textit{Proof of Lemma \ref{lemmaOne}}. To prove the lemma, we first write the centralized critics and decentralized communicating critics based on the Bellman equations, and then investigate their relation. Followed by Lyu et al. \cite{Lyu2023Centralized}, the centralized Q-function (critic) under the Bellman equality is defined as follows\footnote{We use Q-functions instead of critics when involving the Bellman equations.}:
\begin{equation}
Q^{\boldsymbol{\pi}}(\boldsymbol{h}, \boldsymbol{a})=\mathcal{R}(\boldsymbol{h}, \boldsymbol{a}) + \gamma \mathbb{E}_{\boldsymbol{o}|\boldsymbol{h},\boldsymbol{a}}[\sum_{\boldsymbol{a}'}\boldsymbol{\pi}(\boldsymbol{a}'| \boldsymbol{hao})Q^{\boldsymbol{\pi}}(\boldsymbol{hao},\boldsymbol{a}')]
\label{eq:jointQ}
\end{equation}
where $\mathcal{R}(\boldsymbol{h}, \boldsymbol{a}) \doteq \mathbb{E}_{s|\boldsymbol{h}, \boldsymbol{a}}[\mathcal{R}(s,\boldsymbol{a})]$ is the joint history reward function and $s \sim p(s|\boldsymbol{h}, \boldsymbol{a})$ is how agents infer the state distribution based on current information. $Q^{\boldsymbol{\pi}}(\boldsymbol{h}, \boldsymbol{a})$ is the expected long-term performance of the team of agents when each agent (with policy $\pi_i$) has observed the individual history $h_i$ and has opted to perform a first action $a_i$. 

Similarly, we define a decentralized communicating Q-function (critic) following the Bellman equality as follows:
\begin{equation}
Q^{\pi}_i(h_i, a_i, m_{-i})=\mathcal{R}_i(h_i, a_i, m_{-i}) + \gamma \mathbb{E}_{o_i, m'_{-i}|h_i, a_i, m_{-i}}[\sum_{a'_i}\pi_i(a'_i| hao_i, m'_{-i})Q^{\pi}_i(hao_i, a'_i, m'_{-i})]
\label{eq:QwithComm}
\end{equation}
where $\mathcal{R}_i(h_i, a_i, m_{-i}) \doteq \mathbb{E}_{s|h_i, a_i, m_{-i}}[\mathcal{R}(s, (a_i, a_{-i}))]$ for $a_{-i}$ from the same joint action. $\mathcal{R}_i(h_i, a_i, m_{-i})$ is the individual reward function with communication when observing individual history $h_i$, action $a_i$, and received messages $m_{-i}$ when other agents opt for action $a_{-i}$. In the equation, $m'_{-i}$ is the next received messages. The individual reward function is defined to consider all possible states under history information. Given the same joint action considered at the current time step, we derive the relation between the joint history reward function and the individual reward function with communication as follows:

\begin{subequations}
\begin{align}
\mathcal{R}(\boldsymbol{h}, \boldsymbol{a})&=\mathbb{E}_{s|\boldsymbol{h}, \boldsymbol{a}}[\mathcal{R}(s, \boldsymbol{a})] \\
&=\sum_s p(s|\boldsymbol{h}, \boldsymbol{a}) \mathcal{R}(s, \boldsymbol{a}) \\
&= \sum_{s, m_{-i}} p(s, m_{-i}|\boldsymbol{h}, \boldsymbol{a}) \mathcal{R}(s, \boldsymbol{a}) \\
&=\sum_{m_{-i}} p(m_{-i}|\boldsymbol{h}, \boldsymbol{a}) \sum_s p(s|m_{-i}, \boldsymbol{h}, \boldsymbol{a}) \mathcal{R}(s, \boldsymbol{a}) \\
&\stackrel{imp.}{=}\sum_{m_{-i}} p(m_{-i}|\boldsymbol{h}, \boldsymbol{a}) \sum_s p(s|m_{-i}, h_i, a_i) \mathcal{R}(s, \boldsymbol{a}) \\
&\stackrel{def.}{=}\sum_{m_{-i}} p(m_{-i}|\boldsymbol{h}, \boldsymbol{a}) \mathcal{R}_i(h_i, a_i, m_{-i}) \\
&=\mathbb{E}_{m_{-i}|\boldsymbol{h}, \boldsymbol{a}}[\mathcal{R}_i(h_i, a_i, m_{-i})]
\end{align}
\label{eq:reward}
\end{subequations}
The line \ref{eq:reward}c is due to the marginalized expectation w.r.t. messages. In line \ref{eq:reward}d, as received messages can correctly represent sender agents' history and actions (Assumption \ref{assumptionOne}), $s$ is conditionally independent of other agents' information $h_{-i}$ and $a_{-i}$ given received messages $m_{-i}$, and therefore: $p(s|m_{-i}, \boldsymbol{h}, \boldsymbol{a})=p(s|m_{-i}, h_i, a_i)$. The simplification indicates that each receiver agent $i$ can infer the distribution of the current states ($s$) based on its current information, i.e., individual history ($h_i$), action ($a_i$), and received message ($m_{-i}$). In line \ref{eq:reward}f, we use the definition of individual reward function with communication in Equation \ref{eq:QwithComm}. As a result, Equation \ref{eq:reward} shows that the joint history reward function is equivalent to the individual reward function by summing up all possible messages when agents opt for the same joint action.

We then analyze how messages affect future rewards in the Bellman equations, considering the stochastic effect of messages in future time steps. In the following equation, we derive the relation between Q-functions for the next time step. We substitute the Q-function $\mathbb{E}_{m'_{-i}|\boldsymbol{hao}, \boldsymbol{a'}}[Q^{\pi}_i(hao_i, a'_i, m'_{-i})]$ (at the next time step) into the second term of Equation \ref{eq:jointQ} and obtain:

\begin{subequations}
\begin{align}
& \gamma \mathbb{E}_{\boldsymbol{o}|\boldsymbol{h}, \boldsymbol{a}}[\sum_{\boldsymbol{a}'}\boldsymbol{\pi}(\boldsymbol{a}'| \boldsymbol{hao})\mathbb{E}_{m'_{-i}|\boldsymbol{hao}, \boldsymbol{a'}}[Q^{\pi}_i(hao_i, a'_i, m'_{-i})]] \\
&= \gamma \mathbb{E}_{\boldsymbol{o}|\boldsymbol{h}, \boldsymbol{a}}[\sum_{\boldsymbol{a}'}\boldsymbol{\pi}(\boldsymbol{a}'| \boldsymbol{hao})\sum_{m'_{-i}}p(m'_{-i}|\boldsymbol{hao}, \boldsymbol{a'})Q^{\pi}_i(hao_i, a'_i, m'_{-i})] \\
&= \gamma \sum_{\boldsymbol{o}}p(\boldsymbol{o}|\boldsymbol{h}, \boldsymbol{a})\sum_{\boldsymbol{a}'}\boldsymbol{\pi}(\boldsymbol{a}'| \boldsymbol{hao})\sum_{m'_{-i}}p(m'_{-i}|\boldsymbol{hao}, \boldsymbol{a'})Q^{\pi}_i(hao_i, a'_i, m'_{-i}) \\
&= \gamma \sum_{\boldsymbol{o}, \boldsymbol{a}', m'_{-i}}p(\boldsymbol{o}|\boldsymbol{h}, \boldsymbol{a})\boldsymbol{\pi}(\boldsymbol{a}'| \boldsymbol{hao})p(m'_{-i}|\boldsymbol{hao}, \boldsymbol{a'})Q^{\pi}_i(hao_i, a'_i, m'_{-i}) \\
&= \gamma \sum_{\boldsymbol{o}, \boldsymbol{a}', m'_{-i}} \frac{\cancel{p(\boldsymbol{h}, \boldsymbol{o}, \boldsymbol{a})}}{p(\boldsymbol{h}, \boldsymbol{a})} \frac{\cancel{p(\boldsymbol{a}', \boldsymbol{h}, \boldsymbol{a}, \boldsymbol{o}))}}{\cancel{p(\boldsymbol{h}, \boldsymbol{a}, \boldsymbol{o})}} \frac{p(m'_{-i}, \boldsymbol{h}, \boldsymbol{a}, \boldsymbol{o}, \boldsymbol{a'})}{\cancel{p(\boldsymbol{h}, \boldsymbol{a}, \boldsymbol{o}, \boldsymbol{a'})}}Q^{\pi}_i(hao_i, a'_i, m'_{-i}) \\
&= \gamma \sum_{\boldsymbol{o}, \boldsymbol{a}', m'_{-i}}\frac{p(m'_{-i}, \boldsymbol{h}, \boldsymbol{a}, \boldsymbol{o}, \boldsymbol{a'})}{p(\boldsymbol{h}, \boldsymbol{a})}Q^{\pi}_i(hao_i, a'_i, m'_{-i}) \\
&= \gamma \sum_{\boldsymbol{o}, \boldsymbol{a}', m'_{-i}}p(\boldsymbol{o}, \boldsymbol{a}', m'_{-i}|\boldsymbol{h}, \boldsymbol{a})Q^{\pi}_i(hao_i, a'_i, m'_{-i}) \\
&= \gamma \sum_{o_i, a'_i, m'_{-i}}p(o_i, a'_i, m'_{-i}|\boldsymbol{h}, \boldsymbol{a})Q^{\pi}_i(hao_i, a'_i, m'_{-i}) \\
&= \gamma \sum_{m_{-i}, o_i, a'_i, m'_{-i}}p(m_{-i}, o_i, a'_i, m'_{-i}|\boldsymbol{h}, \boldsymbol{a})Q^{\pi}_i(hao_i, a'_i, m'_{-i}) \\
&= \gamma \sum_{m_{-i}, o_i, a'_i, m'_{-i}}p(m_{-i}|\boldsymbol{h}, \boldsymbol{a})p(o_i, a'_i, m'_{-i}|\boldsymbol{h}, \boldsymbol{a},m_{-i})Q^{\pi}_i(hao_i, a'_i, m'_{-i}) \\
&= \gamma \sum_{m_{-i}, o_i, a'_i, m'_{-i}}p(m_{-i}|\boldsymbol{h}, \boldsymbol{a})p(o_i, m'_{-i}|\boldsymbol{h}, \boldsymbol{a},m_{-i})p(a'_i|\boldsymbol{h}, \boldsymbol{a},m_{-i},o_i, m'_{-i})Q^{\pi}_i(hao_i, a'_i, m'_{-i}) \\
&\stackrel{imp.}{=} \gamma \sum_{m_{-i}, o_i, a'_i, m'_{-i}}p(m_{-i}|\boldsymbol{h}, \boldsymbol{a})p(o_i, m'_{-i}|h_i, a_i, m_{-i})\pi_i(a'_i|h_i,a_i,o_i, m'_{-i})Q^{\pi}_i(hao_i, a'_i, m'_{-i}) \\
&= \gamma \sum_{m_{-i}}p(m_{-i}|\boldsymbol{h}, \boldsymbol{a})\sum_{o_i, m'_{-i}}p(o_i, m'_{-i}|h_i, a_i, m_{-i})\sum_{a'_i}\pi_i(a'_i|h_i,a_i,o_i, m'_{-i})Q^{\pi}_i(hao_i, a'_i, m'_{-i}) \\
&= \gamma \mathbb{E}_{m_{-i}|\boldsymbol{h}, \boldsymbol{a}}[\mathbb{E}_{o_i, m'_{-i}|h_i, a_i, m_{-i}}[\sum_{a'_i}\pi_i(a'_i| hao_i, m'_{-i})Q^{\pi}_i(hao_i, a'_i, m'_{-i})]]
\end{align}
\label{eq:Qfunction}
\end{subequations}
In line \ref{eq:Qfunction}g, we sum up the other agents' next observations ($o_{-i}$) and next actions ($a'_{-i}$) since they do not determine agent $i$' Q-values (due to marginalization). In line \ref{eq:Qfunction}i, adding a new variable $m_{-i}$ does not change the expected quantity due to marginalized distribution, where all possible messages are considered in the enumeration. From line \ref{eq:Qfunction}k to line \ref{eq:Qfunction}l, we use Assumption 1 to simplify the distribution $p(o_i, m'_{-i}|\boldsymbol{h}, \boldsymbol{a}, m_{-i})$ to $p(o_i, m'_{-i}|m_{-i}, h_i, a_i)$: the currently received message ($m_{-i}$) has included other agents' current information (i.e., $h_{-i}$, $a_{-i}$), and agent $i$ can infer the next observations and the next received messages based on its current information. We further simplify the distribution $p(a'_i|\boldsymbol{h}, \boldsymbol{a}, m_{-i},o_i, m'_{-i})$ to $p(a'_i|h_i,a_i,m_{-i},o_i, m'_{-i})$, because the next received message $m_{-i}'$ has already included sender agents' (previous) histories $h_{-i}$ and $a_{-i}$ (due to Assumption 1). Since messages are generated to include all historical information, we assume that each agent's policy depends only on current messages and can ignore history information (e.g., previous messages $m_{-i}$) due to Markovian property. This simplifies the distribution $p(a'_i|h_i,a_i,o_i, m_{-i}, m'_{-i})$ to $p(a'_i|h_i,a_i,o_i, m'_{-i})$. The distribution $p(a'_i|h_i,a_i,o_i, m'_{-i})$ indicates that receiver agent $i$ can decide its next action ($a'_i$) based on its next history ($h_i$, $a_i$, $o_i$) and the next received message ($m'_{-i}$). We further denote $p(a'_i|h_i,a_i,o_i, m'_{-i})$ as $\pi_i(a'_i|h_i,a_i,o_i, m'_{-i})$ for notation consistency.

By bringing Equations \ref{eq:reward} and \ref{eq:Qfunction} together, we achieve the following equality:

\begin{equation}
\begin{split}
&\mathcal{R}(\boldsymbol{h}, \boldsymbol{a}) + \gamma \mathbb{E}_{\boldsymbol{o}|\boldsymbol{h}, \boldsymbol{a}}[\sum_{\boldsymbol{a}'}\boldsymbol{\pi}(\boldsymbol{a}'| \boldsymbol{hao})\mathbb{E}_{m'_{-i}|\boldsymbol{hao}, \boldsymbol{a'}}[Q^{\pi}_i(hao_i, a'_i, m'_{-i})]] \\
&\stackrel{eq. \ref{eq:reward} \& \ref{eq:Qfunction}}{=}\mathbb{E}_{m_{-i}|\boldsymbol{h}, \boldsymbol{a}}[\mathcal{R}_i(h_i, a_i, m_{-i})] + \gamma \mathbb{E}_{m_{-i}|\boldsymbol{h}, \boldsymbol{a}}[\mathbb{E}_{o_i, m'_{-i}|h_i, a_i, m_{-i}}[\sum_{a'}\pi_i(a'| hao_i, m'_{-i})Q^{\pi}_i(hao_i, a', m'_{-i})]] \\
&\stackrel{def.}{=} \mathbb{E}_{m_{-i}|\boldsymbol{h}, \boldsymbol{a}}[\mathcal{R}_i(h_i, a_i, m_{-i}) + \gamma \mathbb{E}_{o_i, m'_{-i}|h_i, a_i, m_{-i}}[\sum_{a'}\pi_i(a'| hao_i, m'_{-i})Q^{\pi}_i(hao_i, a', m'_{-i})]] \\
&= \mathbb{E}_{m_{-i}|\boldsymbol{h}, \boldsymbol{a}}[Q^{\pi}_i(h_i, a_i, m_{-i})]
\end{split}
\label{eq:jointocomm}
\end{equation}
where the final equality follows Equation \ref{eq:QwithComm}. Note that $\mathbb{E}_{m'_{-i}|\boldsymbol{hao}, \boldsymbol{a'}}[Q^{\pi}_i(hao_i, a'_i, m'_{-i})]$ and $\mathbb{E}_{m_{-i}|\boldsymbol{h}, \boldsymbol{a}}[Q^{\pi}_i(h_i, a_i, m_{-i})]$ are the same Q-function evaluated at different time steps. Based on Equation \ref{eq:jointocomm}, we have,
\begin{equation}
\mathbb{E}_{m_{-i}|\boldsymbol{h}, \boldsymbol{a}}[Q^{\pi}_i(h_i, a_i, m_{-i})] = \mathcal{R}(\boldsymbol{h}, \boldsymbol{a}) + \gamma \mathbb{E}_{\boldsymbol{o}|\boldsymbol{h}, \boldsymbol{a}}[\sum_{\boldsymbol{a}'}\boldsymbol{\pi}(\boldsymbol{a}'| \boldsymbol{hao})\mathbb{E}_{m'_{-i}|\boldsymbol{hao}, \boldsymbol{a'}}[Q^{\pi}_i(hao_i, a'_i, m'_{-i})]]
\label{eq:jointocommNew}
\end{equation}
Essentially, after substituting the decentralized communicating Q-function ($\mathbb{E}_{m'_{-i}|\boldsymbol{hao}, \boldsymbol{a'}}[Q^{\pi}_i(hao_i, a'_i, m'_{-i})]$) into the left side of the centralized Q-function under the Bellman equation (Equation \ref{eq:jointQ}), we still get back the decentralized communicating Q-function ($\mathbb{E}_{m_{-i}|\boldsymbol{h}, \boldsymbol{a}}[Q^{\pi}_i(h_i, a_i, m_{-i})]$). Followed by Lyu et al's theory \cite{Lyu2023Centralized}, the centralized Q-function $Q^{\pi}(\boldsymbol{h}, \boldsymbol{a})$ is the unique fixed point of the Bellman equation defined in Equation \ref{eq:jointQ} (due to the contraction mapping of the Bellman operator). Since $\mathbb{E}_{m_{-i}|\boldsymbol{h}, \boldsymbol{a}}[Q^{\pi}_i(h_i, a_i, m_{-i})]$ appears to be also the solution of the same Bellman equation (Equation \ref{eq:jointQ}), we have $Q^{\pi}(\boldsymbol{h}, \boldsymbol{a})=\mathbb{E}_{m_{-i}|\boldsymbol{h}, \boldsymbol{a}}[Q^{\pi}_i(h_i, a_i, m_{-i})]$, which completes the proof.

\subsection{Proof of Theorem 1}
\label{app:variance}

We have the following theorem:

\setcounter{theorem}{0}
\begin{theorem}
The DCCDA sample gradient has a variance greater or equal than that of the
CTDE sample gradient in idealistic communication setting: $Var(\hat{g}^i_{DCCDA}) \geq Var(\hat{g}^i_{CTDE})$.
\label{theoremOne}
\end{theorem}

\textit{Proof of Theorem \ref{theoremOne}}. We first check the relation between the two expected gradients $g^i_{DCCDA}$ and $g^i_{CTDE}$, which can greatly simplify the later variance analysis. Based on Lemma \ref{lemmaOne}, as $Q^{\pi}(\boldsymbol{h}, \boldsymbol{a})=\mathbb{E}_{m_{-i}|\boldsymbol{h},\boldsymbol{a}}[Q^{\pi}_i(h_i, a_i, m_{-i})]$, we have,
\begin{subequations}
\begin{align}
g^i_{DCCDA} &= \mathbb{E}_{\boldsymbol{h},\boldsymbol{a},\boldsymbol{m}}[Q^{\pi}_i(h_i, a_i, m_{-i})\nabla_{\theta_i}\log \pi_i(a_i|h_i, \theta_i)] \\
&= \mathbb{E}_{\boldsymbol{h},\boldsymbol{a}}[\mathbb{E}_{\boldsymbol{m}|\boldsymbol{h},\boldsymbol{a}}[Q^{\pi}_i(h_i, a_i, m_{-i})\nabla_{\theta_i}\log \pi_i(a_i|h_i, \theta_i)] \\
&= \mathbb{E}_{\boldsymbol{h},\boldsymbol{a}}[\mathbb{E}_{m_{-i}|\boldsymbol{h},\boldsymbol{a}}[Q^{\pi}_i(h_i, a_i, m_{-i})\nabla_{\theta_i}\log \pi_i(a_i|h_i, \theta_i)]] \\
&= \mathbb{E}_{\boldsymbol{h},\boldsymbol{a}}[\mathbb{E}_{m_{-i}|\boldsymbol{h},\boldsymbol{a}}[Q^{\pi}_i(h_i, a_i, m_{-i})]\nabla_{\theta_i}\log \pi_i(a_i|h_i, \theta_i)] \\
&\stackrel{lem. 1}{=}  \mathbb{E}_{\boldsymbol{h},\boldsymbol{a}}[Q^{\pi}(\boldsymbol{h}, \boldsymbol{a})\nabla_{\theta_i}\log \pi_i(a_i|h_i, \theta_i)] \\
&= g^i_{CTDE}
\end{align}
\label{eq:bias}
\end{subequations}
where line \ref{eq:bias}c is due to the quantity inside the expectation does not depend on agent $i$'s message $m_{i}$.

Based on Equation \ref{eq:bias}, we come to the following proof of Theorem \ref{theoremOne}. Note that $\hat{g}^i_{CTDE}=Q^{\boldsymbol{\pi}}(\boldsymbol{h}, \boldsymbol{a})\nabla_{\theta_i}\log \pi_i(a_i|h_i, \theta_i)$ and $\hat{g}^i_{DCCDA}=Q^{\pi}_i(h_i, a_i, m_{-i})\nabla_{\theta_i}\log \pi_i(a_i|h_i, \theta_i)$. In the following equation, in order to simplify the notation, we denote $S=\nabla_{\theta_i}\log \pi_i(a_i|h_i, \theta_i)^T\nabla_{\theta_i}\log \pi_i(a_i|h_i, \theta_i)$, which is the inner product of $\nabla_{\theta_i}\log \pi_i(a_i|h_i, \theta_i)$ (due to the square of it):

\begin{subequations}
\begin{align}
&Var(\hat{g}^i_{DCCDA}) - Var(\hat{g}^i_{CTDE}) \\
&\stackrel{def.}{=} \left(\mathbb{E}_{\boldsymbol{h},\boldsymbol{a},\boldsymbol{m}}[(\hat{g}^i_{DCCDA})^2] - \left (\mathbb{E}_{\boldsymbol{h},\boldsymbol{a},\boldsymbol{m}}[\hat{g}^i_{DCCDA}] \right)^2 \right) - \left(\mathbb{E}_{\boldsymbol{h},\boldsymbol{a}}[(\hat{g}^i_{CTDE})^2] - \left (\mathbb{E}_{\boldsymbol{h},\boldsymbol{a}}[\hat{g}^i_{CTDE}] \right)^2 \right)  \\
&= \left(\mathbb{E}_{\boldsymbol{h},\boldsymbol{a},\boldsymbol{m}}[(\hat{g}^i_{DCCDA})^2] - \mathbb{E}_{\boldsymbol{h},\boldsymbol{a}}[(\hat{g}^i_{CTDE})^2] \right) - \left( \left( \mathbb{E}_{\boldsymbol{h},\boldsymbol{a},\boldsymbol{m}}[\hat{g}^i_{DCCDA}] \right)^2 - \left (\mathbb{E}_{\boldsymbol{h},\boldsymbol{a}}[\hat{g}^i_{CTDE}] \right)^2 \right)  \\
&\stackrel{eq. \ref{eq:bias}}{=} \left(\mathbb{E}_{\boldsymbol{h},\boldsymbol{a},\boldsymbol{m}}[(\hat{g}^i_{DCCDA})^2] - \mathbb{E}_{\boldsymbol{h},\boldsymbol{a}}[(\hat{g}^i_{CTDE})^2] \right) - \underbrace{ \left((g^i_{DCCDA})^2 - (g^i_{CTDE})^2\right)}_{=0}  \\
&=  \mathbb{E}_{\boldsymbol{h},\boldsymbol{a},\boldsymbol{m}}[(\hat{g}^i_{DCCDA})^2] - \mathbb{E}_{\boldsymbol{h},\boldsymbol{a}}[(\hat{g}^i_{CTDE})^2]  \\
&\stackrel{def.}{=}  \mathbb{E}_{\boldsymbol{h},\boldsymbol{a},\boldsymbol{m}}[(Q^{\pi}_i(h_i, a_i, m_{-i})\nabla_{\theta_i}\log \pi_i(a_i|h_i, \theta_i))^2] - \mathbb{E}_{\boldsymbol{h},\boldsymbol{a}}[(Q^{\pi}(\boldsymbol{h}, \boldsymbol{a})\nabla_{\theta_i}\log \pi_i(a_i|h_i, \theta_i))^2] \\
&= \mathbb{E}_{\boldsymbol{h},\boldsymbol{a},\boldsymbol{m}}[Q^{\pi}_i(h_i, a_i, m_{-i})^2S] - \mathbb{E}_{\boldsymbol{h},\boldsymbol{a}}[Q^{\pi}(\boldsymbol{h}, \boldsymbol{a})^2S] \\
&= \mathbb{E}_{\boldsymbol{h},\boldsymbol{a}}[\mathbb{E}_{\boldsymbol{m}|\boldsymbol{h},\boldsymbol{a}}[Q^{\pi}_i(h_i, a_i, m_{-i})^2S]] - \mathbb{E}_{\boldsymbol{h},\boldsymbol{a}}[Q^{\pi}(\boldsymbol{h}, \boldsymbol{a})^2S] \\
&= \mathbb{E}_{\boldsymbol{h},\boldsymbol{a}}[\mathbb{E}_{m_{-i}|\boldsymbol{h},\boldsymbol{a}}[Q^{\pi}_i(h_i, a_i, m_{-i})^2S]] - \mathbb{E}_{\boldsymbol{h},\boldsymbol{a}}[Q^{\pi}(\boldsymbol{h}, \boldsymbol{a})^2S] \\
&= \mathbb{E}_{\boldsymbol{h},\boldsymbol{a}}[\left (\mathbb{E}_{m_{-i}|\boldsymbol{h},\boldsymbol{a}}[Q^{\pi}_i(h_i, a_i, m_{-i})^2] - Q^{\pi}(\boldsymbol{h}, \boldsymbol{a})^2 \right ) S] \\
&= \mathbb{E}_{\boldsymbol{h},\boldsymbol{a}, \boldsymbol{m}}[\left (Q^{\pi}_i(h_i, a_i, m_{-i})^2 - Q^{\pi}(\boldsymbol{h}, \boldsymbol{a})^2 \right ) S] \\
&\stackrel{lem. \ref{lemmaOne}}{=}  \mathbb{E}_{\boldsymbol{h},\boldsymbol{a}}[\underbrace{\left (\mathbb{E}_{m_{-i}|\boldsymbol{h},\boldsymbol{a}}[Q^{\pi}_i(h_i, a_i, m_{-i})^2] - (\mathbb{E}_{m_{-i}|\boldsymbol{h},\boldsymbol{a}}[Q^{\pi}_i(h_i, a_i, m_{-i})])^2 \right )}_{u} S] \\
&\ge 0
\end{align}
\label{eq:variance}
\end{subequations}
where line \ref{eq:variance}b follows the definition of variance. Line \ref{eq:variance}b follows the definitions that $g^i_{DCCDA}= \mathbb{E}[\hat{g}^i_{DCCDA}]$ and $g^i_{CTDE}= \mathbb{E}[\hat{g}^i_{CTDE}]$. Moreover, due to Equation \ref{eq:bias}, we have $g^i_{DCCDA}=g^i_{CTDE}$ and then $(g^i_{DCCDA})^2- (g^i_{CTDE})^2=0$ in line \ref{eq:variance}d. In line \ref{eq:variance}f, we replace $g^i_{DCCDA}$ and $g^i_{CTDE}$ with the defined formula and it shows a square of the gradient $\nabla_{\theta_i}\log \pi_i(a_i|h_i, \theta_i)$. We then use $S$ to simplify the notation. Line \ref{eq:variance}i is because the quantity inside the expectation does not depend on agent $i$'s message $m_i$. Line \ref{eq:variance}l is according to Lemma \ref{lemmaOne}. The final inequality in line \ref{eq:variance}m follows because $u \ge 0$ by Jensen’s inequality: $\mathbb{E}[X^2] \ge (\mathbb{E}[X])^2$, and $S$ is the inner product of a vector itself and therefore non-negative. Therefore, the DCCDA sample gradient has a variance greater or equal than that of the CTDE sample gradient, i.e., $Var(\hat{g}^i_{DCCDA}) \geq Var(\hat{g}^i_{CTDE})$, which completes the proof.

\section{Proofs of the Theoretical Results in Non-idealistic Communication Setting}
\label{sec:proofsNonIdealistic}

We further consider a non-idealistic communication setting where messages are corrupted with a noisy term. The noisy term could come from the imperfection of decoders, e.g., due to the use of neural networks. Therefore, from receiver agent $i$'s perspective, received message $m^{i}_j$ from sender agent $j$ is dedicated to incorporate a noisy term $\epsilon_i \in \mathcal{E}$ when sender agent $j$'s sends a message: $m^{i}_j=<m_j, \epsilon_i>$, where $m_j \sim f^{msg}(m_j|h_j, a_j)$ is a message broadcast by sender agent $j$ and $\epsilon_i$ is the noise associated to agent $i$. In order to simplify the analysis, we assume there exists a decoder that could decode $h_j$ and $a_j$ from message $m_j$ and therefore we could omit $f_j$, i.e., $m^{i}_j= <h_j, a_j, \epsilon_i>$, while messages can still be imperfect and remain noisy. Then, the message used by receiver agent $i$ is denoted as $m_{-i}=<m^{i}_1,...,m^{i}_{i-1},m^{i}_{i+1},...,m^{i}_N>=<h_{-i}, a_{-i}, \epsilon_i>$

Based on the above non-idealistic communication setting, the communicating critic of receiver agent $i$ is denoted as $Q^{\pi}_i(h_i, a_i, m_{-i})$, and we have: $Q^{\pi}_i(h_i,a_i, m_{-i})=Q^{\pi}_i(h_i,a_i, <h_{-i}, a_{-i}, \epsilon_i>)=Q^{\pi}_i(\boldsymbol{h}, \boldsymbol{a}, \epsilon_i)$, where noise in communication is lifted to Q-values, leading to individual but centralized critics with additive noise.

We further investigate how noise affects the value estimation in critics and also the relation between individual but centralized critics with additive noise $Q^{\pi}_i(\boldsymbol{h}, \boldsymbol{a}, \epsilon_i)$ and individual but centralized critics without noise $Q^{\pi}_i(\boldsymbol{h}, \boldsymbol{a})$, which is essential for our variance analysis. Due to the noise in communication, rewards become noisy as well. Given history $\boldsymbol{h} \in \boldsymbol{H}$, action $\boldsymbol{a} \in \boldsymbol{\mathcal{A}}$, and noise term $\epsilon_i \in \mathcal{E}$, we define individual but centralized critics with noise at time step $t$ as:
\begin{equation}
\begin{aligned}
& Q^{\pi}_{i,t}(\boldsymbol{h}, \boldsymbol{a}, \epsilon_i) \\
&= \mathbb{E}_{s_{t+k},\boldsymbol{a}_{t+k}, \epsilon^i_{t+k}}[\sum_{k=0}^{T}\gamma^k \mathcal{R}_i(s_{t+k}, \boldsymbol{a}_{t+k}, \epsilon^i_{t+k}) | \boldsymbol{h}_t=\boldsymbol{h}, \boldsymbol{a}_t=\boldsymbol{a}, \epsilon^i_t=\epsilon_i] \\
&= \sum_{s_t, s_{t+1} \in \mathcal{S}} p(s_t, s_{t+1}|\boldsymbol{h}, \boldsymbol{a}) \Big(\mathcal{R}_i(s_t, \boldsymbol{a}, \epsilon_i) + \gamma \sum_{\boldsymbol{a}_{t+1} \in \boldsymbol{\mathcal{A}}, \boldsymbol{h}_{t+1} \in \boldsymbol{H}} p(\boldsymbol{h}_{t+1}|s_{t+1}) \boldsymbol{\pi}(\boldsymbol{a}_{t+1}|\boldsymbol{h}_{t+1}) \sum_{\epsilon^i_{t+1}} p(\epsilon^i_{t+1}) Q^{\pi}_{i,t+1}(\boldsymbol{h}_{t+1}, \boldsymbol{a}_{t+1}, \epsilon^i_{t+1}) \Big)
\end{aligned}
\label{eq:noisyQfunction}
\end{equation}
where $p(s_t, s_{t+1}|\boldsymbol{h}_t, \boldsymbol{a}_t)=p(s_t|\boldsymbol{h}_t, \boldsymbol{a}_t)p(s_{t+1}|\boldsymbol{h}_t, \boldsymbol{a}_t,s_t)$ represents the probabilities transiting from state $s_t$ to state $s_{t+1}$ under joint action $\boldsymbol{a}$ when observing history $\boldsymbol{h}_t$, $p(\boldsymbol{h}_{t+1}|s_{t+1})$ represents the probabilities of history $\boldsymbol{h}_{t+1}$ under future state $s_{t+1}$, $\boldsymbol{\pi}(\boldsymbol{a}_{t+1}|\boldsymbol{h}_{t+1})$ represents the history-based policy, and $\mathcal{R}_i(s_t, \boldsymbol{a}_t, \epsilon^i_t)$ is the noisy reward of agent $i$ at time step $t$. In Equation \ref{eq:noisyQfunction}, rewards considering noisy messages are accumulated when observing history $\boldsymbol{h}$, the first joint action to be $\boldsymbol{a}$, and the first noise to be $\epsilon_i$. Without noise, we define individual but centralized critics at time step $t$ as:
\begin{equation}
\begin{aligned}
& Q^{\pi}_{i,t}(\boldsymbol{h}, \boldsymbol{a}) \\
&= \mathbb{E}_{s_{t+k},\boldsymbol{a}_{t+k}}[\sum_{k=0}^{T}\gamma^k \mathcal{R}_i(s_{t+k}, \boldsymbol{a}_{t+k}, s_{t+k+1}) | \boldsymbol{h}_t=\boldsymbol{h}, \boldsymbol{a}_t=\boldsymbol{a}] \\
&= \sum_{s_t, s_{t+1} \in \mathcal{S}} p(s_t, s_{t+1}|\boldsymbol{h},\boldsymbol{a}) \Big(\mathcal{R}_i(s_t, \boldsymbol{a}) + \gamma \sum_{\boldsymbol{a}_{t+1} \in \boldsymbol{\mathcal{A}}, \boldsymbol{h}_{t+1} \in \boldsymbol{H}} p(\boldsymbol{h}_{t+1}|s_{t+1})\boldsymbol{\pi}(\boldsymbol{a}_{t+1}|\boldsymbol{h}_{t+1}) Q^{\pi}_{i,t+1}(\boldsymbol{h}_{t+1}, \boldsymbol{a}_{t+1}) \Big) 
\end{aligned}
\end{equation}
where $\mathcal{R}_i(s_{t}, \boldsymbol{a}_{t})$ is the true reward of agent $i$ at time step $t$. Without noise, rewards in $Q^{\pi}_{i,t}(\boldsymbol{h}, \boldsymbol{a})$ are accumulated when observing history $\boldsymbol{h}$ and the first joint action to be $\boldsymbol{a}$. Due to shared rewards in the Dec-POMDP setting, we have: $Q^{\pi}_{i,t}(\boldsymbol{h}, \boldsymbol{a})=Q^{\pi}_t(\boldsymbol{h}, \boldsymbol{a})$ by convergence, where $Q^{\pi}_t(\boldsymbol{h}, \boldsymbol{a})$ is the centralized Q-function at time step $t$.

Based on the individual but centralized critics with and without noise, we can see that the noise affects the estimation of Q-values when accumulating rewards. In single-agent reinforcement learning, Wang et al. \cite{Wang2020Noise} shows that noisy rewards can be compensated by utilizing a surrogate reward function, leading to unbiased value estimation. However, removing the effect of noise that we defined in value estimation may lead to increased variance when summing up noisy rewards. Inspired by the surrogate rewards proposed by Wang et al. \cite{Wang2020Noise}, we prove that in multi-agent reinforcement learning scenarios, using a surrogate reward function can also remove the effect of noise in value estimation (thereby become unbiased), while at the cost of increased variance. In the following proof, we consider a binary case where rewards indicate either success ($\mathbf{r}_+$) or failure ($\mathbf{r}_-$), while this can be generalized to rewards beyond binary (see Wang et al. \cite{Wang2020Noise}). Note that due to the shared rewards in the Dec-POMDP setting, agents have the same values of success or failure, i.e., $\mathbf{r}_+=\mathbf{r}^1_+=\mathbf{r}^2_+=...=\mathbf{r}^N_+$ and $\mathbf{r}_-=\mathbf{r}^1_-=\mathbf{r}^2_-=...=\mathbf{r}^N_-$, where $\mathbf{r}^i_+$ and $\mathbf{r}^i_-$ are rewards obtained by agent $i$ for success and failure, respectively. In the binary reward case, noisy rewards can be characterized by the noise rate parameter $e$, where we have:
$$
e \; :=\; p(\mathcal{R}_i(s_t, \boldsymbol{a}_t, \epsilon^i_t) \neq \mathcal{R}_i(s_t, \boldsymbol{a}_t))
$$
where $e$ defines the probability that noisy rewards are different from noise-free rewards under noise $\epsilon^i_t$ and $\mathcal{R}_i(s_{t}, \boldsymbol{a}_{t}) \in \mathbf{R} :=\{\mathbf{r}_+, \mathbf{r}_-\}$. $e$ implies that the probability of $\mathcal{R}_i(s_{t}, \boldsymbol{a}_{t}, \epsilon^i_t) \neq \mathcal{R}_i(s_{t}, \boldsymbol{a}_{t})$ can be affected by the noise term $\epsilon^i_t$. The noise term captures how communication under noisy conditions alters agents' chances of success or failure. With shared rewards, agents have the same values of $e$.

We determine the noisy reward $\mathcal{R}_i(s_{t}, \boldsymbol{a}_{t}, \epsilon^i_t)$ based on the noisy term as follows:

Suppose the noise-free (true) reward indicates success (i.e., $\mathcal{R}_i(s_{t}, \boldsymbol{a}_{t})=\mathbf{r}_+$) and there exists a value $\delta_{\epsilon} \in \mathcal{E}$, if the noise term $\epsilon^i_t \geq \delta_{\epsilon}$, then the noisy reward equals to the true reward and thus $\mathcal{R}_i(s_{t}, \boldsymbol{a}_{t}, \epsilon^i_t)=\mathbf{r}_+$, otherwise if the noise term $\epsilon^i_t < \delta_{\epsilon}$, the reward is considered to be erroneous and thus $\mathcal{R}_i(s_{t}, \boldsymbol{a}_{t}, \epsilon^i_t)=\mathbf{r}_-$. 


Let the set of noise terms $\boldsymbol{\epsilon}_+ = \{\epsilon^i_t \in \mathcal{E} | \epsilon^i_t \geq \delta_{\epsilon}\}$ represents the cases that noisy rewards equal to true rewards and $\boldsymbol{\epsilon}_- = \{\epsilon^i_t \in \mathcal{E} | \epsilon^i_t < \delta_{\epsilon}\}$ represents the cases that noisy rewards differ from true rewards. We have $\boldsymbol{\epsilon}_+ \cup \boldsymbol{\epsilon}_- = \mathcal{E}$ and $\boldsymbol{\epsilon}_+ \cap \boldsymbol{\epsilon}_- = \emptyset$. Then we denote $E=\{\boldsymbol{\epsilon}_+, \boldsymbol{\epsilon}_-\}$. We determine $e$ as:
$$
e = p(\mathcal{R}_i(s_{t}, \boldsymbol{a}_{t}, \epsilon^i_t) \neq \mathcal{R}_i(s_{t}, \boldsymbol{a}_{t}))= p(\boldsymbol{\epsilon}_-)
$$
Then we have, 
$$
1- e = p(\mathcal{R}_i(s_{t}, \boldsymbol{a}_{t}, \epsilon^i_t) = \mathcal{R}_i(s_{t}, \boldsymbol{a}_{t})) = p(\boldsymbol{\epsilon}_+)
$$
where $\boldsymbol{\epsilon}_+ \in E$ and $\boldsymbol{\epsilon}_- \in E$.


In the following two lemmas, we show that the effect of noisy rewards can be removed upon the inspiration of Wang et al \cite{Wang2020Noise}, leading to the equality between individual but centralized Q-function (without noise) and a defined surrogate Q-function (with noise). The equality between the two Q-functions significantly simplifies the variance analysis in CTDE policy gradients and the noisy version of DCCDA policy gradients. However, it still presents the issue of increased variance, as demonstrated later in Theorem \ref{theoremNoiseDCCDA}.

We list the definition of important symbols used in the proof of Lemma 2 for better reading:
\begin{itemize}
    \item $r^i \in \{\mathbf{r}_+, \mathbf{r}_-\}$ is the true reward of agent $i$ where noise is not presented.
    \item $r^i_\epsilon \in \{\mathbf{r}_+, \mathbf{r}_-\}$ is the noisy reward of agent $i$ where noise is presented.
\end{itemize}

Wang et al. \cite{Wang2020Noise} propose a surrogate reward function (Equation 1 in the paper of Wang et al.) to help remove the effect of noise in value estimation. Inspired by Wang et al., we define a surrogate reward function $\hat{\mathcal{R}}_i(s_t, \boldsymbol{a}_t, r^i, \epsilon^i_t)$ of agent $i$ as a function of state $s_t$, actions $\boldsymbol{a}_t$, the true reward $r^i$ and the noise term $\epsilon^i_t$:
\begin{equation}
\hat{\mathcal{R}}_i(s_t, \boldsymbol{a}_t, r^i, \epsilon^i_t)
\;:=\;
\begin{cases}
\dfrac{(1 - e)\mathbf{r}_+ - e \, \mathbf{r}_-}{1-2e},
& if (r^i=\mathbf{r}_+\, \&\, \epsilon^i_t \in \boldsymbol{\epsilon}_{+})\, or\, (r^i=\mathbf{r}_-\, \&\, \epsilon^i_t \in \boldsymbol{\epsilon}_{-}),\\[2em]
\dfrac{(1-e)\mathbf{r}_- - e\,\mathbf{r}_+}{1-2e},
& if (r^i=\mathbf{r}_+\, \&\, \epsilon^i_t \in \boldsymbol{\epsilon}_{-})\, or\, (r^i=\mathbf{r}_-\, \&\, \epsilon^i_t \in \boldsymbol{\epsilon}_{+}).
\end{cases}
\label{eq:surroRewards}
\end{equation}
where the true rewards and noise term determine noisy rewards, i.e., $r^i_\epsilon = \mathbf{r}_+$ is equivalent to $(r^i=\mathbf{r}_+\, \&\, \epsilon^i_t \in \boldsymbol{\epsilon}_{+})\, or\, (r^i=\mathbf{r}_-\, \&\, \epsilon^i_t \in \boldsymbol{\epsilon}_{-})$, and $r^i_\epsilon = \mathbf{r}_-$ is equivalent to $(r^i=\mathbf{r}_+\, \&\, \epsilon^i_t \in \boldsymbol{\epsilon}_{-})\, or\, (r^i=\mathbf{r}_-\, \&\, \epsilon^i_t \in \boldsymbol{\epsilon}_{+})$.

Based on our defined surrogate reward function $\hat{\mathcal{R}}_i(s_t, \boldsymbol{a}_t, r^i, \epsilon^i_t)$, we come to the following lemma:

\begin{lemma}
We set true reward $r^i=\mathcal{R}_i(s_t, \boldsymbol{a}_t)$, where $r^i \in \{\mathbf{r}_+, \mathbf{r}_-\}$. For any value of $r^i$, with noise term $\epsilon^i_t$, we have: $\mathbb{E}_{\epsilon^i_t}[\hat{\mathcal{R}}_i(s_t, \boldsymbol{a}_t, r^i, \epsilon^i_t)]=\mathcal{R}_i(s_t, \boldsymbol{a}_t)$ in non-idealistic communication setting.
\label{lemmaTwo}
\end{lemma}

\textit{Proof of Lemma \ref{lemmaTwo}}. In the following proof, we compute the expectation of surrogate rewards under associated probabilities when the true reward is specified. We prove that under any value of the true reward $r^i=\mathcal{R}_i(s_t, \boldsymbol{a}_t)$, the expectation of surrogate reward $\hat{\mathcal{R}}_i(s_t, \boldsymbol{a}_t, r^i, \epsilon^i_t)$ equals to the true reward $\mathcal{R}_i(s_t, \boldsymbol{a}_t)$.

When true reward $r^i=\mathbf{r}_+$, based on the definition of the surrogate reward in Equation \ref{eq:surroRewards}, the expected values of surrogate rewards under noise will be:
\begin{subequations}
\begin{align}
& \mathbb{E}_{\epsilon^i_t}[\hat{\mathcal{R}}_i(s_t, \boldsymbol{a}_t, r^i=\mathbf{r}_+, \epsilon^i_t)] = (1-e) \cdot \dfrac{(1 - e)\mathbf{r}_+ \;-\; e \,\mathbf{r}_-}{1 -2e} + e \cdot \dfrac{(1 - e)\mathbf{r}_- \;-\; e \,\mathbf{r}_+}{1 -2e} \\
&= \dfrac{(1 - e)(1-e)\,\mathbf{r}_+ \;-\; e(1-e)\,\mathbf{r}_-+e(1-e)\mathbf{r}_--e^2\mathbf{r}_+}{1 - 2e} \\
&= \dfrac{(1 - 2e)\mathbf{r}_+}{1 -2e} \\
&= \mathbf{r}_+ \\
&= r^i = \mathcal{R}_i(s_t, \boldsymbol{a}_t)
\end{align}
\label{eq:calRewards}
\end{subequations}
where the value in line \ref{eq:calRewards}a comes from the surrogate rewards multiplied with the probabilities $p(\boldsymbol{\epsilon}_+)$ and $p(\boldsymbol{\epsilon}_-)$. When true reward $r^i=\mathbf{r}_-$, it also verifies that $\mathbb{E}_{\epsilon^i_t}[\hat{\mathcal{R}}_i(s_t, \boldsymbol{a}_t, r^i=\mathbf{r}_-, \epsilon^i_t)]=\mathbf{r}_-$. So we have $\mathbb{E}_{\epsilon^i_t}[\hat{\mathcal{R}}_i(s_t, \boldsymbol{a}_t, r^i, \epsilon^i_t)]=\mathcal{R}_i(s_t, \boldsymbol{a}_t)$ for any value of $r^i$, which completes the proof.


We list the definition of important symbols used in the proof of Lemma 3 and the following theorem for better reading:
\begin{itemize}
    \item $\mathbf{R}=\{\mathbf{r}_+, \mathbf{r}_-\}=\{\mathbf{r}^i_n\}_{n \in \{+, -\}}$ is the set of true rewards of agent $i$ where $\mathbf{r}^i_+$ represents success and $\mathbf{r}^i_-$ represents failure. Due to shared rewards, we have $\mathbf{r}^i_+=\mathbf{r}_+$ and $\mathbf{r}^i_-=\mathbf{r}_-$. Note that we express $r^i_t \in \mathbf{R}$ for reward variable at time step $t$.
    \item $\hat{\mathbf{R}}_n=\{\hat{\mathbf{r}}^i_{n, l}\}_{l \in \{+,-\}}$ is the set of surrogate reward of agent $i$ when true reward is $\mathbf{r}^i_n$. In the set $\hat{\mathbf{R}}_n$, $\hat{\mathbf{r}}^i_{n, +}$ is the surrogate reward if noisy term $\epsilon^i_t \in \boldsymbol{\epsilon}_+$ and given true reward $\mathbf{r}^i_n$. And, $\hat{\mathbf{r}}^i_{n, -}$ is the surrogate reward if noisy term $\epsilon^i_t \in \boldsymbol{\epsilon}_-$ and given true reward $r^i_n$.
\end{itemize}

Similar to Equation \ref{eq:noisyQfunction}, we define a surrogate Q-function $\hat{Q}^{\pi}_i(\boldsymbol{h}, \boldsymbol{a}, \epsilon_i)$ at time step $t$ as:
\begin{equation}
\begin{aligned}
& \hat{Q}^{\pi}_{i,t}(\boldsymbol{h}, \boldsymbol{a}, \epsilon_i) = \mathbb{E}_{s_{t+k},\boldsymbol{a}_{t+k}, r^i_{t+k}, \epsilon^i_{t+k}}[\sum_{k=0}^{T}\gamma^k \hat{\mathcal{R}}_i(s_{t+k}, \boldsymbol{a}_{t+k}, r^i_{t+k}, \epsilon^i_{t+k}) | \boldsymbol{h}_t=\boldsymbol{h}, \boldsymbol{a}_t=\boldsymbol{a}, \epsilon^i_t=\epsilon_i] \\
&= \sum_{s_t, s_{t+1} \in \mathcal{S}} \sum_{r^i \in \mathbf{R}} p(s_t, s_{t+1}, r^i|\boldsymbol{h}, \boldsymbol{a}) \Big(\hat{\mathcal{R}}_i(s_t, \boldsymbol{a}, r^i, \epsilon_i) + \\
&\gamma \sum_{\boldsymbol{a}_{t+1} \in \boldsymbol{\mathcal{A}}, \boldsymbol{h}_{t+1} \in \boldsymbol{H}} p(\boldsymbol{h}_{t+1}|s_{t+1})\boldsymbol{\pi}(\boldsymbol{a}_{t+1}|\boldsymbol{h}_{t+1}) \sum_{\epsilon^i_{t+1}} p(\epsilon^i_{t+1}) \hat{Q}^{\pi}_{i,t+1}(\boldsymbol{h}_{t+1}, \boldsymbol{a}_{t+1}, \epsilon^i_{t+1}) \Big)
\end{aligned}
\label{eq:surrogateQ}
\end{equation}
where surrogate rewards are accumulated when observing history $\boldsymbol{h}$, the first joint action to be $\boldsymbol{a}$, and the first noise to be $\epsilon_i$.

According to Lemma \ref{lemmaTwo}, the surrogate rewards and true rewards are related at every time step. Based on this, we would like to investigate how the accumulation of rewards in Q-functions relates to each other. Since individual but centralized Q-function equals to centralized Q-function with shared rewards, we have the following lemma:
\begin{lemma}
Given $\mathcal{R}_i(s_t, \boldsymbol{a}_t)=\mathbb{E}_{\epsilon^i_t}[\hat{\mathcal{R}}_i(s_t, \boldsymbol{a}_t, r^i, \epsilon^i_t)]$, the centralized Q-function equals to the expectation of surrogate Q-function in non-idealistic communication setting: $Q^{\pi}(\boldsymbol{h}, \boldsymbol{a})=\mathbb{E}_{\epsilon_i}[\hat{Q}^{\pi}_i(\boldsymbol{h}, \boldsymbol{a}, \epsilon_i)]$.
\label{lemmaThree}
\end{lemma}

\textit{Proof of Lemma \ref{lemmaThree}}. In the following proof, we first relate true rewards and surrogate rewards under transition probabilities. Then we relate individual but centralized Q-function and surrogate Q-function upon the summation over true rewards and surrogate rewards per step, respectively. Finally we achieve the equality between the centralized Q-function and the surrogate Q-function.

According to Lemma \ref{lemmaTwo}, for any true reward $r^i=\mathbf{r}^i_n$, we have:
\begin{equation}
\mathbb{E}_{\epsilon^i_t}[\hat{\mathcal{R}}_i(s_t, \boldsymbol{a}_t, r^i=\mathbf{r}^i_n, \epsilon^i_t)]= p(\boldsymbol{\epsilon}_+) \cdot \hat{\mathbf{r}}^i_{n, +} + p(\boldsymbol{\epsilon}_-) \cdot \hat{\mathbf{r}}^i_{n, -} = \sum_{l \in \{+, -\}} p(\boldsymbol{\epsilon}_l) \hat{\mathbf{r}}^i_{n,l} = \mathbf{r}^i_n
\label{eq:SumSurrRwd}
\end{equation}

We further use $p(s_t, s_{t+1}, \mathbf{r}^i_n | \boldsymbol{h}_t, \boldsymbol{a}_t)$ to represent the probabilities of a certain true reward $\mathbf{r}^i_n$ when transiting from state $s_t$ to the next state $s_{t+1}$ under joint actions $\boldsymbol{a}_t$ given current history $\boldsymbol{h}_t$. The true rewards and surrogate rewards for every time step $t$ satisfy:
\begin{subequations}
\begin{align}
\sum_{s_t, s_{t+1} \in \mathcal{S}} p(s_t, s_{t+1}| \boldsymbol{h}_t,\boldsymbol{a}_t) \, \mathcal{R}_i(s_t, \boldsymbol{a}_t) &= \sum_{s_t, s_{t+1} \in \mathcal{S}} \sum_{n \in \{+, -\}} p(s_t, s_{t+1}, \mathbf{r}^i_n| \boldsymbol{h}_t, \boldsymbol{a}_t) \mathbf{r}^i_n \\
&\stackrel{eq. \ref{eq:SumSurrRwd}}{=} \sum_{s_t, s_{t+1} \in \mathcal{S}} \sum_{n \in \{+, -\}} p(s_t, s_{t+1}, \mathbf{r}^i_n|\boldsymbol{h}_t,\boldsymbol{a}_t) \sum_{l \in \{+,-\}} p(\boldsymbol{\epsilon}_l) \hat{\mathbf{r}}^i_{n,l} \\
&= \sum_{s_t, s_{t+1} \in \mathcal{S}} \sum_{r^i \in \mathbf{R}} p(s_t, s_{t+1}, r^i| \boldsymbol{h}_t,\boldsymbol{a}_t) \sum_{\epsilon^i_t \in \mathcal{E}} p(\epsilon^i_t) \hat{\mathcal{R}}_i(s_t, \boldsymbol{a}_t, r^i, \epsilon^i_t)
\end{align}
\label{eq:trueSurrogate}
\end{subequations}
where $p(s_t, s_{t+1}|\boldsymbol{h}_t,\boldsymbol{a}_t)$ is the transition probabilities giving current history $\boldsymbol{h}_t$ and action $\boldsymbol{a}_t$. Line \ref{eq:trueSurrogate}a sums over all possible values of true rewards multiplied by their probabilities. Line \ref{eq:trueSurrogate}b is due to Equation \ref{eq:SumSurrRwd}. In line \ref{eq:trueSurrogate}c, we achieve the summation over surrogate rewards by replacing the index with reward variable: a) the summation over index $n$ is transformed into the summation over $r^i \in \mathbf{R}$; b) the summation over the set $E=\{\boldsymbol{\epsilon}_l\}_{l \in \{+,-\}}$ is transformed into the summation over all possible $\epsilon^i_t$ as $\boldsymbol{\epsilon}_+ \cup \boldsymbol{\epsilon}_-= \mathcal{E}$.

Equation \ref{eq:trueSurrogate} implies that the true rewards and surrogate rewards under transitions per time step can be equal in expectation, which can lead to the equality between individual but centralized Q-function (accumulating true rewards) and surrogate Q-function (accumulating surrogate rewards). In the following derivations, we first prove that the equality hold at the last time step. Then we can achieve that the equality holds for every time step by induction.

\textbf{\textit{Base Step}:} At terminal time step $T$, we have:
\begin{subequations}
\begin{align}
Q^{\pi}_{i,T}(\boldsymbol{h}_{T}, \boldsymbol{a}_{T}) &= \sum_{s_{T} \in \mathcal{S}} p(s_T, s_{terminal}|\boldsymbol{h}_{T},\boldsymbol{a}_{T}) \mathcal{R}_i(s_{T}, \boldsymbol{a}_{T}) \\
&\stackrel{eq. \ref{eq:trueSurrogate}}{=} \sum_{s_{T} \in \mathcal{S}} \sum_{r^i \in \mathbf{R}} p(s_{T}, s_{terminal}, r^i| \boldsymbol{h}_{T},\boldsymbol{a}_{T}) \sum_{\epsilon^i_{T} \in \mathcal{E}} p(\epsilon^i_{T}) \hat{\mathcal{R}}_i(s_{T}, \boldsymbol{a}_{T}, r^i, \epsilon^i_{T}) \\
&= \sum_{\epsilon^i_{T} \in \mathcal{E}} p(\epsilon^i_{T}) \Big( \sum_{s_{T} \in \mathcal{S}} \sum_{r^i \in \mathbf{R}} p(s_{T}, s_{terminal}, r^i| \boldsymbol{h}_{T},\boldsymbol{a}_{T}) \hat{\mathcal{R}}_i(s_{T}, \boldsymbol{a}_{T}, r^i, \epsilon^i_{T}) \Big) \\
&\stackrel{eq. \ref{eq:surrogateQ}}{=} \sum_{\epsilon^i_{T} \in \mathcal{E}} p(\epsilon^i_{T}) \hat{Q}^{\pi}_{i,T}(\boldsymbol{h}_{T}, \boldsymbol{a}_{T}, \epsilon^i_{T})
\end{align}
\label{eq:lastQvalues}
\end{subequations}
where $s_{terminal}$ is the terminal state. Line \ref{eq:lastQvalues}a is due to the definition of Q-values in the last transitions and we do not use the discounted term. Line \ref{eq:lastQvalues}b is due to Equation \ref{eq:trueSurrogate} by replacing true rewards with the expectation of surrogate rewards. Line \ref{eq:lastQvalues}c holds since the summation over noisy term $\epsilon^i_{T}$ at line \ref{eq:lastQvalues}b is factored out over the summation. The equation between lines \ref{eq:lastQvalues}c and \ref{eq:lastQvalues}d holds due to Equation \ref{eq:surrogateQ}. Note we do not use the discounted term in \ref{eq:surrogateQ} because $T$ is the terminal step. 

\textbf{\textit{Induction step}:} We now assume that $Q^{\pi}_{i,t+1}(\boldsymbol{h}_{t+1}, \boldsymbol{a}_{t+1})=\sum_{\epsilon^i_{t+1} \in \mathcal{E}} p(\epsilon^i_{t+1}) \hat{Q}^{\pi}_{i,t+1}(\boldsymbol{h}_{t+1}, \boldsymbol{a}_{t+1}, \epsilon^i_{t+1})$ holds for time step $t+1<T$, and prove that this holds also for time step $t$, i.e., we prove that $Q^{\pi}_{i,t}(\boldsymbol{h}_{t}, \boldsymbol{a}_{t})=\sum_{\epsilon^i_{t} \in \mathcal{E}} p(\epsilon^i_{t}) \hat{Q}^{\pi}_{i,t}(\boldsymbol{h}_{t}, \boldsymbol{a}_{t}, \epsilon^i_{t})$.

\begin{subequations}
\begin{align}
&Q^{\pi}_{i,t}(\boldsymbol{h}_{t}, \boldsymbol{a}_{t}) \\
&= \sum_{s_{t}, s_{t+1} \in \mathcal{S}} p(s_{t}, s_{t+1}|\boldsymbol{h}_{t},\boldsymbol{a}_{t}) \Big( \mathcal{R}_i(s_{t}, \boldsymbol{a}_{t}) + \gamma \sum_{\boldsymbol{a}_{t+1} \in \boldsymbol{\mathcal{A}}, \boldsymbol{h}_{t+1} \in \boldsymbol{H}} p(\boldsymbol{h}_{t+1}|s_{t+1})\boldsymbol{\pi}(\boldsymbol{a}_{t+1}|\boldsymbol{h}_{t+1}) \underbrace{Q^{\pi}_{i,t+1}(\boldsymbol{h}_{t+1}, \boldsymbol{a}_{t+1})}_{\stackrel{ind.step}{=}\sum_{\epsilon^i_{t+1} \in \mathcal{E}} p(\epsilon^i_{t+1}) \hat{Q}^{\pi}_{i,t+1}(\boldsymbol{h}_{t+1}, \boldsymbol{a}_{t+1}, \epsilon^i_{t+1})} \Big) \\
&\stackrel{eq. \ref{eq:trueSurrogate}}{=} \sum_{s_{t}, s_{t+1} \in \mathcal{S}} \sum_{r^i \in \mathbf{R}} p(s_{t}, s_{t+1}, r^i|\boldsymbol{h}_{t},\boldsymbol{a}_{t}) \sum_{\epsilon^i_{t} \in \mathcal{E}} p(\epsilon^i_{t}) \Big( \hat{\mathcal{R}}_i(s_{t}, \boldsymbol{a}_{t}, r^i, \epsilon^i_{t}) + \\
&\gamma \sum_{\boldsymbol{a}_{t+1} \in \boldsymbol{\mathcal{A}}, \boldsymbol{h}_{t+1} \in \boldsymbol{H}} p(\boldsymbol{h}_{t+1}|s_{t+1})\boldsymbol{\pi}(\boldsymbol{a}_{t+1}|\boldsymbol{h}_{t+1})\sum_{\epsilon^i_{t+1} \in \mathcal{E}} p(\epsilon^i_{t+1}) \hat{Q}^{\pi}_{i,t+1}(\boldsymbol{h}_{t+1}, \boldsymbol{a}_{t+1}, \epsilon^i_{t+1}) \Big) \\
&= \sum_{\epsilon^i_{t} \in \mathcal{E}} p(\epsilon^i_{t}) \Big( \sum_{s_{t}, s_{t+1} \in \mathcal{S}} \sum_{r^i \in \mathbf{R}} p(s_{t}, s_{t+1}, r^i|\boldsymbol{h}_{t},\boldsymbol{a}_{t}) \Big( \hat{\mathcal{R}}_i(s_{t}, \boldsymbol{a}_{t}, r^i, \epsilon^i_{t}) + \\
&\gamma \sum_{\boldsymbol{a}_{t+1} \in \boldsymbol{\mathcal{A}}, \boldsymbol{h}_{t+1} \in \boldsymbol{H}} p(\boldsymbol{h}_{t+1}|s_{t+1})\boldsymbol{\pi}(\boldsymbol{a}_{t+1}|\boldsymbol{h}_{t+1})  \sum_{\epsilon^i_{t+1} \in \mathcal{E}} p(\epsilon^i_{t+1}) \hat{Q}^{\pi}_{i,t+1}(\boldsymbol{h}_{t+1}, \boldsymbol{a}_{t+1}, \epsilon^i_{t+1}) \Big) \Big) \\
&\stackrel{eq. \ref{eq:surrogateQ}}{=} \sum_{\epsilon^i_{t} \in \mathcal{E}} p(\epsilon^i_{t}) \hat{Q}^{\pi}_{i,t}(\boldsymbol{h}_{t}, \boldsymbol{a}_{t}, \epsilon^i_{t})
\end{align}
\label{eq:lastTwoQvalues}
\end{subequations}
where derivation \ref{eq:lastTwoQvalues}b is due to the definition of $Q^{\pi}_{i,t}(\boldsymbol{h}_{t}, \boldsymbol{a}_{t})$. Moreover, in derivation \ref{eq:lastTwoQvalues}b, we use the assumption $Q^{\pi}_{i,t+1}(\boldsymbol{h}_{t+1}, \boldsymbol{a}_{t+1})=\sum_{\epsilon^i_{t+1} \in \mathcal{E}} p(\epsilon^i_{t+1}) \hat{Q}^{\pi}_{i,t+1}(\boldsymbol{h}_{t+1}, \boldsymbol{a}_{t+1}, \epsilon^i_{t+1})$, leading to the discounted term in \ref{eq:lastTwoQvalues}d. We also use Equation \ref{eq:trueSurrogate} to replace true rewards with the expectation of surrogate rewards in derivation \ref{eq:lastTwoQvalues}c. Derivations \ref{eq:lastTwoQvalues}e-\ref{eq:lastTwoQvalues}f holds since the summation over
noisy term $\epsilon^i_t$ at derivation \ref{eq:lastTwoQvalues}c is factored out over the summation. The equation between derivations \ref{eq:lastTwoQvalues}e-\ref{eq:lastTwoQvalues}f and \ref{eq:lastTwoQvalues}g holds due to Equation \ref{eq:surrogateQ} (the surrogate Q-function). Then, we achieve the equality $Q^{\pi}_{i,t}(\boldsymbol{h}_{t}, \boldsymbol{a}_{t})=\sum_{\epsilon^i_{t} \in \mathcal{E}} p(\epsilon^i_{t}) \hat{Q}^{\pi}_{i,t}(\boldsymbol{h}_{t}, \boldsymbol{a}_{t}, \epsilon^i_{t})$. By induction, we conclude that $Q^{\pi}_i(\boldsymbol{h}, \boldsymbol{a})=\sum_{\epsilon_i \in \mathcal{E}} p(\epsilon_i) \hat{Q}^{\pi}_{i}(\boldsymbol{h}, \boldsymbol{a}, \epsilon_i)$ hold for any time step (and we drop the time step). Due to shared rewards and the definition of expectation we have $Q^{\pi}_i(\boldsymbol{h}, \boldsymbol{a})=Q^{\pi}(\boldsymbol{h}, \boldsymbol{a})$ and $\sum_{\epsilon_i \in \mathcal{E}} p(\epsilon^i) \hat{Q}^{\pi}_{i}(\boldsymbol{h}, \boldsymbol{a}, \epsilon^i)=\mathbb{E}_{\epsilon_i} [\hat{Q}^{\pi}_i(\boldsymbol{h}, \boldsymbol{a}, \epsilon_i)]$. Therefore, we complete the proof that $Q^{\pi}(\boldsymbol{h}, \boldsymbol{a})=\mathbb{E}_{\epsilon_i} [\hat{Q}^{\pi}_i(\boldsymbol{h}, \boldsymbol{a}, \epsilon_i)]$.

\subsection{Proof of Theorem 2}
\label{app:varianceNoise}

We use Q-functions $\hat{Q}^{\pi}_i(\boldsymbol{h}, \boldsymbol{a}, \epsilon_i)$ and $Q^{\pi}(\boldsymbol{h}, \boldsymbol{a})$ as critics for decentralized actors $\pi_i(a_i|h_i, \theta_i)$. Then, we have the following policy gradients:
$$
g^i_{DCCDA-noise} = \mathbb{E}_{\boldsymbol{h},\boldsymbol{a}, \epsilon_i}[\hat{Q}^{\pi}_i(\boldsymbol{h},\boldsymbol{a}, \epsilon_i)\nabla_{\theta_i}\log \pi_i(a_i|h_i, \theta_i)]
$$
$$
g^i_{CTDE} =\mathbb{E}_{\boldsymbol{h},\boldsymbol{a}}[Q^{\pi}(\boldsymbol{h},\boldsymbol{a})\nabla_{\theta_i}\log \pi_i(a_i|h_i, \theta_i)]
$$
where $g^i_{DCCDA-noise}$ is the noise version of DCCDA and the noise in communication is lifted to a surrogate Q-function.

Based on $g^i_{DCCDA-noise}$ and $g^i_{CTDE}$, we have the following theorem, which follows similar procedures as in the proof of idealistic communication setting:
\setcounter{theorem}{1}
\begin{theorem}
The noisy version of DCCDA sample gradient has a variance greater or equal than that of the CTDE sample gradient in non-idealistic communication setting: $Var(\hat{g}^i_{DCCDA-noise}) \geq Var(\hat{g}^i_{CTDE})$.
\label{theoremNoiseDCCDA}
\end{theorem}

\textit{Proof of Theorem \ref{theoremNoiseDCCDA}}. In the following proof, we first check the relation between the two expected gradients $g^i_{DCCDA-noise}$ and $g^i_{CTDE}$, which can greatly simplify the later variance analysis. Then we compare the variance of gradients $g^i_{DCCDA-noise}$ and $g^i_{CTDE}$.

Based on Lemma \ref{lemmaThree}, as $Q^{\pi}(\boldsymbol{h}, \boldsymbol{a})=\mathbb{E}_{\epsilon_i}[Q^{\pi}_i(\boldsymbol{h}, \boldsymbol{a}, \epsilon_i)]$:
\begin{subequations}
\begin{align}
g^i_{DCCDA-noise} &= \mathbb{E}_{\boldsymbol{h},\boldsymbol{a},\epsilon_i}[Q^{\pi}_i(\boldsymbol{h},\boldsymbol{a}, \epsilon_i)\nabla_{\theta_i}\log \pi_i(a_i|h_i, \theta_i)] \\
&= \mathbb{E}_{\boldsymbol{h},\boldsymbol{a}}[\mathbb{E}_{\epsilon_i}[Q^{\pi}_i(\boldsymbol{h},\boldsymbol{a}, \epsilon_i)\nabla_{\theta_i}\log \pi_i(a_i|h_i, \theta_i)] \\
&\stackrel{lem. \ref{lemmaThree}}{=}  \mathbb{E}_{\boldsymbol{h},\boldsymbol{a}}[Q^{\pi}(\boldsymbol{h}, \boldsymbol{a})\nabla_{\theta_i}\log \pi_i(a_i|h_i, \theta_i)] \\
&= g^i_{CTDE}
\end{align}
\label{eq:biasNoise}
\end{subequations}
where line \ref{eq:biasNoise}c is due to Lemma \ref{lemmaThree}. Note that the noise term is not included in actors as actors do not communicate.

Based on derivation \ref{eq:biasNoise}a-\ref{eq:biasNoise}d, we come to the following proof of Theorem \ref{theoremNoiseDCCDA}. Note that $\hat{g}^i_{CTDE}$ is used to denote $Q^{\pi}(\boldsymbol{h}, \boldsymbol{a})\nabla_{\theta_i}\log \pi_i(a_i|h_i, \theta_i)$ and $\hat{g}^i_{DCCDA-noise}$ is used to denote $Q^{\pi}_i(\boldsymbol{h}, \boldsymbol{a}, \epsilon_i)\nabla_{\theta_i}\log \pi_i(a_i|h_i, \theta_i)$. In the following equation, in order to simplify the notation, we denote $S=\nabla_{\theta_i}\log \pi_i(a_i|h_i, \theta_i)^T\nabla_{\theta_i}\log \pi_i(a_i|h_i, \theta_i)$, which is the inner product of $\nabla_{\theta_i}\log \pi_i(a_i|h_i, \theta_i)$ (due to the square of it):

\begin{subequations}
\begin{align}
&Var(\hat{g}^i_{DCCDA-noise}) - Var(\hat{g}^i_{CTDE}) \\
&\stackrel{def.}{=} \left(\mathbb{E}_{\boldsymbol{h},\boldsymbol{a},\epsilon_i}[(\hat{g}^i_{DCCDA-noise})^2] - \left (\mathbb{E}_{\boldsymbol{h},\boldsymbol{a},\epsilon_i}[\hat{g}^i_{DCCDA-noise}] \right)^2 \right) - \left(\mathbb{E}_{\boldsymbol{h},\boldsymbol{a}}[(\hat{g}^i_{CTDE})^2] - \left (\mathbb{E}_{\boldsymbol{h},\boldsymbol{a}}[\hat{g}^i_{CTDE}] \right)^2 \right)  \\
&= \left(\mathbb{E}_{\boldsymbol{h},\boldsymbol{a},\epsilon_i}[(\hat{g}^i_{DCCDA-noise})^2] - \mathbb{E}_{\boldsymbol{h},\boldsymbol{a}}[(\hat{g}^i_{CTDE})^2] \right) - \left( \left( \mathbb{E}_{\boldsymbol{h},\boldsymbol{a},\epsilon_i}[\hat{g}^i_{DCCDA-noise}] \right)^2 - \left (\mathbb{E}_{\boldsymbol{h},\boldsymbol{a}}[\hat{g}^i_{CTDE}] \right)^2 \right)  \\
&\stackrel{eq. \ref{eq:biasNoise}}{=} \left(\mathbb{E}_{\boldsymbol{h},\boldsymbol{a},\epsilon_i}[(\hat{g}^i_{DCCDA-noise})^2] - \mathbb{E}_{\boldsymbol{h},\boldsymbol{a}}[(\hat{g}^i_{CTDE})^2] \right) - \underbrace{ \left((g^i_{DCCDA-noise})^2 - (g^i_{CTDE})^2\right)}_{=0}  \\
&= \mathbb{E}_{\boldsymbol{h},\boldsymbol{a},\epsilon_i}[(\hat{g}^i_{DCCDA-noise})^2] - \mathbb{E}_{\boldsymbol{h},\boldsymbol{a}}[(\hat{g}^i_{CTDE})^2]  \\
&\stackrel{def.}{=}  \mathbb{E}_{\boldsymbol{h},\boldsymbol{a},\epsilon_i}[(Q^{\pi}_i(\boldsymbol{h},\boldsymbol{a}, \epsilon_i)\nabla_{\theta_i}\log \pi_i(a_i|h_i, \theta_i))^2] - \mathbb{E}_{\boldsymbol{h},\boldsymbol{a}}[(Q^{\pi}(\boldsymbol{h}, \boldsymbol{a})\nabla_{\theta_i}\log \pi_i(a_i|h_i, \theta_i))^2] \\
&= \mathbb{E}_{\boldsymbol{h},\boldsymbol{a},\epsilon_i}[Q^{\pi}_i(\boldsymbol{h},\boldsymbol{a}, \epsilon_i)^2S] - \mathbb{E}_{\boldsymbol{h},\boldsymbol{a}}[Q^{\pi}(\boldsymbol{h}, \boldsymbol{a})^2S] \\
&= \mathbb{E}_{\boldsymbol{h},\boldsymbol{a}}[\mathbb{E}_{\epsilon_i}[Q^{\pi}_i(\boldsymbol{h},\boldsymbol{a}, \epsilon_i)^2S]] - \mathbb{E}_{\boldsymbol{h},\boldsymbol{a}}[Q^{\pi}(\boldsymbol{h}, \boldsymbol{a})^2S] \\
&= \mathbb{E}_{\boldsymbol{h},\boldsymbol{a}}[\mathbb{E}_{\epsilon_i}[Q^{\pi}_i(\boldsymbol{h},\boldsymbol{a}, \epsilon_i)^2S] - Q^{\pi}(\boldsymbol{h}, \boldsymbol{a})^2S] \\
&\stackrel{lem. \ref{lemmaThree}}{=}  \mathbb{E}_{\boldsymbol{h},\boldsymbol{a}}[\underbrace{\left (\mathbb{E}_{\epsilon_i}[Q^{\pi}_i(\boldsymbol{h},\boldsymbol{a}, \epsilon_i)^2] - (\mathbb{E}_{\epsilon_i}[Q^{\pi}_i(\boldsymbol{h},\boldsymbol{a}, \epsilon_i)])^2 \right )}_{u} S] \\
&\ge 0
\end{align}
\label{eq:varianceNoise}
\end{subequations}
where line \ref{eq:varianceNoise}b follows the definition of variance. Line \ref{eq:varianceNoise}d is due to $g^i_{DCCDA-noise}=g^i_{CTDE}$ according to derivation \ref{eq:biasNoise}a-\ref{eq:biasNoise}d. Line \ref{eq:varianceNoise}j is due to Lemma \ref{lemmaThree}. The final inequality in line \ref{eq:varianceNoise}k follows because $u \ge 0$ by Jensen’s inequality: $\mathbb{E}[X^2] \ge (\mathbb{E}[X])^2$, and $S$ is the inner product of a vector itself and therefore non-negative. Therefore, the noise version of DCCDA sample gradient has a variance greater or equal than that of the CTDE sample gradient, i.e., $Var(\hat{g}^i_{DCCDA-noise}) \geq Var(\hat{g}^i_{CTDE})$, which completes the proof.

\section{Proofs of the Theoretical Results of the Optimal Baseline}
\label{sec:proofsOptimalBaseline}

\subsection{Proof of Theorem 3}
\label{app:baseline}

In this section, we derive the optimal message-dependent baseline. The computation of the message-dependent baseline use each agent's critic as well as encoded messages, where messages can either be noise-free or noisy. Therefore, the message-dependent baseline can be used in both idealistic or non-idealistic communication setting. We have the following theorem:

\begin{theorem}
The optimal message-dependent baseline for DCCDA-OB gradient estimator is,
\begin{equation}
\begin{split}
b_i^*(h_i, m_{-i}) = \frac{\mathbb{E}_{a_i}[Q_i(h_i, a_i, m_{-i}) S]}{\mathbb{E}_{a_i}[S]}
\end{split}
\label{eq:obFinal}
\end{equation}
where $S=\nabla_{\theta_i}\log \pi_i(a_i|h_i, \theta_i)^T\nabla_{\theta_i}\log \pi_i(a_i|h_i, \theta_i)$.
\label{theoremThree}
\end{theorem}

\textit{Proof of Theorem \ref{theoremThree}.} To prove the theorem, we firstly prove that the message-dependent baseline does not change the policy gradients $g^i_{DCCDA}$ (i.e., $g^i_{DCCDA-OB}=g^i_{DCCDA}$), which will be used to simplify the variance measurement of $g^i_{DCCDA-OB}$. Note that the noisy version of the DCDDA policy gradients, $g^i_{DCCDA-noise}$, also applies to the following derivations, where we can replace $m_{-i}$ with $<h_{-i}, a_{-i}, \epsilon_i>$. 

Therefore, we have,
\begin{equation}
\begin{split}
&\mathbb{E}_{\boldsymbol{h},\boldsymbol{a},\boldsymbol{m}}[b_i(h_i, m_{-i}) \nabla_{\theta_i}\log \pi_i(a_i|h_i, \theta_i)] = \mathbb{E}_{\boldsymbol{h},a_{-i},\boldsymbol{m}}[b_i(h_i, m_{-i}) \mathbb{E}_{a_i}[\nabla_{\theta_i}\log \pi_i(a_i|h_i, \theta_i)]] \\
&= \mathbb{E}_{\boldsymbol{h},a_{-i},\boldsymbol{m}}[b_i(h_i, m_{-i}) \sum_{a_i}\pi_i(a_i|h_i, \theta_i)\nabla_{\theta_i}\log \pi_i(a_i|h_i, \theta_i)] \\
&= \mathbb{E}_{\boldsymbol{h},a_{-i},\boldsymbol{m}}[b_i(h_i, m_{-i}) \sum_{a_i} \cancel{\pi_i(a_i|h_i, \theta_i)}\frac{\nabla_{\theta_i} \pi_i(a_i|h_i, \theta_i)}{\cancel{\pi_i(a_i|h_i, \theta_i)}}] \\
&= \mathbb{E}_{\boldsymbol{h},a_{-i},\boldsymbol{m}}[b_i(h_i, m_{-i}) \sum_{a_i} \nabla_{\theta_i} \pi_i(a_i|h_i, \theta_i)] = \mathbb{E}_{\boldsymbol{h},a_{-i},\boldsymbol{m}}[b_i(h_i, m_{-i}) \nabla_{\theta_i}1] = 0
\end{split}
\label{eq:baseline}
\end{equation}
where the second to the last line is due to the sum of all probabilities over agent $i$'s action is 1. By integrating Equation \ref{eq:baseline} into the policy gradient $g^i_{DCCDA-OB}=\mathbb{E}_{\boldsymbol{h},\boldsymbol{a},\boldsymbol{m}}[\left ( Q_i(h_i, a_i, m_{-i}) - b_i(h_i, m_{-i}) \right ) \nabla_{\theta_i}\log \pi_i(a_i|h_i, \theta_i)]$, we obtain the policy gradient $g^i_{DCCDA}=\mathbb{E}_{\boldsymbol{h},\boldsymbol{a},\boldsymbol{m}}[Q^{\pi}_i(h_i, a_i, m_{-i})\nabla_{\theta_i}\log \pi_i(a_i|h_i, \theta_i)]$. Therefore, we have $g^i_{DCCDA-OB}=g^i_{DCCDA}$. The equality shows that the baseline $b_i(h_i, m_{-i})$ does not introduce bias to $g^i_{DCCDA}$ in expectation. Nevertheless, $g^i_{DCCDA}$ and $g^i_{DCCDA-OB}$ may have different variance properties. We first derive the variance of the DCCDA policy gradient estimate with the message-dependent baseline. We simplify the expression by using $b_i$ to denote the baseline $b_i(h_i, m_{-i})$. Therefore, by the definition of variance, we have:
\begin{equation}
\begin{split}
& Var(\hat{g}^i_{DCCDA-OB}) = \mathbb{E}_{\boldsymbol{h},\boldsymbol{a},\boldsymbol{m}}[(\hat{g}^i_{DCCDA-OB})^2] - \left (\mathbb{E}_{\boldsymbol{h},\boldsymbol{a},\boldsymbol{m}}[\hat{g}^i_{DCCDA-OB}] \right)^2 \\
&= \mathbb{E}_{\boldsymbol{h},\boldsymbol{a},\boldsymbol{m}}[\left ( (Q_i(h_i, a_i, m_{-i}) - b_i) \nabla_{\theta_i}\log \pi_i(a_i|h_i, \theta_i) \right )^2] - \left (\mathbb{E}_{\boldsymbol{h},\boldsymbol{a},\boldsymbol{m}}[(Q_i(h_i, a_i, m_{-i}) - b_i) \nabla_{\theta_i}\log \pi_i(a_i|h_i, \theta_i)] \right)^2 \\
&= \mathbb{E}_{\boldsymbol{h},\boldsymbol{a},\boldsymbol{m}}[\left ( (Q_i(h_i, a_i, m_{-i}) - b_i)) \nabla_{\theta_i}\log \pi_i(a_i|h_i, \theta_i) \right )^2] - \left (\mathbb{E}_{\boldsymbol{h},\boldsymbol{a},\boldsymbol{m}}[Q_i(h_i, a_i, m_{-i}) \nabla_{\theta_i}\log \pi_i(a_i|h_i, \theta_i)] \right)^2 \\
\end{split}
\label{eq:baselineVar}
\end{equation}
where line 2 follows the definition of $\hat{g}^i_{DCCDA-OB}$ and the last line is due to the fact $\mathbb{E}_{\boldsymbol{h},\boldsymbol{a},\boldsymbol{m}}[b_i(h_i, m_{-i}) \nabla_{\theta_i}\log \pi_i(a_i|h_i, \theta_i)]$ is zero according to Equation \ref{eq:baseline}. Note that $S=\nabla_{\theta_i}\log \pi_i(a_i|h_i, \theta_i)^T\nabla_{\theta_i}\log \pi_i(a_i|h_i, \theta_i)$ is used to denote the inner product of the gradient $\nabla_{\theta_i}\log \pi_i(a_i|h_i, \theta_i)$ for notation simplification. We seek the optimal baseline that would minimize this variance by setting the derivatives with respect to the baseline $b_i$ to be zero:
\begin{equation}
\begin{split}
&\frac{\partial}{\partial b_i}\left[ Var(\hat{g}^i_{DCCDA-OB}) \right] \\
&= \frac{\partial}{\partial b_i}\left[ \mathbb{E}_{\boldsymbol{h},\boldsymbol{a},\boldsymbol{m}}[\left ( Q_i(h_i, a_i, m_{-i}) - b_i \right )^2 S ] \right] + \frac{\partial}{\partial b_i}\left[(\mathbb{E}_{\boldsymbol{h},\boldsymbol{a},\boldsymbol{m}}[Q_i(h_i, a_i, m_{-i}) \nabla_{\theta_i}\log \pi_i(a_i|h_i, \theta_i)] \right)^2] \\
&= \frac{\partial}{\partial b_i}\left[ \mathbb{E}_{\boldsymbol{h},\boldsymbol{a},\boldsymbol{m}}[\left ( Q_i(h_i, a_i, m_{-i}) - b_i \right )^2 S ] \right] \\
&= 0
\end{split}
\label{eq:VarZero}
\end{equation}
where the term in the second line does not depend on the baseline $b_i$, and therefore the derivative is 0. By writing out the term in brackets from the second to the last line in Equation \ref{eq:VarZero}, we have:
\begin{equation}
\begin{split}
& \mathbb{E}_{\boldsymbol{h},\boldsymbol{a},\boldsymbol{m}}[\left ( Q_i(h_i, a_i, m_{-i}) - b_i \right )^2 S ] \\
&= \mathbb{E}_{\boldsymbol{h},\boldsymbol{a},\boldsymbol{m}}[\left ( Q_i(h_i, a_i, m_{-i})^2 - 2b_iQ_i(h_i, a_i, m_{-i}) + b_i^2 \right ) S] \\
&= \mathbb{E}_{\boldsymbol{h},\boldsymbol{a},\boldsymbol{m}}[Q_i(h_i, a_i, m_{-i})^2 S - 2b_iQ_i(h_i, a_i, m_{-i})S + b_i^2S] \\
&= \mathbb{E}_{\boldsymbol{h},\boldsymbol{a},\boldsymbol{m}}[Q_i(h_i, a_i, m_{-i})^2 S] + \mathbb{E}_{\boldsymbol{h}, a_{-i},\boldsymbol{m}}[- 2b_i\mathbb{E}_{a_i}[Q_i(h_i, a_i, m_{-i}) S] + b_i^2 \mathbb{E}_{a_i}[S]]
\end{split}
\label{eq:baselineVarSim}
\end{equation}
where the last line is because $b_i$ does not depend on actions $a_i$. By integrating Equation \ref{eq:baselineVarSim} into Equation \ref{eq:VarZero} we have:
\begin{equation}
\frac{\partial}{\partial b_i}\left[ Var(\hat{g}^i_{DCCDA-OB}) \right] = \mathbb{E}_{\boldsymbol{h}, a_{-i},\boldsymbol{m}}[-2\mathbb{E}_{a_i}[Q_i(h_i, a_i, m_{-i}) S] + 2 b_i(h_i, m_{-i}) \mathbb{E}_{a_i}[S]] = 0
\label{eq:optimalB}
\end{equation}

Therefore, the optimal baseline is,
\begin{equation}
\begin{split}
b_i^*(h_i, m_{-i}) = \frac{\mathbb{E}_{a_i}[Q_i(h_i, a_i, m_{-i}) S]}{\mathbb{E}_{a_i}[S]}
\end{split}
\end{equation}
where the expectation enumerates all possible actions of agent $i$. Then, we complete the proof.

\subsection{Proof of Corollary 1}
\label{app:reducedVar}

In this section, we prove that with the optimal message-dependent baseline, the variance of DCCDA policy gradient is reduced. Note that Corollary \ref{corollaryOne} also holds for non-idealistic communication setting, i.e., $Var(\hat{g}^i_{DCCDA-OB}) \leq Var(\hat{g}^i_{DCCDA-noise})$, where we replace message $m_{-i}$ with $<h_{-i}, a_{-i}, \epsilon_i>$ and follow the same derivations. 

We have the following corollary based on Theorem \ref{theoremThree}:

\setcounter{theorem}{0}
\begin{corollary}
The variance of DCCDA policy gradients is reduced with the optimal message-dependent baseline: $Var(\hat{g}^i_{DCCDA-OB}) \leq Var(\hat{g}^i_{DCCDA})$.
\label{corollaryOne}
\end{corollary}

\textit{Proof of Corollary \ref{corollaryOne}.} The proof is achieved by integrating the optimal baseline $b_i^*$ derived from Theorem \ref{theoremThree} back to Equation \ref{eq:baselineVar}, where $Var(\hat{g}^i_{DCCDA}) =\mathbb{E}_{\boldsymbol{h},\boldsymbol{a},\boldsymbol{m}}[(Q_i(h_i, a_i, m_{-i})\nabla_{\theta_i}\log \pi_i(a_i|h_i, \theta_i))^2] - (\mathbb{E}_{\boldsymbol{h},\boldsymbol{a},\boldsymbol{m}}[Q_i(h_i, a_i, m_{-i}) \nabla_{\theta_i}\log \pi_i(a_i|h_i, \theta_i)])^2$ by definition. Then we have:
\begin{equation}
\begin{split}
Var(\hat{g}^i_{DCCDA-OB}) = Var(\hat{g}^i_{DCCDA}) - \mathbb{E}_{\boldsymbol{h}, a_{-i},\boldsymbol{m}}[\frac{(\mathbb{E}_{a_i}[Q_i(h_i, a_i, m_{-i}) S])^2}{\mathbb{E}_{a_i}[S]}]
\end{split}
\label{eq:withOB}
\end{equation}
where the second term on the right in the Equation is non-negative. Therefore, we have: $Var(\hat{g}^i_{DCCDA-OB}) \leq Var(\hat{g}^i_{DCCDA})$, which completes the proof.


\section{Settings, Implementations, Algorithms, Parameters, and Additional Results}

\subsection{Comparison in MADRL Settings}
\label{app:comSettings}

\begin{figure}[t]
\centering 
\includegraphics[width=0.46\textwidth]{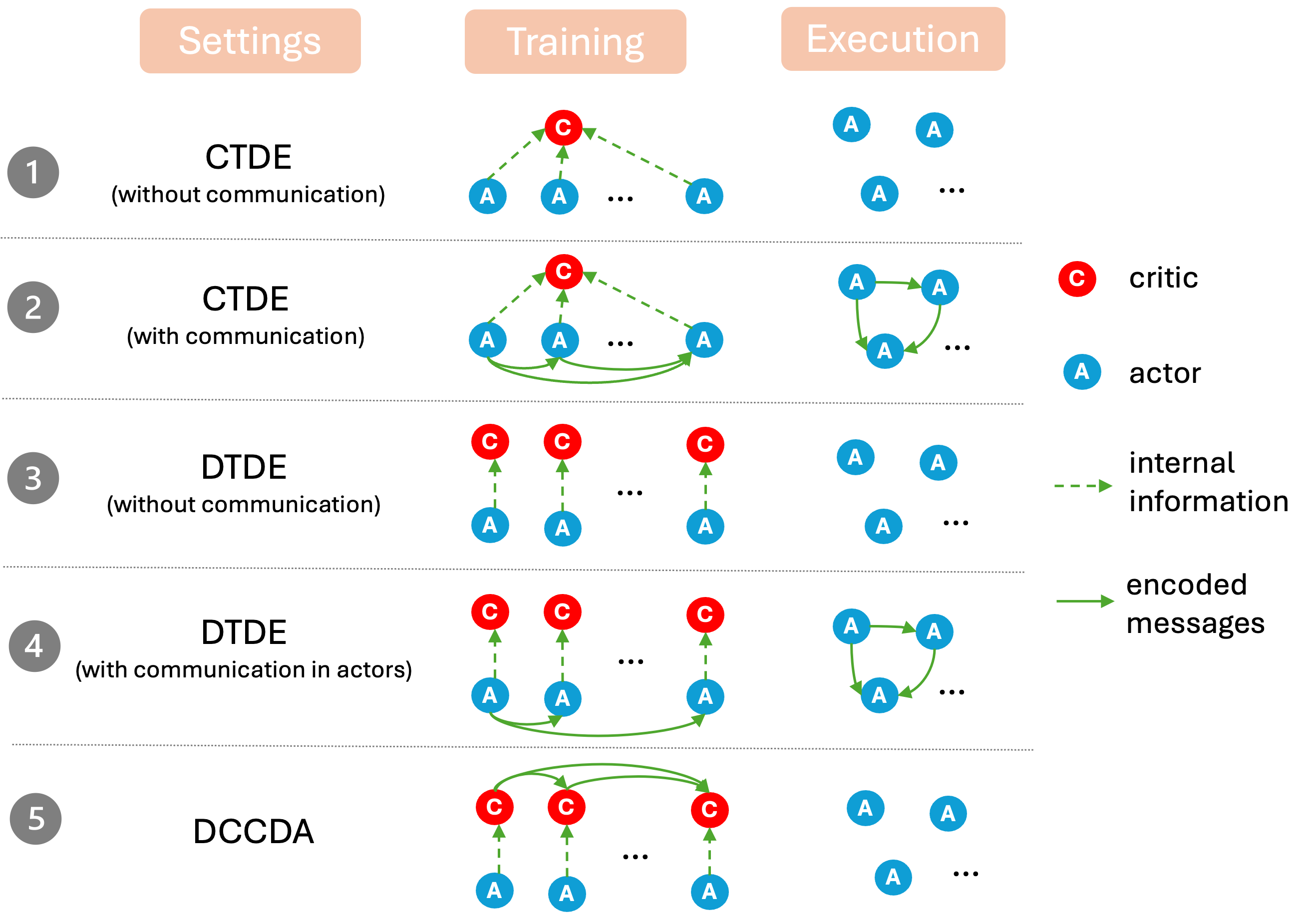} 
\caption{{The training and execution phases for CTDE (without communication), CTDE (with communication), DTDE (without communication), and DCCDA using actor-critic methods.}}
\label{fig:comparison}
\end{figure}

To position our focused DCCDA setting within MADRL, we illustrate various settings, with and without communication, across training and execution phases in Figure \ref{fig:comparison}. Note that we specifically focus on actor-critic methods, which align with the DCCDA setting used in our work. As motivated in the introduction, Settings $2\&4$ allows agents communicating during policy execution, which may not satisfy practical requirements of security and privacy. Also, Setting $3$ is fundamentally different from other settings as communication is not utilized during the training phase. 
It should be noted that Setting $1$ utilizes global information during the centralized training, which can be comparable with a situation where all information is communicated during the training phase. For this reason, we compare our DCCDA setting with Setting $1$ in both theoretical analysis and experiments.

\subsection{Implementations Details and Algorithms}
\label{app:extAlg}

We present two cases demonstrating the extension of two DCCDA algorithms using our proposed techniques. Firstly, we select the state-of-the-art method under the DCCDA setting, GAAC \cite{Liu2020G2ANet}, and illustrate the extension, GAAC-OB-KL, in Algorithm \ref{alg:algoGAAC}. Due to the absence of learning communication methods under the DCCDA setting, we opt for the state-of-the-art method under the DTDE setting (without communication), IPPO \cite{Yu2022PPO}, and introduce a standard communication architecture, resulting in IPPO-Comm. Subsequently, IPPO-Comm is further extended with our proposed techniques, forming IPPO-Comm-OB-KL, which is illustrated in Algorithm \ref{alg:algoIPPO}. Note that we highlight the communication process, including determining message content, exchange, and the optimization of communication models, as \textcolor{blue}{BLUE} in Algorithms \ref{alg:algoGAAC} and \ref{alg:algoIPPO}.

\textbf{Communication Architecture used in IPPO-Comm}. In IPPO-Comm, we integrate a communication architecture that considers the historical information and current actions of sender agents in messages, which is in line with our Assumption 1 to encode all available local information into messages. Specifically, we have explored various methods to generate and encode actions. Ultimately, we chose the most effective approach: concatenating the policy distribution (which probabilistically indicates the actions) along with the history/local information (hidden states from the LSTM when using histories as input) of the sender agents. Next, the concatenation is encoded through an MLP, and the output layer uses softmax to produce a distribution of messages. These messages are then selected and sent to receiver agents.  

\textbf{Algorithm Descriptions}. In Algorithm \ref{alg:algoGAAC}, each agent's actor, critic, and communication model are initialized first. During each training iteration, agents communicate through the communication process introduced by GAAC. We utilize the implementation of GAAC from a publicly accessible repository.\footnote{The open-source code is available at https://github.com/starry-sky6688/MARL-Algorithms.} In the algorithm, agents store observations, actions, messages, rewards, new observations, Q-values, and action probabilities in their buffer for training including computing the value of the message-dependent baseline and the KL divergence term. When training is enabled, agents train their policies using the gradients defined in $g^{i}_{DCCDA-OB-KL} = g^i_{DCCDA-OB} + \beta \nabla_{\theta_i} \mathcal{L}_{KL}(\theta_i)$. Critics are trained in a DQN-like manner, and the communication model is trained according to GAAC. We further highlight the communication process introduced by GAAC as BLUE in Algorithm \ref{alg:algoGAAC}. Similarly, we highlight the communication process in IPPO-Comm-OB-KL in Algorithm \ref{alg:algoIPPO}. We use the implementation of IPPO from a publicly accessible repository.\footnote{The open-source code is available at https://github.com/marlbenchmark/on-policy.} Essentially, IPPO-Comm-OB-KL and GAAC-OB-KL differ only in how agents communicate and train their communication models.

\begin{algorithm}[t]
\caption{GAAC-OB-KL using regularized policies and message-dependent baselines}\label{alg:algoGAAC}
\begin{algorithmic}[1]
\State Initialize $\theta_i$ and $\phi_i$, the parameters for each agent's actor and critic. Initialize the communication model for each agent.
\For{each training iteration}
\State Initialize data buffer $D_i$ for each agent $i$
\State Get initial observations $\boldsymbol{o}_0=\{o^1_0,...,o^i_0,...,o^N_0\}$ and set initial history $\boldsymbol{h}_0$
\For{$t$ = 0 to \textit{max_steps_per_episode}}
\For{each agent $i$}
\State Decide an action $a^i_t$ and output the corresponding probability distribution: $p^i_t \leftarrow \pi_i(a^i_t|h^i_t, \theta_i)$ 
\State \textcolor{blue}{Generate messages $m^i_t$ from encoded observations and actions}
\State \textcolor{blue}{Send messages to other agents}
\State \textcolor{blue}{Aggregate received messages through a two-layer attention mechanism}
\State Generate the corresponding Q-value: $q^i_t \leftarrow Q_i(h^i_t,a^i_t,m^{-i}_t)$
\EndFor
\State Get new observations $\boldsymbol{o}_{t+1}$ and rewards $r_t$ by performing actions
\State Insert experience $(o^i_t, m^{-i}_{t}, a^i_{t}, r_t, o^i_{t+1}, q^i_t, p^i_t)$ into $D_i$ and update $h^i_t$ for each agent $i$
\EndFor
\For{each agent $i$}
\State Sample a train batch $b_i$ from buffer $D_i$
\State Calculate the KL objective $\mathcal{L}_{KL}(\theta_i)$ using sampled experience
\State Compute the message-dependent baseline $b_i^*(h_i, m_{-i})$ using sampled experience
\State Update $\theta_i$ with Adam/RMSProp following the gradient defined in $g^{i}_{DCCDA-OB-KL}$
\EndFor
\State \textcolor{blue}{Update the communication model evaluated by $Q_i(h^i_t,a^i_t,m^{-i}_t)$ for each agent $i$}
\State Update the critic parameter $\phi_i$ for each agent $i$ with TD-learning
\EndFor
\end{algorithmic}
\end{algorithm}

\begin{algorithm}[t]
\caption{IPPO-Comm-OB-KL using regularized policies and message-dependent baselines}\label{alg:algoIPPO}
\begin{algorithmic}[1]
\State Initialize $\theta_i$ and $\phi_i$, the parameters for each agent's actor and critic. Initialize the communication model for each agent.
\For{each training iteration}
\State Initialize data buffer $D_i$ for each agent $i$
\State Get initial observations $\boldsymbol{o}_0=\{o^1_0,...,o^i_0,...,o^N_0\}$ and set initial history $\boldsymbol{h}_0$
\For{$t$ = 0 to \textit{max_steps_per_episode}}
\For{each agent $i$}
\State Decide an action $a^i_t$ and output the corresponding probability distribution: $p^i_t \leftarrow \pi_i(a^i_t|h^i_t, \theta_i)$ 
\State \textcolor{blue}{Generate messages $m^i_t$ from individual history and policy distribution}
\State \textcolor{blue}{Send messages to other agents}
\State Generate the corresponding Q-value: $q^i_t \leftarrow Q_i(h^i_t,a^i_t,m^{-i}_t)$
\EndFor
\State Get new observations $\boldsymbol{o}_{t+1}$ and rewards $r_t$ by performing actions
\State Insert experience $(o^i_t, m^{-i}_{t}, a^i_{t}, r_t, o^i_{t+1}, q^i_t, p^i_t)$ into $D_i$ and update $h^i_t$ for each agent $i$
\EndFor
\For{each agent $i$}
\State Sample a train batch $b_i$ from buffer $D_i$
\State Calculate the KL objective $\mathcal{L}_{KL}(\theta_i)$ using sampled experience
\State Compute the message-dependent baseline $b_i^*(h_i, m_{-i})$ using sampled experience
\State Update $\theta_i$ with Adam/RMSProp following the gradient defined in $g^{i}_{DCCDA-OB-KL}$
\EndFor
\State \textcolor{blue}{Update the communication model evaluated by $Q_i(h^i_t,a^i_t,m^{-i}_t)$ for each agent $i$}
\State Update the critic parameter $\phi_i$ for each agent $i$ with TD-learning
\EndFor
\end{algorithmic}
\end{algorithm}

\textbf{Implementations of Inner Product $S$}. Recall that we use $S=\nabla_{\theta_i}\log \pi_i(a_i|h_i, \theta_i)^T\nabla_{\theta_i}\log \pi_i(a_i|h_i, \theta_i)$ to denote the inner production of the gradient $\nabla_{\theta_i}\log \pi_i(a_i|h_i, \theta_i)$. This can be computationally challenging due to the extremely high dimension of the parameter space (parametrized by a neural network). Inspired by the work of \cite{Kuba2021Setting}, we use the softmax policy, which allows the product to be computed in an analytical form. We further leverage the implementation of the inner product from \cite{Kuba2021Setting} to build our message-dependent baseline, explicitly incorporating messages in the replay buffer and during the computation of the baseline values.

\subsection{Comparison in Methods}
\label{app:compMethods}

We illustrate the essential components of all methods and how they differ from each other in MADRL setting, critics and policy regularization techniques in Table \ref{tab:allmethods}. Notably, IPPO-Comm-OB-KL and GAAC-OB-KL inherently differ other methods due to the proposed message-dependent baseline and the regularization concerning communication. COMA, MAPPO, and MAT use baseline techniques based on state-value or action-value functions that do not account for encoded messages. The communication method GAAC does not employ a baseline technique. IPPO-Comm, on the other hand, follows the same training strategies as IPPO, which includes a baseline based on state-value functions.

\begin{table}[t]
    \caption{The essential components of all methods.}
    \label{tab:allmethods}
\begin{center}
\begin{sc}
\begin{tabular}{p{0.2\linewidth}p{0.1\linewidth}p{0.35\linewidth}p{0.3\linewidth}}
        \toprule
        methods & settings & critics & regularization \\
        \midrule
        COMA & CTDE & $Q(s, \boldsymbol{a})-\mathbb{E}_{a_i}[Q(s,a_i, a_{-i})]$ & no \\
        MAPPO & CTDE & $Q(s, \boldsymbol{a}) - V(s)$ & entropy \\
        MAT & CTDE & $Q(s, \boldsymbol{a}) - V(s)$ & entropy \\
        IPPO-Comm & DCCDA & $Q_i(h_i, a_i, m_{-i}) - V_i(h_i)$ & entropy \\
        IPPO-Comm-OB-KL & DCCDA & \parbox{5cm}{$Q_i(h_i, a_i, m_{-i}) - b_i(h_i, m_{-i})$ (Eq. 1 in the main paper)} & KL (Eq. 3 in the main paper) \\
        GAAC & DCCDA& $Q_i(h_i, a_i, m_{-i})$ & no \\
        GAAC-OB-KL & DCCDA & \parbox{5cm}{$Q_i(h_i, a_i, m_{-i}) - b_i(h_i, m_{-i})$ (Eq. 1 in the main paper)} & KL (Eq. 3 in the main paper) \\
        \bottomrule
\end{tabular}
\end{sc}
\end{center}
\vskip -0.1in
\end{table}

\subsection{Statistical Tests}
\label{app:statistical}

In Table \ref{tab:allMeanSTD}, we report the median win rate and standard deviation on all evaluated methods in SMAC and Traffic Junction. We also report the mean and 95\% confidence interval of all methods in the last 100 evaluation periods in Table \ref{tab:MeanConfidence}. GAAC-OB-KL and IPPOComm-OB-KL achieve a higher win rate compared to all the other methods. In the meanwhile, IPPO-Comm-OB-KL and GAAC-OBKL have a lower or similar variance in win rate than other methods except for COMA, which performs much worse in the traffic junction domain.

\subsection{Parameter Choices}
\label{app:parmChoices}

Hyper-parameters used for IPPO-Comm-OB-KL and GAAC-OB-KL in the SMAC domain and Traffic Junction are shown in Table \ref{tab:parameters}. Note that we use \emph{hidden dim} to refer to the hidden dimension of the actor and critic model. We use \emph{comm dim} to denote the hidden dimension of the communication model. We further use \emph{attention dim} to denote the hidden dimension of the attention model used by GAAC for aggregating messages from the other agents. Note that hyperparameters for IPPO-Comm-OB-KL and GAAC-OB-KL were optimised using a grid search over learning rate and batch sizes with the grid centred on the hyperparameters used in the original papers (e.g., GAAC and IPPO) and parameter performance tested in all used environments. We further search the optimal parameters introduced by the KL objective (i.e., the temperature parameter and the scaling factor) and report the corresponding performance in Section \ref{app:addRes}.

\begin{table}[t]
\centering 
\caption{Important hyperparameters in SMAC and Traffic Junction.} 
\begin{sc}
\begin{tabular}{p{0.2\linewidth}|p{0.2\linewidth}|p{0.16\linewidth}|p{0.2\linewidth}|p{0.16\linewidth}}
\hline
& \multicolumn{2}{c|}{SMAC} & \multicolumn{2}{c}{Traffic Junction} \\
\hline
 Hyperparameters & IPPO-Comm-OB-KL &  GAAC-OB-KL & IPPO-Comm-OB-KL &  GAAC-OB-KL \\
\hline
\emph{actor lr} & 5e-4 & 5e-4 & 1e-3 & 1e-3 \\
\emph{critic lr} & 5e-4 & 5e-4 & 1e-2 & 1e-2 \\
\emph{comm lr} & 5e-4 & 5e-4 & 1e-3 & 1e-3 \\
\emph{gamma} & 0.99 & 0.99 & 0.99 & 0.99 \\
\emph{update epoch} & 10 & 10 & 10 & 10 \\
\emph{mini batch} & 1 & 1 & 1 & 1 \\
\emph{optimizer} & Adam & Adam & Adam & Adam \\
\emph{optim eps} & 1e-5 & 1e-5 & 1e-3 & 1e-3 \\
\emph{max grad norm} & 10 & 10 & 10 & 10 \\
\emph{hidden dim} & 64 & 64 & 64 & 64 \\
\emph{comm dim} & 64 & 64 & 64 & 64 \\
\emph{attention dim} & None & 32 & None & 32 \\
\emph{eval episodes} & 32 & 32 & 32 & 32 \\
\hline
\end{tabular}
\end{sc}
\label{tab:parameters}
\end{table}

\begin{table}[t]
    \caption{Compute time in SMAC (1o_10b_vs_1r) and Traffic Junction (hard).}
     \label{tab:computeTime}
\begin{center}
\begin{small}
\begin{sc}
    \begin{tabular}{p{0.12\linewidth}p{0.08\linewidth}p{0.08\linewidth}p{0.08\linewidth}p{0.08\linewidth}p{0.12\linewidth}p{0.08\linewidth}p{0.12\linewidth}}
        \toprule
    & COMA & MAPPO & MAT & IPPO-Comm & IPPO-Comm-OB-KL & GAAC & GAAC-OB-KL \\ \midrule
    1o\_10b\_vs\_1r & 1 day & 2 days & 2 days & 2.5 days & 2.5 days & 3.5 days & 4 days \\
    hard & 3 hours & 0.5 day & 1 day & 1 day & 1 day & 2 days & 2 days \\ \bottomrule
    \end{tabular}
\end{sc}
\end{small}
\end{center}
\end{table}

\subsection{Additional Results}
\label{app:addRes}

\textbf{Compute Time}. The experiments reported in the paper were conducted in parallel on a cluster using CPUs (32 cores). Regarding compute time, we set a maximum of 4 days for SMAC tasks and a maximum of 2 days for Traffic Junction tasks. We used the map 1o\_10b\_vs\_1r from SMAC and the map hard from Traffic Junction as examples and reported the consumed time for each method per seed in Table \ref{tab:computeTime}. As we can see, our proposed techniques do not significantly increase computation time. IPPO-Comm-OB-KL consumes a similar amount of time compared to IPPO-Comm, and GAAC-OB-KL has a similar or slightly higher computation time than GAAC. Due to the complex two-layer attention mechanism, GAAC and GAAC-OB-KL require more time than IPPO-Comm and IPPO-Comm-OB-KL. As a result, the computational efficiency of the communication methods under DCCDA with our proposed techniques largely depends on the specific communication methods used. Nevertheless, all methods were able to finish within the desired time limit.

\begin{figure}[t]
\centering 
\includegraphics[width=0.3\textwidth]{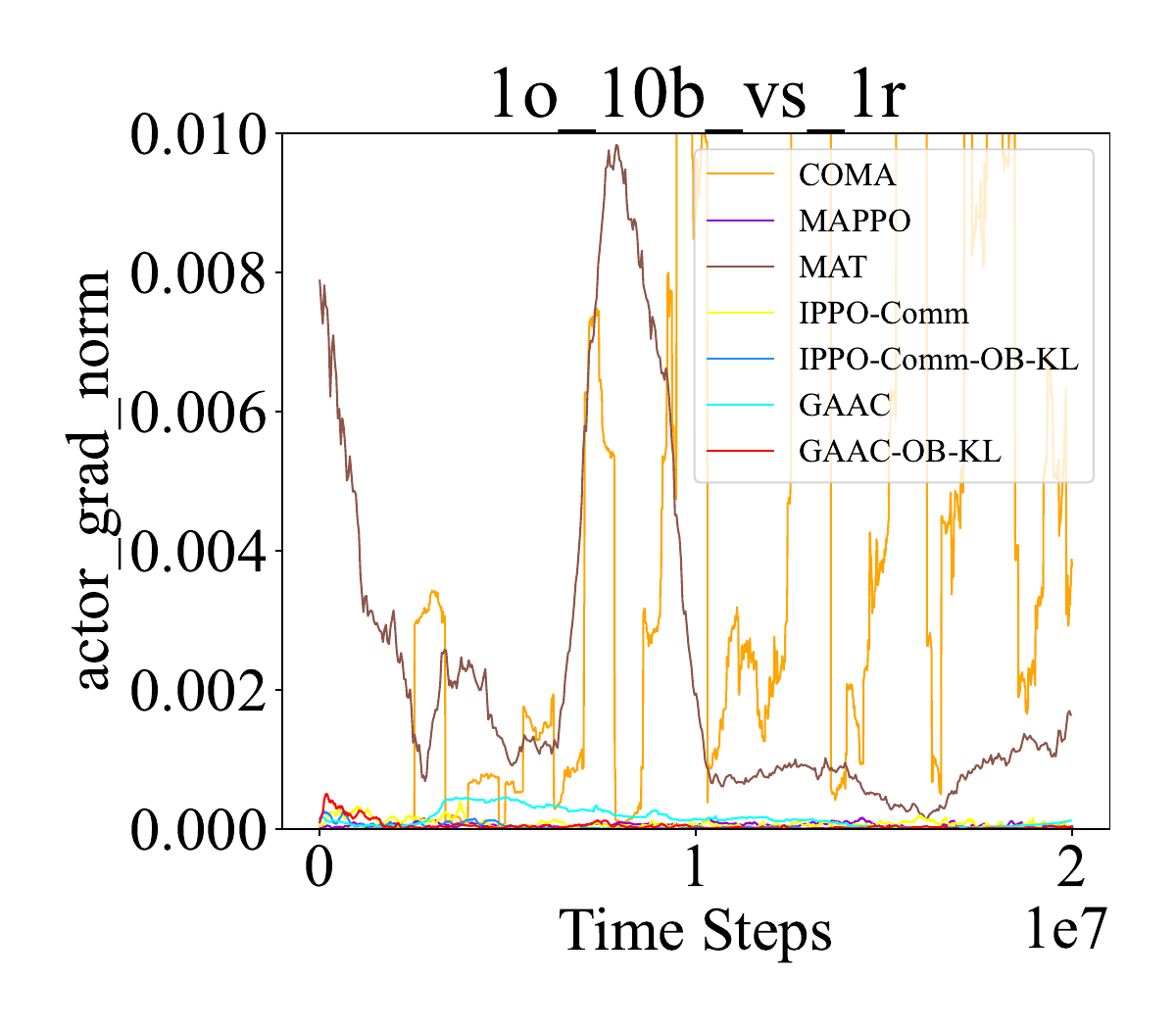} \hspace*{\fill}
\includegraphics[width=0.3\textwidth]{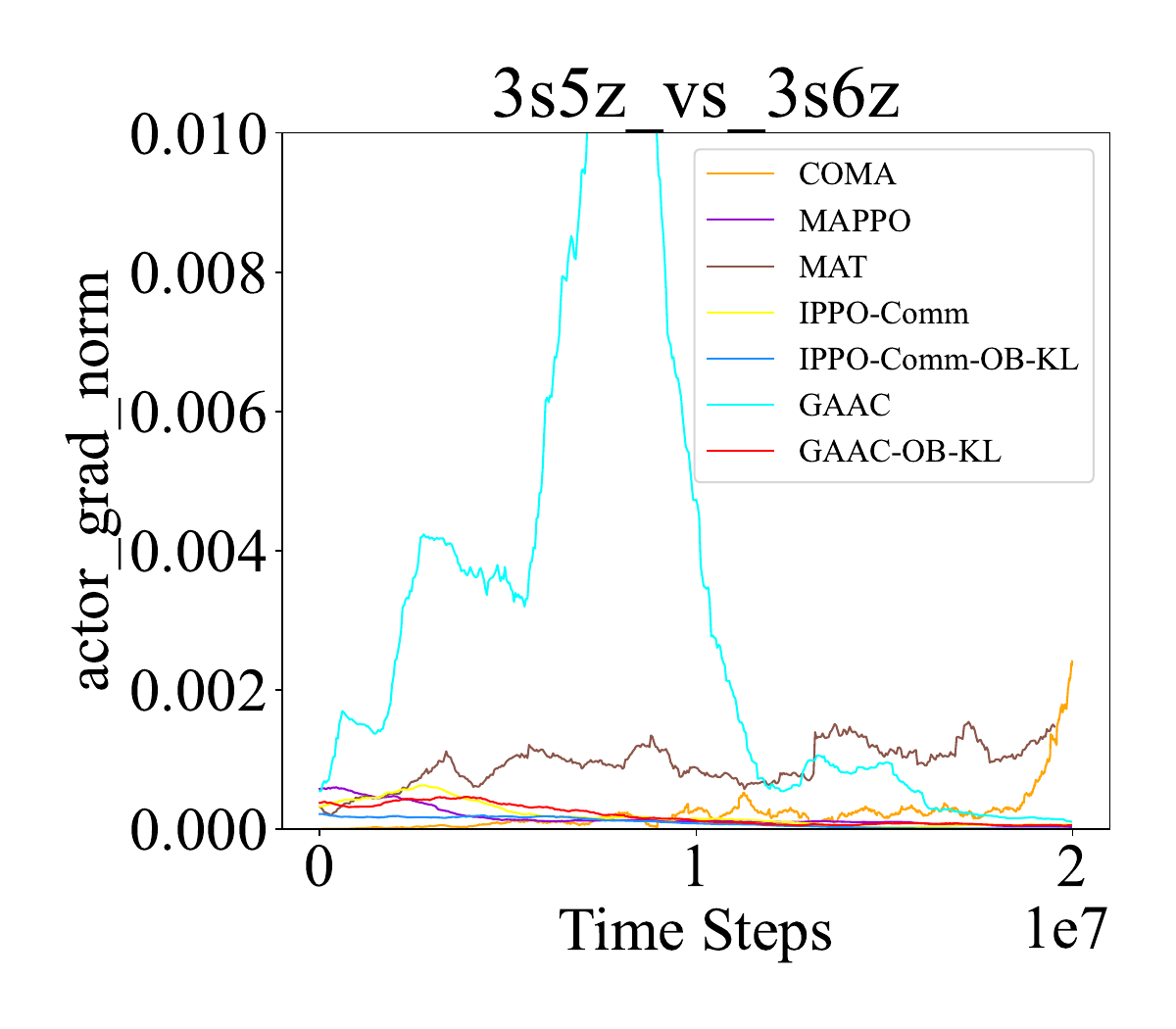} \hspace*{\fill}
\includegraphics[width=0.3\textwidth]{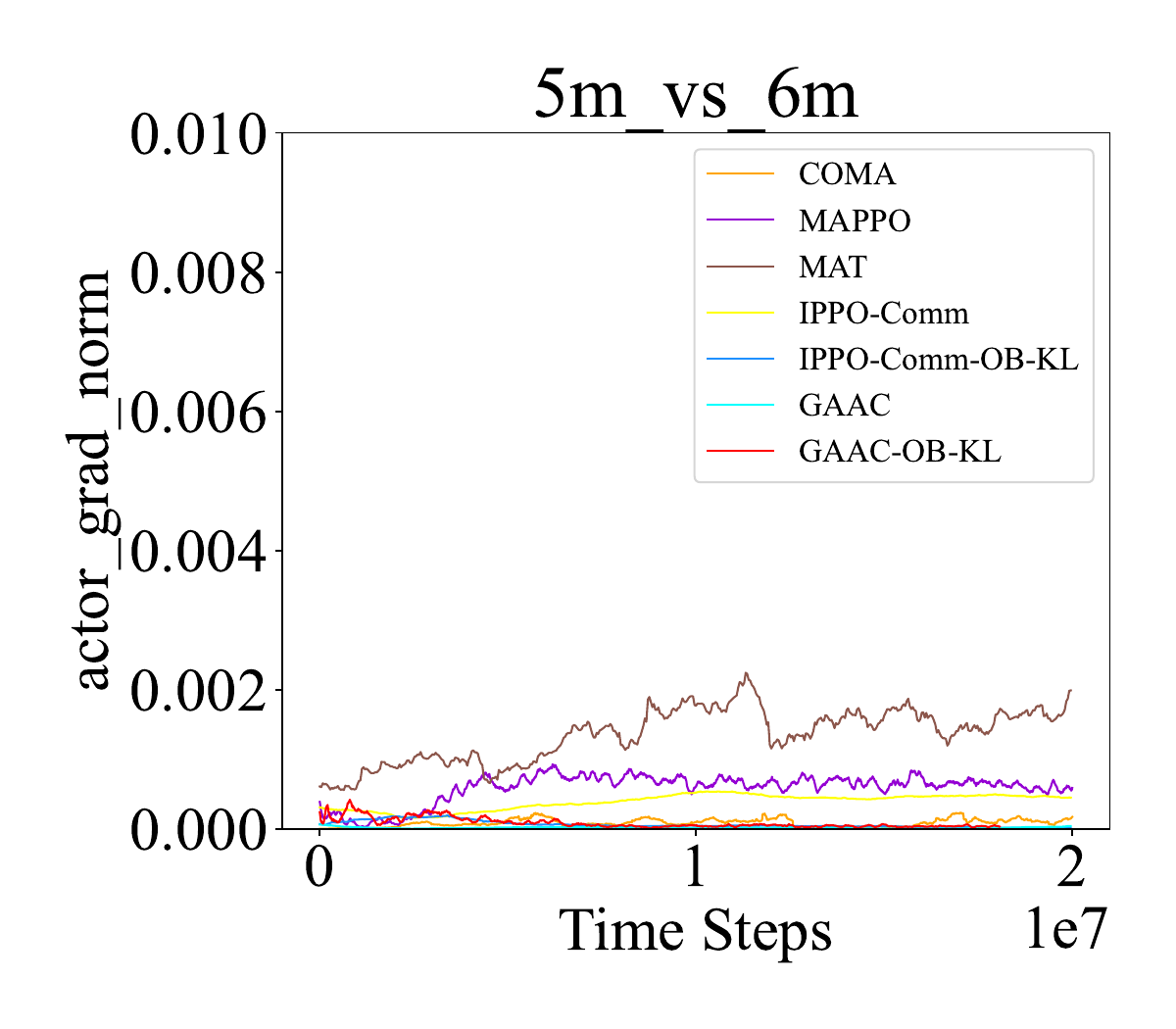} \hspace*{\fill} \\
\includegraphics[width=0.3\textwidth]{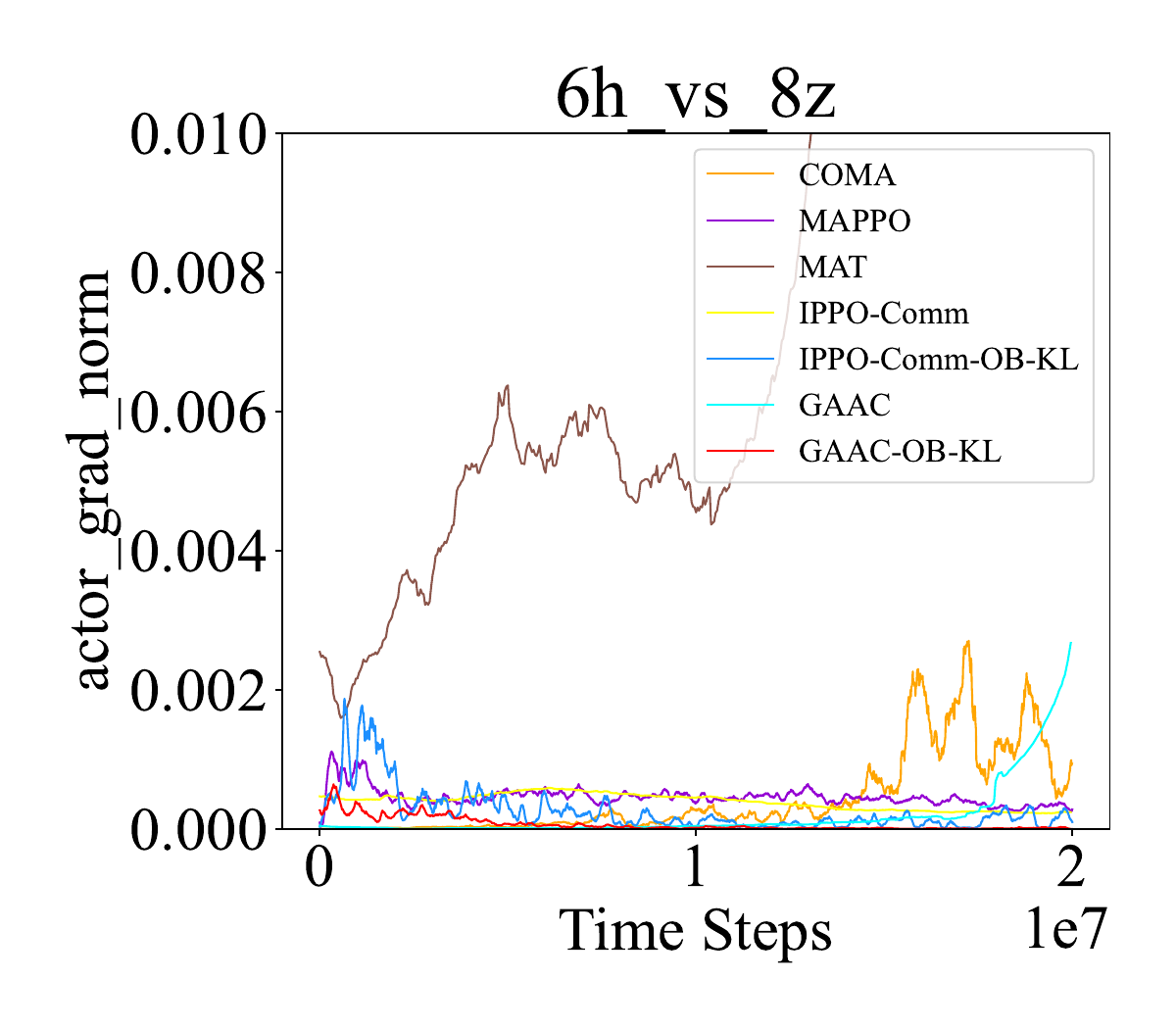} \hspace*{\fill}
\includegraphics[width=0.3\textwidth]{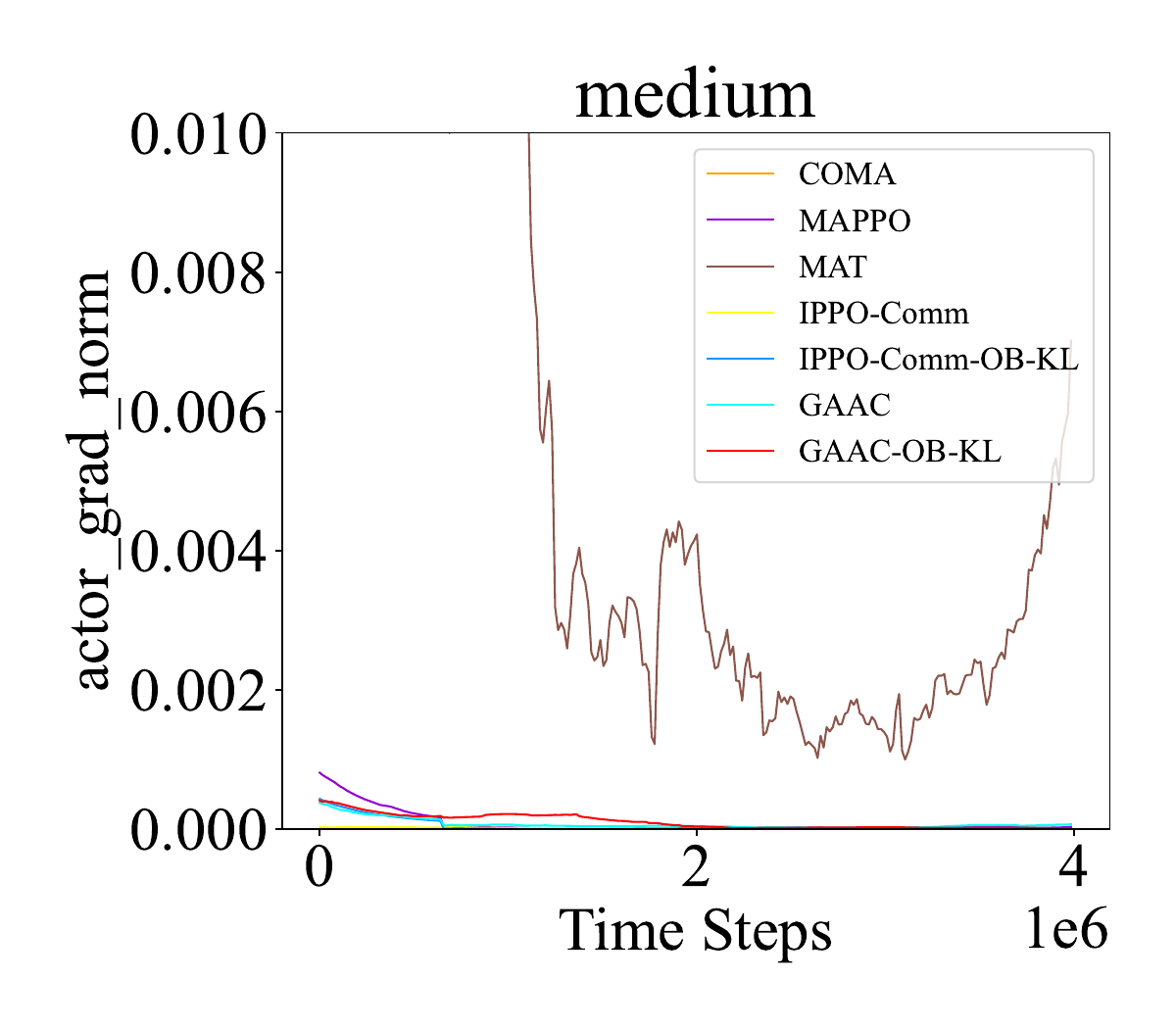} \hspace*{\fill}
\includegraphics[width=0.3\textwidth]{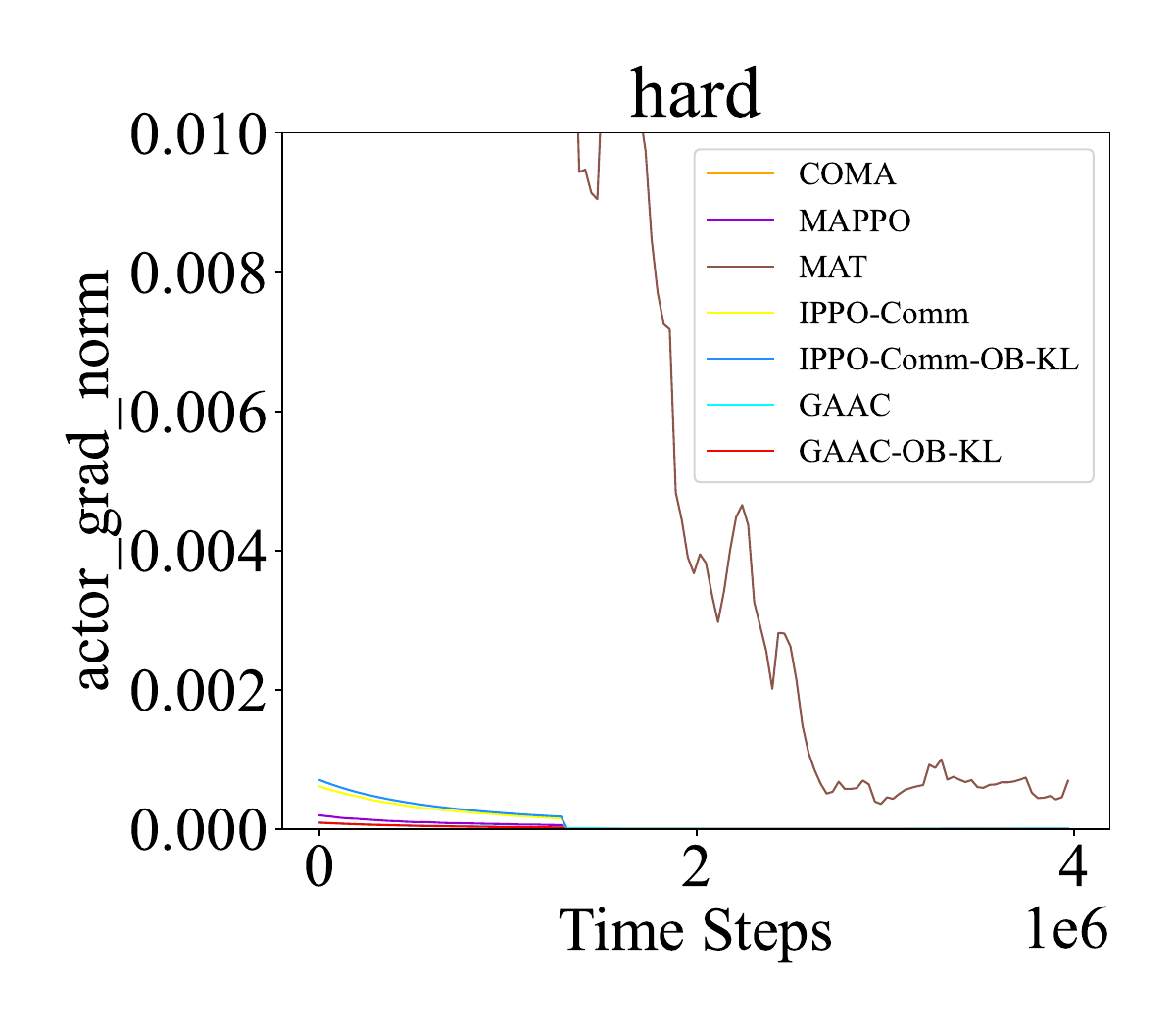} \hspace*{\fill}
\caption{{Variance in policy gradient norm of all methods.}}
\label{fig:varGradNorm}
\end{figure}

\textbf{Variance in Gradient Norm}. We present the variance in gradient norm across training steps for all methods in Figure \ref{fig:varGradNorm}. As shown, GAAC-OB-KL and IPPO-Comm-OB-KL exhibit significantly lower variance in policy gradients throughout training, indicating a more stable learning process compared to CTDE methods and those without the proposed techniques.

\begin{figure}[t]
\centering 
\subfloat[IPPO-Comm-OB-KL \\ in 1o\_10b\_vs\_1r]{\includegraphics[width=0.24\textwidth]{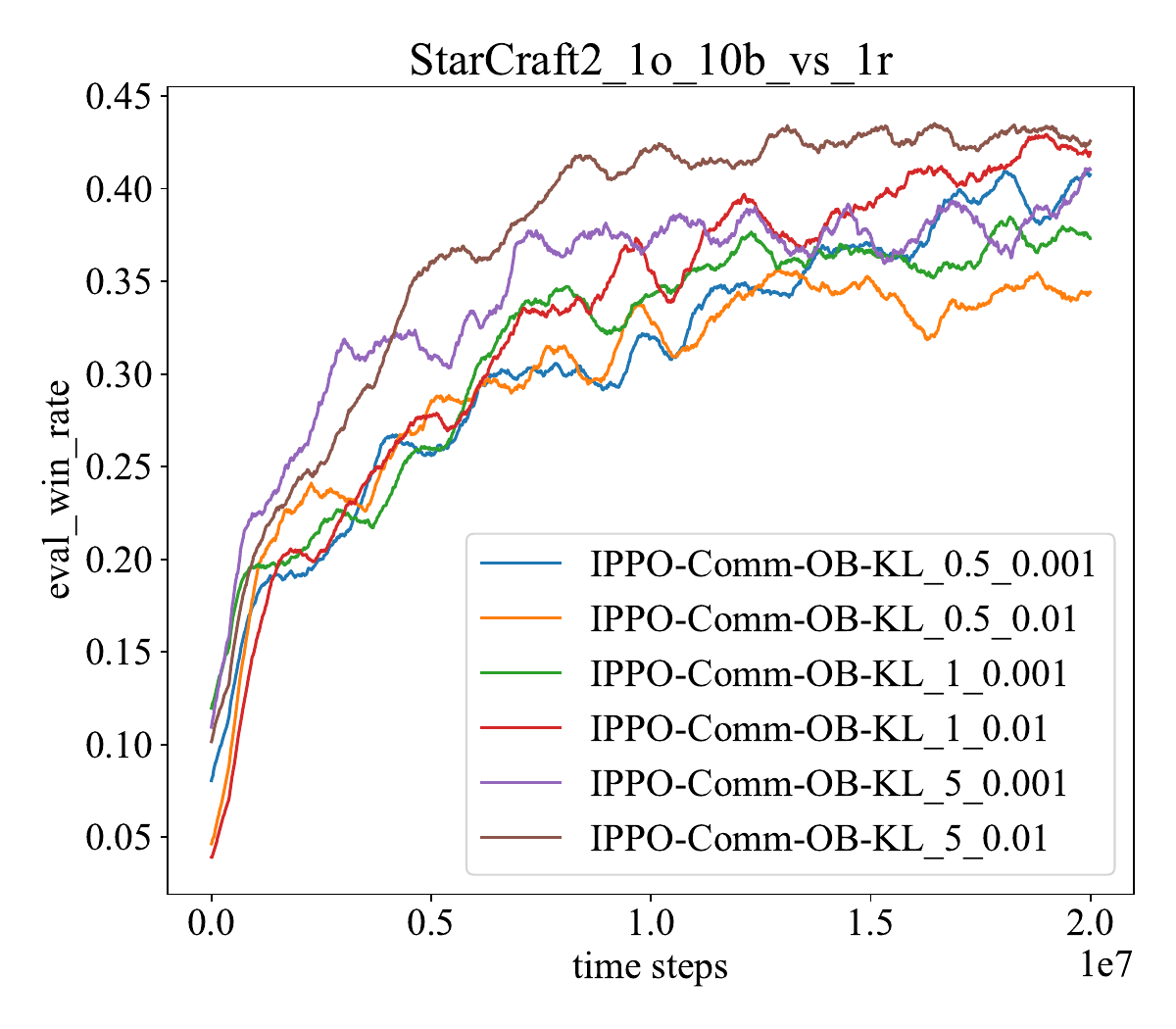}} \hspace*{\fill} 
\subfloat[GAAC-OB-KL \\ in 1o\_10b\_vs\_1r]{\includegraphics[width=0.24\textwidth]{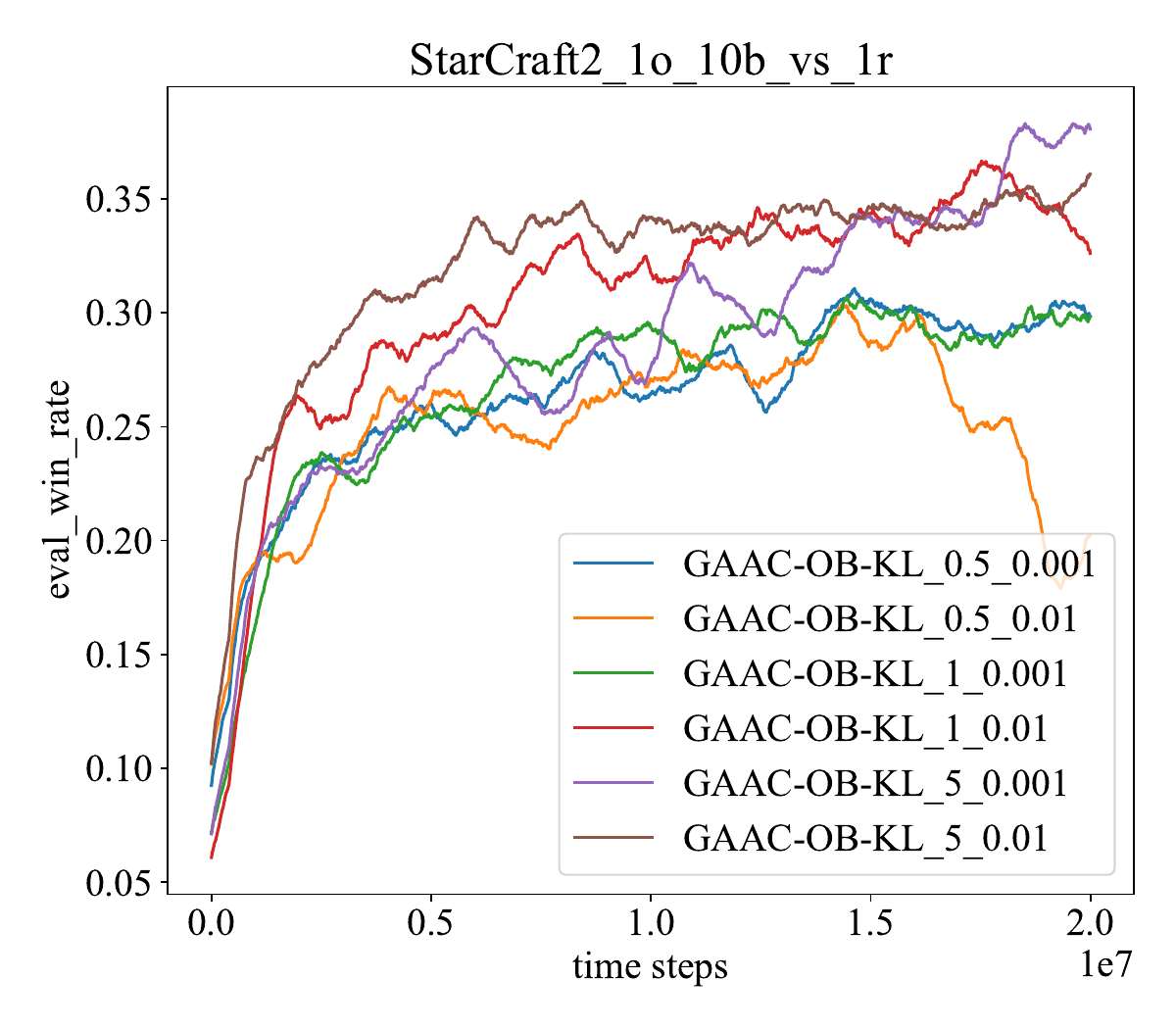}} \hspace*{\fill} 
\subfloat[IPPO-Comm-OB-KL \\ in 3s5z\_vs\_3s6z]{\includegraphics[width=0.24\textwidth]{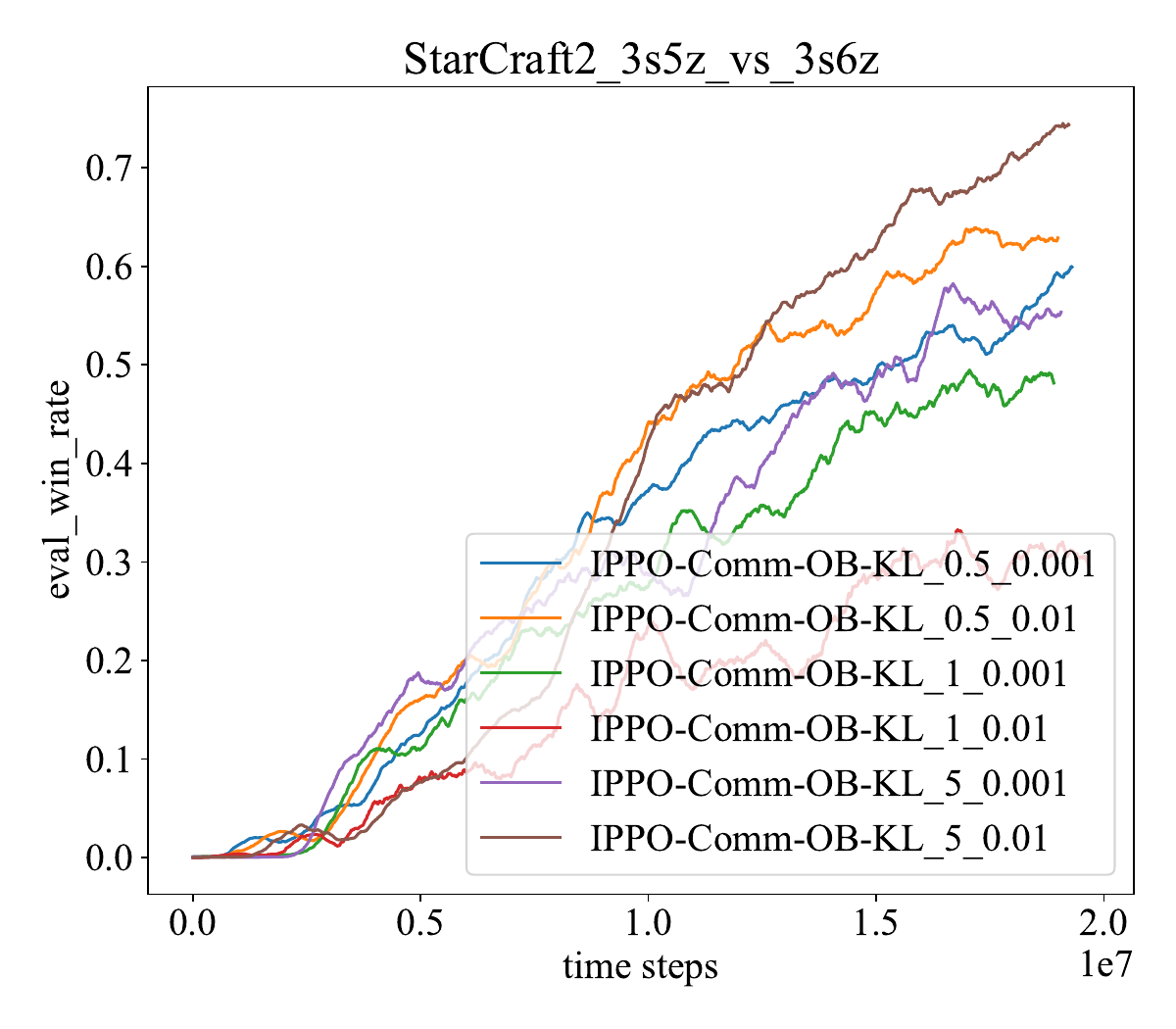}} \hspace*{\fill} 
\subfloat[GAAC-OB-KL \\ in 3s5z\_vs\_3s6z]{\includegraphics[width=0.24\textwidth]{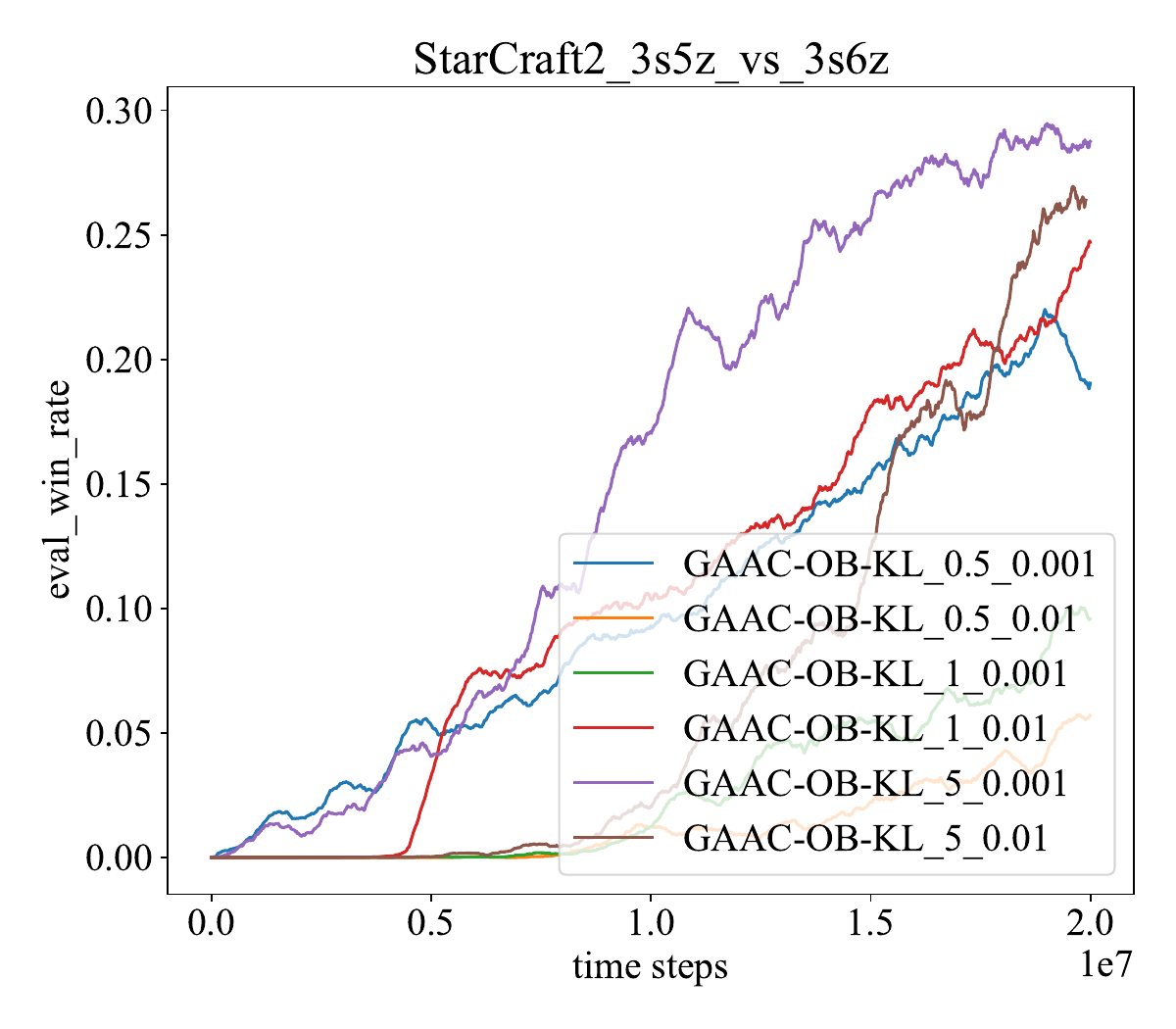}} \\
\subfloat[IPPO-Comm-OB-KL \\ in 5m\_vs\_6m]{\includegraphics[width=0.24\textwidth]{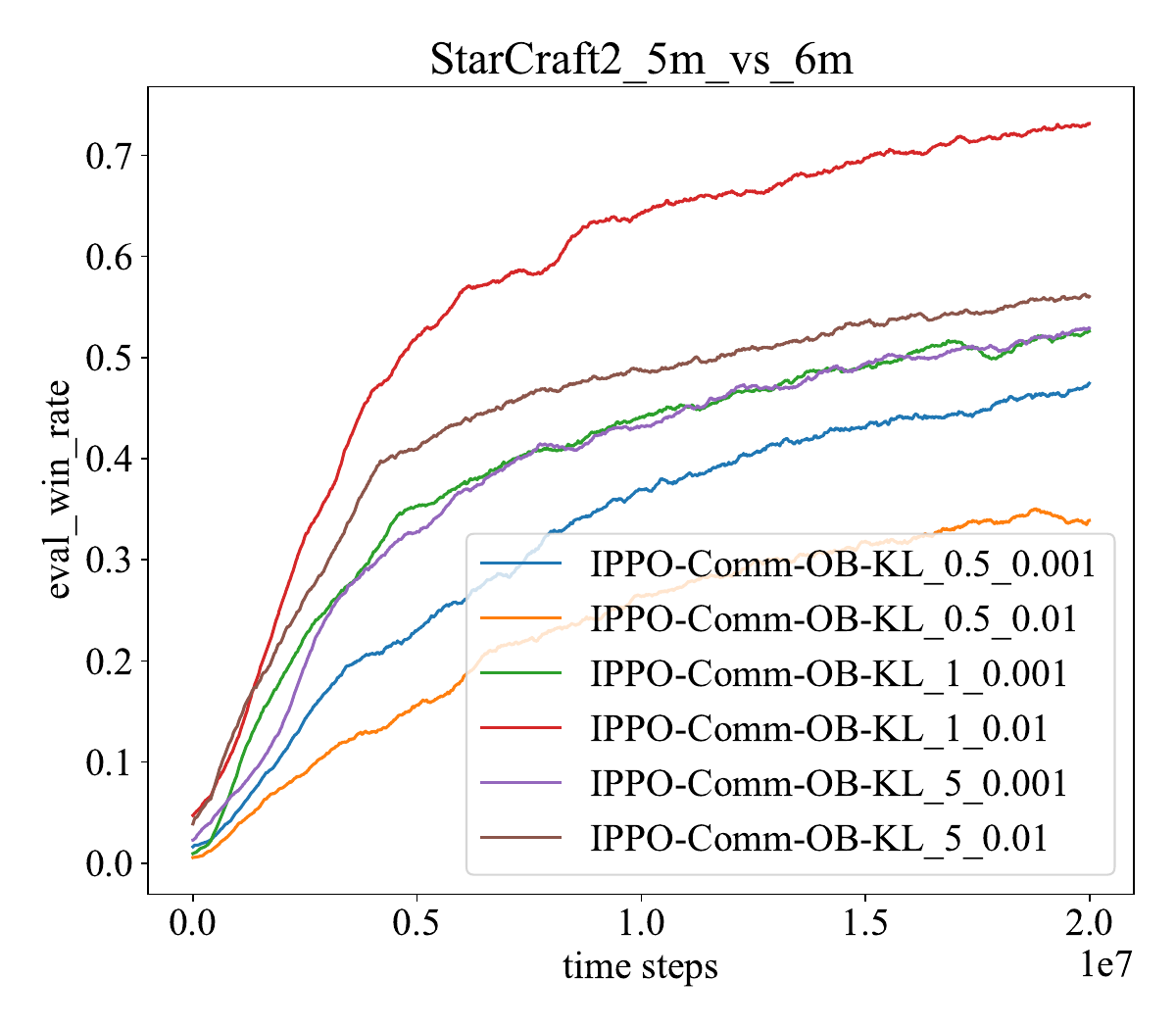}} \hspace*{\fill} 
\subfloat[GAAC-OB-KL \\ in 5m\_vs\_6m]{\includegraphics[width=0.24\textwidth]{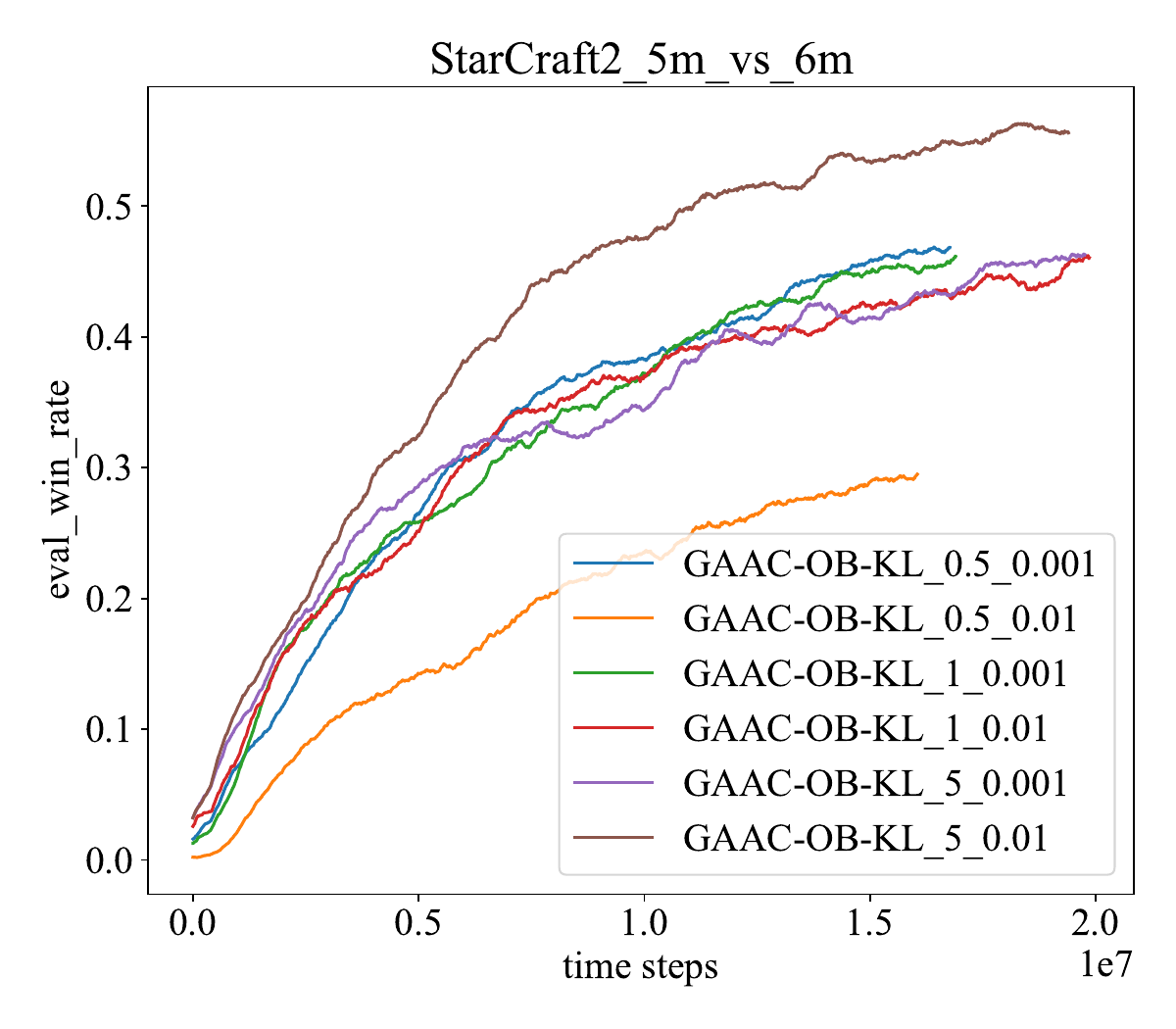}} \hspace*{\fill} 
\subfloat[IPPO-Comm-OB-KL \\ in 6h\_vs\_8z]{\includegraphics[width=0.24\textwidth]{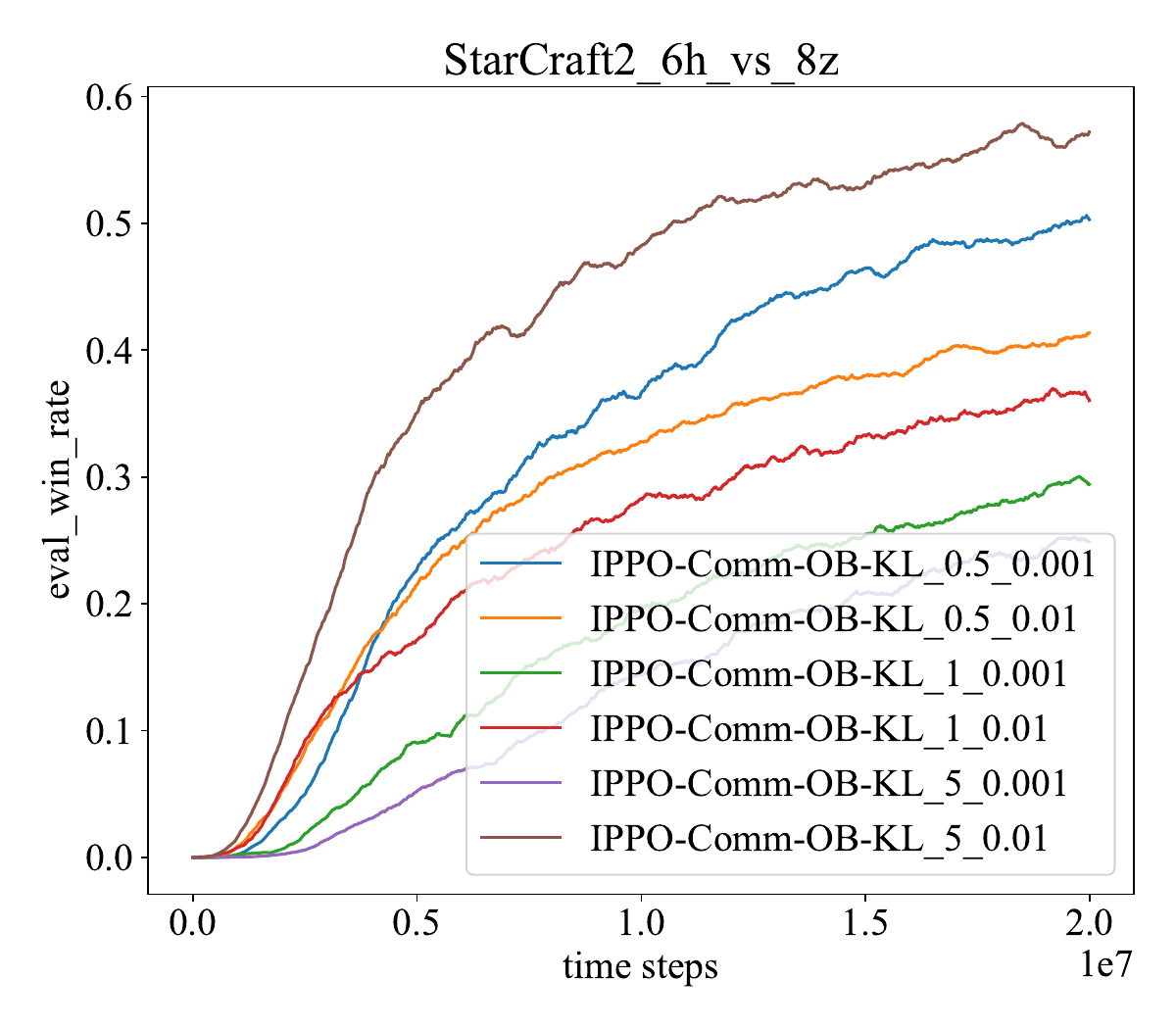}} \hspace*{\fill} 
\subfloat[GAAC-OB-KL \\ in 6h\_vs\_8z]{\includegraphics[width=0.24\textwidth]{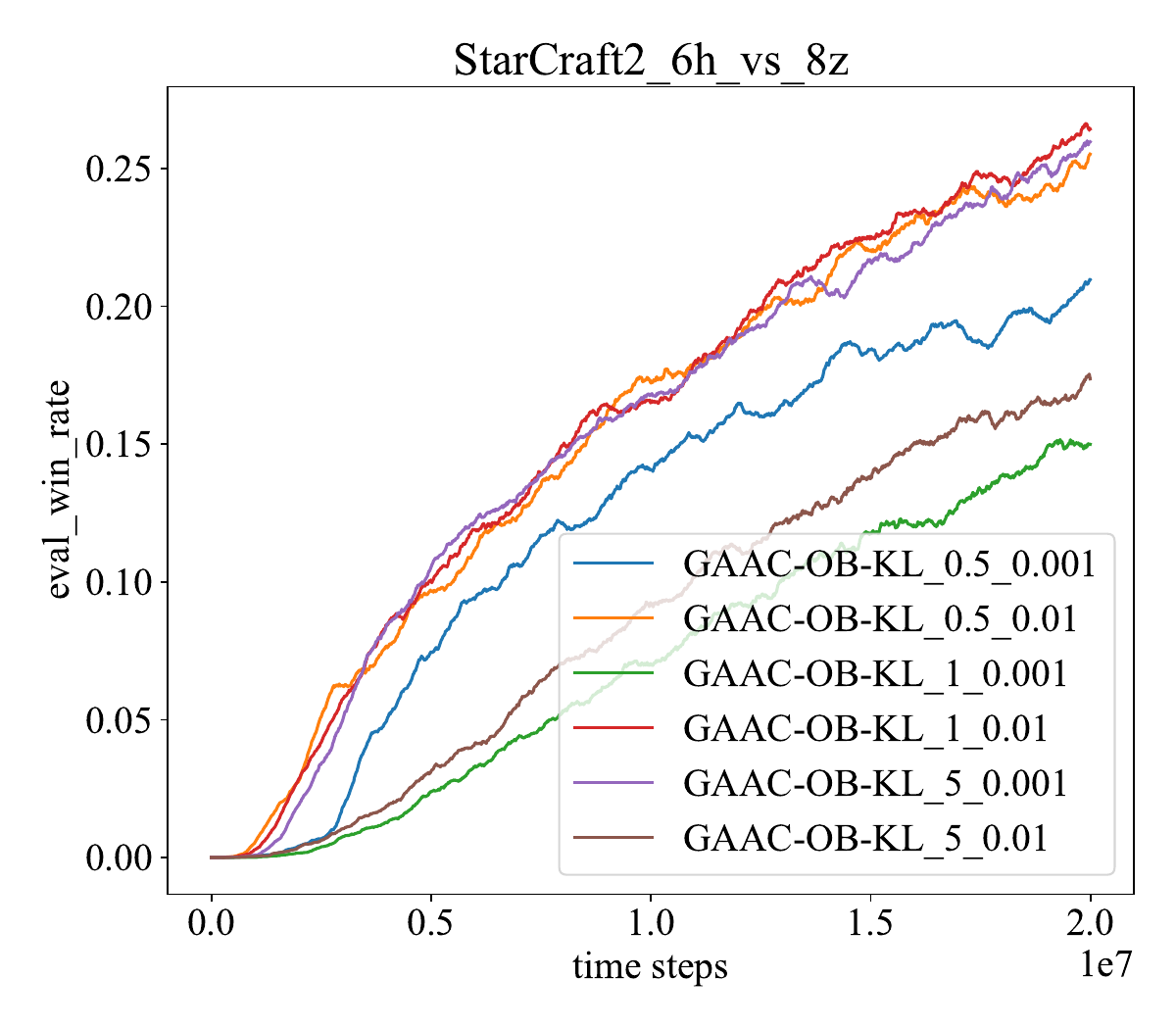}} \\
\subfloat[IPPO-Comm-OB-KL \\ in medium]{\includegraphics[width=0.24\textwidth]{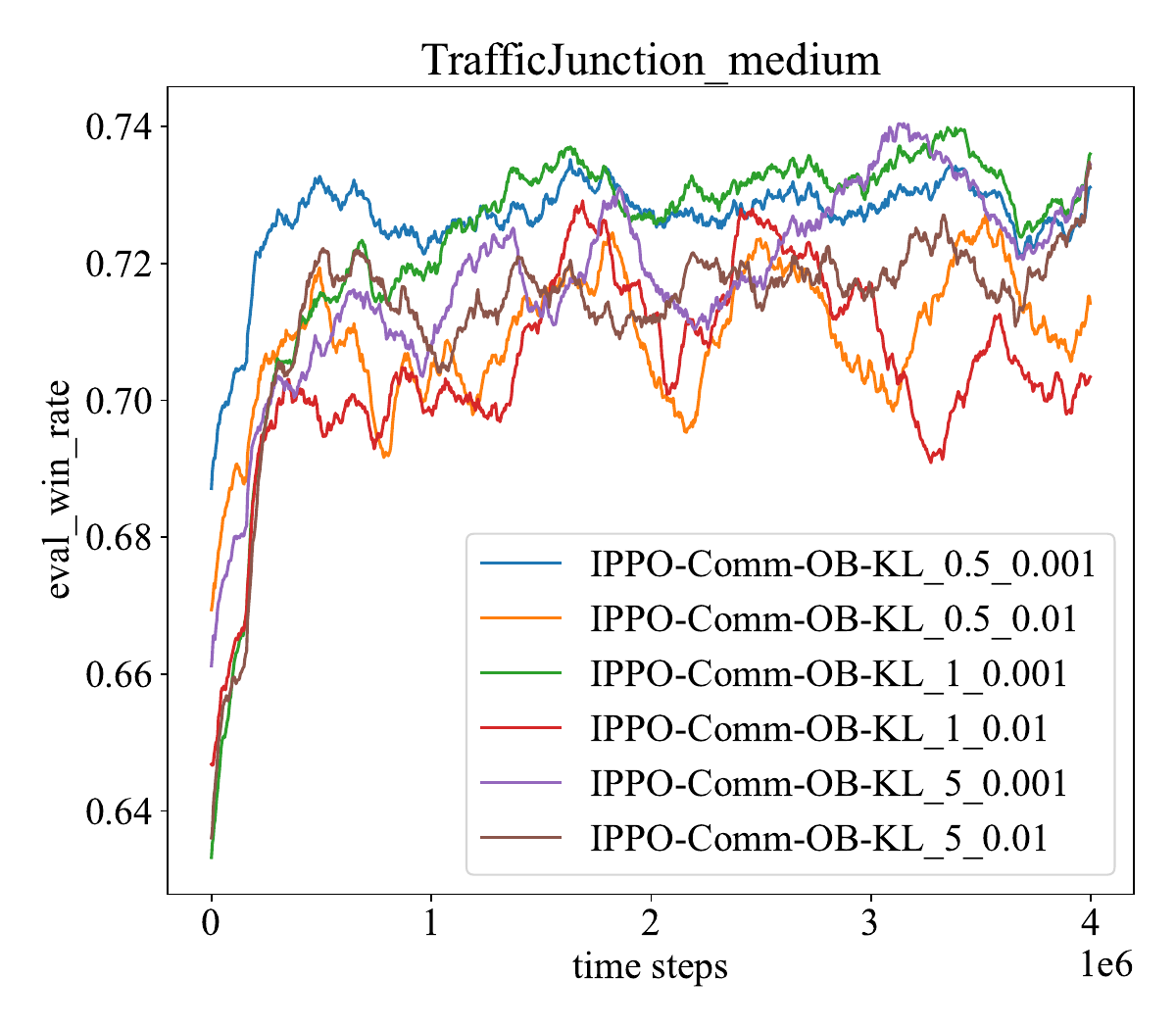}} \hspace*{\fill} 
\subfloat[GAAC-OB-KL \\ in medium]{\includegraphics[width=0.24\textwidth]{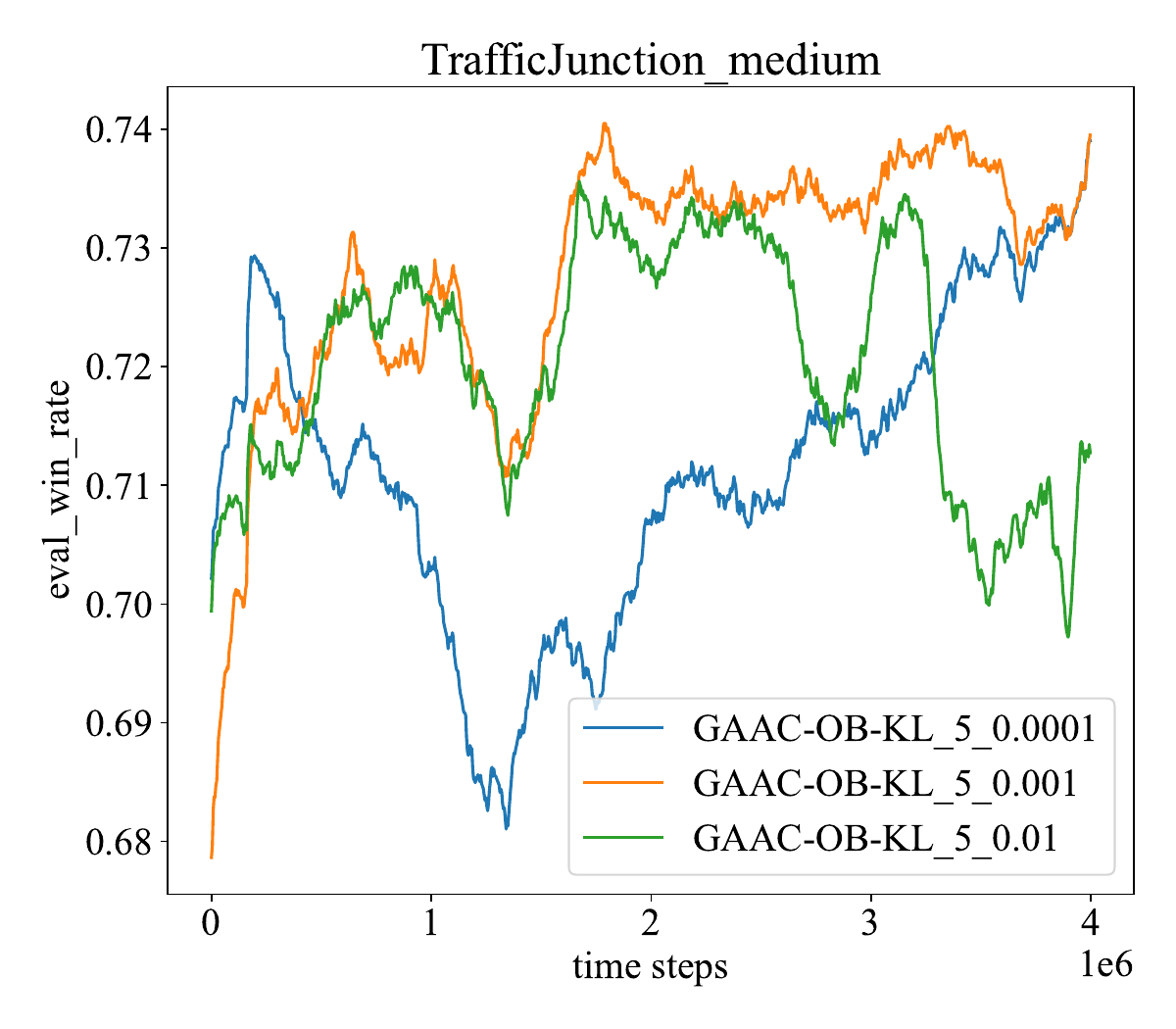}} \hspace*{\fill} 
\subfloat[IPPO-Comm-OB-KL \\ in hard]{\includegraphics[width=0.24\textwidth]{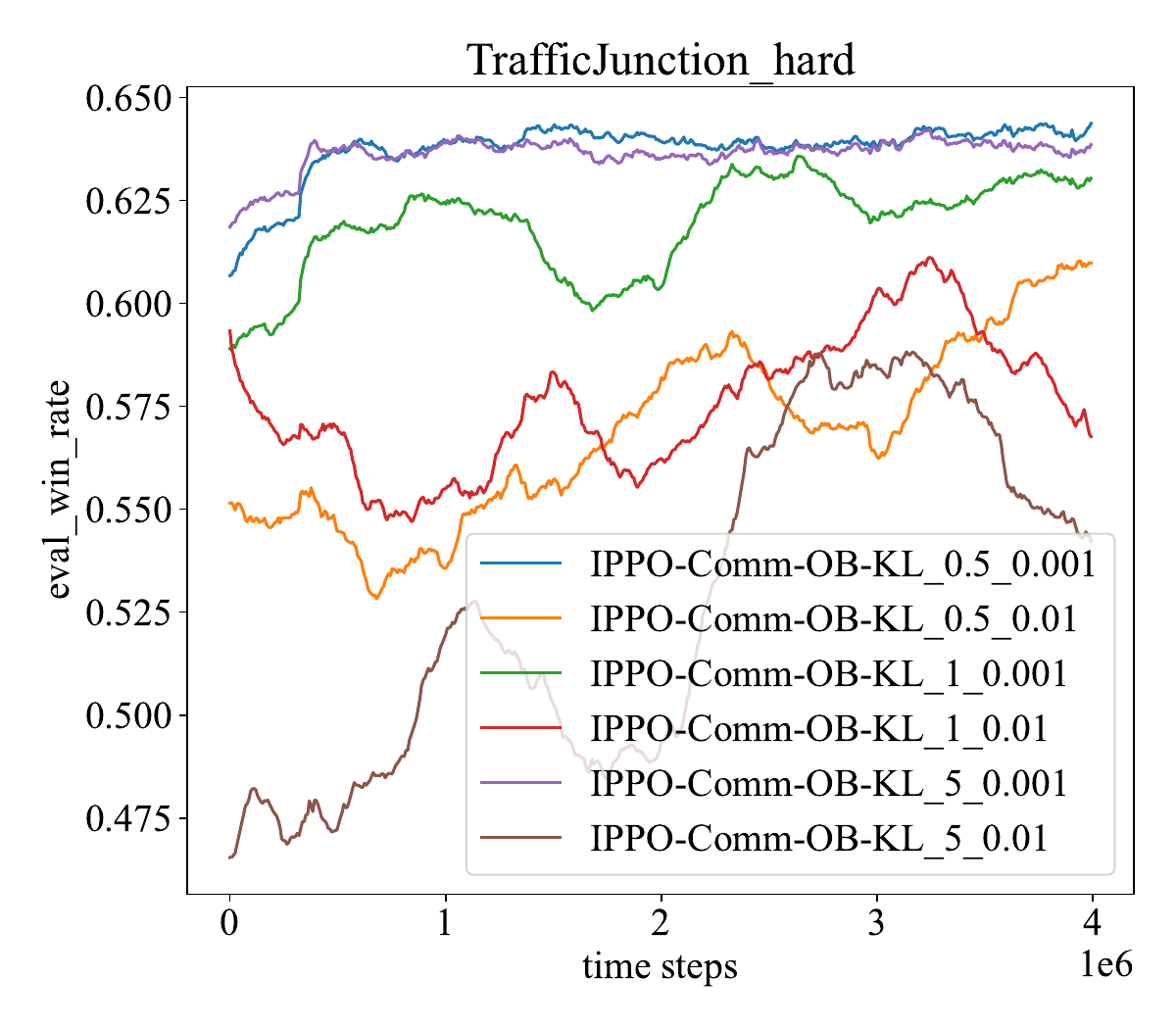}} \hspace*{\fill} 
\subfloat[GAAC-OB-KL \\ in hard]{\includegraphics[width=0.24\textwidth]{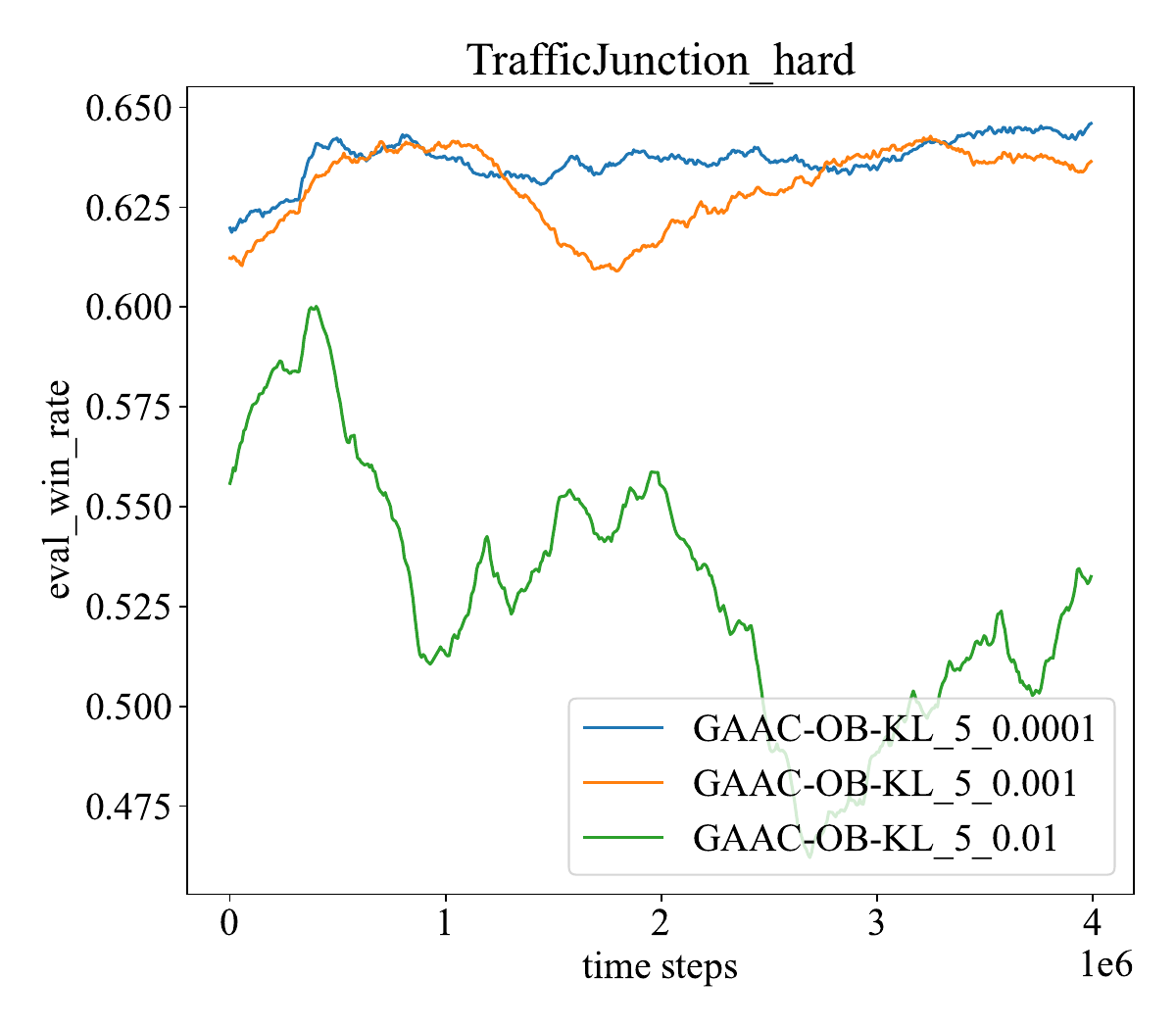}} \hspace*{\fill} 
\caption{{Averaged win rate  under different $\alpha$ and $\beta$ in 4 tasks of SMAC and 2 tasks of Traffic Junction.}}
\label{fig:finetuneAll}
\end{figure}

\textbf{Grid Search}. We conduct a grid search to fine-tune the temperature parameter $\alpha$ and the scaling factor $\beta$ of IPPO-Comm-OB-KL and GAAC-OB-KL. We show the performance of IPPO-Comm-OB-KL and GAAC-OB-KL under different combinations of $\alpha$ and $\beta$ in all 6 tasks (as shown in Figure \ref{fig:finetuneAll}). The label names in plots follow the format of IPPO-Comm-OB-KL\_$\alpha$\_$\beta$ and GAAC-OB-KL\_$\alpha$\_$\beta$. Note that we present the performance with the best parameters in the main paper.

\begin{table}[t]
    \caption{Median win rate and standard deviation on all evaluated methods in SMAC and Traffic Junction.}
    \label{tab:allMeanSTD}
\begin{center}
\begin{small}
\begin{sc}
    \begin{tabular}{p{0.2\linewidth}p{0.1\linewidth}p{0.1\linewidth}p{0.1\linewidth}p{0.1\linewidth}p{0.1\linewidth}p{0.1\linewidth}}
        \toprule
          & 1o\_10b\_vs\_1r & 3s5z\_vs\_3s6z & 5m\_vs\_6m & 6h\_vs\_8z & medium & hard \\ \midrule
COMA & 33.8 \scriptsize{(3.3)} & 0.0 \scriptsize{(0.0)} & 0.1 \scriptsize{(0.2)} & 0.0 \scriptsize{(0.0)} & 60.9 \scriptsize{(6.6)} & 42.2 \scriptsize{(6.8)} \\
MAPPO & 21.5 \scriptsize{(2.4)} & 65.9 \scriptsize{(4.6)} & 32.5 \scriptsize{(2.4)} & 33.9 \scriptsize{(3.0)} & 71.8 \scriptsize{(3.6)} & 57.4 \scriptsize{(4.9)} \\
MAT & 58.0 \scriptsize{(5.0)} & 4.1 \scriptsize{(2.0)} & 3.1 \scriptsize{(0.9)} & 13.5 \scriptsize{(2.5)} & 72.5 \scriptsize{(3.7)} & 57.8 \scriptsize{(4.9)} \\
IPPO-Comm & 2.9 \scriptsize{(1.3)} & 29.2 \scriptsize{(4.2)} & 22.2 \scriptsize{(2.3)} & 35.5 \scriptsize{(2.6)} & 70.4 \scriptsize{(6.4)} & 58.2 \scriptsize{(4.6)} \\
IPPO-Comm-OB & 27.6 \scriptsize{(3.4)} & 49.3 \scriptsize{(4.1)} & 70.3 \scriptsize{(2.7)} & 37.3 \scriptsize{(2.2)} & 69.4 \scriptsize{(3.6)} & 61.0 \scriptsize{(3.7)} \\
IPPO-Comm-KL & 1.5 \scriptsize{(0.8)} & 25.6 \scriptsize{(3.9)} & 33.2 \scriptsize{(2.8)} & 35.8 \scriptsize{(2.8)} & 71.0 \scriptsize{(3.8)} & 60.1 \scriptsize{(4.6)} \\
IPPO-Comm-OB-KL & 42.9 \scriptsize{(3.3)} & 70.9 \scriptsize{(4.0)} & 72.8 \scriptsize{(2.8)} & 56.4 \scriptsize{(2.5)} & 72.4 \scriptsize{(2.7)} & 64.2 \scriptsize{(3.2)} \\
GAAC & 10.9 \scriptsize{(2.6)} & 0.0 \scriptsize{(0.0)} & 3.0 \scriptsize{(1.1)} & 0.0 \scriptsize{(0.0)} & 70.5 \scriptsize{(3.4)} & 59.9 \scriptsize{(4.6)} \\
GAAC-OB & 26.0 \scriptsize{(3.6)} & 0.7 \scriptsize{(0.6)} & 32.1 \scriptsize{(3.0)} & 34.9 \scriptsize{(2.4)} & 70.3 \scriptsize{(3.4)} & 58.2 \scriptsize{(4.7)} \\
GAAC-KL & 2.0 \scriptsize{(0.9)} & 21.7 \scriptsize{(3.5)} & 33.2 \scriptsize{(2.8)} & 36.2 \scriptsize{(2.8)} & 70.7 \scriptsize{(3.2)} & 59.7 \scriptsize{(4.3)} \\
GAAC-OB-KL & 35.0 \scriptsize{(2.9)} & 23.0 \scriptsize{(3.1)} & 56.0 \scriptsize{(2.6)} & 25.3 \scriptsize{(2.4)} & 73.2 \scriptsize{(2.8)} & 64.5 \scriptsize{(3.0)} \\
\bottomrule
    \end{tabular}
    \end{sc}
\end{small}
\end{center}
\end{table}

\begin{table*}[ht]
    \caption{Bootstrap mean and 95\% confidence interval of all evaluated methods in SMAC and Traffic Junction. We mark the maximum mean value in each column in bold and underline.}
\label{tab:MeanConfidence}
\begin{center}
\begin{small}
\begin{sc}
    \begin{tabular}{p{0.2\linewidth}p{0.1\linewidth}p{0.1\linewidth}p{0.1\linewidth}p{0.1\linewidth}p{0.1\linewidth}p{0.1\linewidth}}
        \toprule
          & 1o\_10b\_vs\_1r & 3s5z\_vs\_3s6z & 5m\_vs\_6m & 6h\_vs\_8z & medium & hard \\ \midrule

COMA & 33 (33,34) & 0 (0,0) & 0 (0,0) & 0 (0,0) & 60 (59,62) & 42 (40,43) \\
MAPPO & 21 (21,21) & 65 (64,66) & 32 (32,32) & 33 (33,34) & 71 (71,72) & 57 (56,58) \\
MAT & \underline{\textbf{58} (57,58)} & 4 (3,4) & 3 (2,3) & 13 (13,14) & 72 (71,73) & 57 (56,58) \\
IPPO-Comm & 2 (2,3) & 29 (28,30) & 22 (21,22) & 35 (34,35) & 70 (69,71) & 58 (57,59) \\
IPPO-Comm-OB & 27 (26,28) & 49 (48,50) & 70 (69,70) & 37 (36,37) & 69 (68,70) & 61 (60,61) \\
IPPO-Comm-KL & 1 (1,1) & 25 (24,26) & 33 (32,33) & 35 (35,36) & 71 (70,71) & 60 (59,60) \\
IPPO-Comm-OB-KL & 42 (42,43) & \underline{\textbf{70} (70,71)} & \underline{\textbf{72} (72,73)} & \underline{\textbf{56} (55,56)} & 72 (71,72) & 64 (63,64) \\
GAAC & 10 (10,11) & 0 (0,0) & 3 (2,3) & 0 (0,0) & 70 (69,71) & 59 (58,60) \\
GAAC-OB & 26 (25,26) & 0 (0,0) & 32 (31,32) & 34 (34,35) & 70 (69,70) & 58 (57,59) \\
GAAC-KL & 2 (1,2) & 21 (21,22) & 33 (32,33) & 36 (35,36) & 70 (70,71) & 59 (58,60) \\
GAAC-OB-KL & 35 (34,35) & 23 (22,23) & 56 (55,56) & 25 (24,25) & \underline{\textbf{73} (72,73)} & \underline{\textbf{64} (63,65)} \\
    \bottomrule
    \end{tabular}
    \end{sc}
\end{small}
\end{center}
\end{table*}

\end{document}